\documentclass[10pt,twocolumn,letterpaper]{article}

\usepackage{pgfplots}
\pgfplotsset{compat=newest}
\usepackage{tikz}

\usepackage{iccv}
\usepackage{times}
\usepackage{epsfig}
\usepackage{graphicx}
\usepackage{amsmath}
\usepackage{amssymb}
\usepackage{amsthm}

\usepackage[utf8x]{inputenc}
\usepackage{color}
\usepackage{xcolor}
\usepackage{booktabs,siunitx} 
\usepackage{makecell}
\usepackage{enumitem}
\usepackage[caption=false]{subfig}
\usepackage[noadjust]{cite}
\usepackage{rotating}
\usepackage[misc]{ifsym}
\usepackage{authblk}

\usepackage[pagebackref=true,breaklinks=true,colorlinks,bookmarks=false]{hyperref}

\makeatletter
\renewcommand\AB@affilsepx{, \protect\Affilfont}
\makeatother

\if@mathematic
   
\else
   
\fi

\definecolor{road}                {RGB}{128, 64,128}
\definecolor{sidewalk}            {RGB}{244, 35,232}
\definecolor{building}            {RGB}{ 70, 70, 70}
\definecolor{wall}                {RGB}{102,102,156}
\definecolor{fence}               {RGB}{190,153,153}
\definecolor{pole}                {RGB}{153,153,153}
\definecolor{traffic light}       {RGB}{250,170, 30}
\definecolor{traffic sign}        {RGB}{220,220,  0}
\definecolor{vegetation}          {RGB}{107,142, 35}
\definecolor{terrain}             {RGB}{152,251,152}
\definecolor{sky}                 {RGB}{ 70,130,180}
\definecolor{person}              {RGB}{220, 20, 60}
\definecolor{rider}               {RGB}{255,  0,  0}
\definecolor{car}                 {RGB}{  0,  0,142}
\definecolor{truck}               {RGB}{  0,  0, 70}
\definecolor{bus}                 {RGB}{  0, 60,100}
\definecolor{train}               {RGB}{  0, 80,100}
\definecolor{motorcycle}          {RGB}{  0,  0,230}
\definecolor{bicycle}             {RGB}{119, 11, 32}
\definecolor{void}                {RGB}{  0,  0,  0}

\newcommand{\vecnorm}[1]{\left\|#1\right\|}
\newtheorem{thm}{Theorem}

\newcommand\ver[1]{\rotatebox[origin=c]{90}{#1}}

\newcommand{\yes}{\checkmark}
\newcommand{\no}{$\times$}

\newcommand{\best}[1]{\textbf{#1}}

\newcommand{\PAR}[1]{\vskip4pt \noindent{\bf #1~}}

\iccvfinalcopy 

\def\iccvPaperID{0921} 

\begin{document}

\title{Guided Curriculum Model Adaptation and Uncertainty-Aware Evaluation for Semantic Nighttime Image Segmentation}

\author[1]{Christos Sakaridis}
\author[1]{Dengxin Dai}
\author[1,2]{Luc Van Gool}

\affil[1]{ETH Z\"urich}
\affil[2]{KU Leuven}

\maketitle

\begin{abstract}
Most progress in semantic segmentation reports on daytime images taken under favorable illumination conditions. We instead address the problem of semantic segmentation of nighttime images and improve the state-of-the-art, by adapting daytime models to nighttime without using nighttime annotations. Moreover, we design a new evaluation framework to address the substantial \emph{uncertainty} of semantics in nighttime images. Our central contributions are: 1) a curriculum framework to gradually adapt semantic segmentation models from day to night via labeled synthetic images and unlabeled real images, both for progressively darker times of day, which exploits cross-time-of-day correspondences for the real images to guide the inference of their labels; 2) a novel uncertainty-aware annotation and evaluation framework and metric for semantic segmentation, designed for adverse conditions and including image regions \emph{beyond human recognition capability} in the evaluation in a principled fashion; 3) the \emph{Dark Zurich} dataset, which comprises 2416 unlabeled nighttime and 2920 unlabeled twilight images with correspondences to their daytime counterparts plus a set of 151 nighttime images with fine pixel-level annotations created with our protocol, which serves as a first benchmark to perform our novel evaluation. Experiments show that our guided curriculum adaptation significantly outperforms state-of-the-art methods on real nighttime sets both for standard metrics and our uncertainty-aware metric. Furthermore, our uncertainty-aware evaluation reveals that selective invalidation of predictions can lead to better results on data with ambiguous content such as our nighttime benchmark and profit safety-oriented applications which involve invalid inputs.
\end{abstract}

\section{Introduction}
\label{sec:intro}

The state of the art in semantic segmentation is rapidly improving in recent years. Despite the advance, most methods are designed to operate at daytime, under favorable illumination conditions. However, many outdoor applications require robust vision systems that perform well at all times of day, under challenging lighting conditions, and in bad weather~\cite{vision:atmosphere}. Currently, the popular approach to solving perceptual tasks such as semantic segmentation is to train deep neural networks~\cite{pspnet,refinenet,dilated:convolution} using large-scale human annotations~\cite{pascal:2011,Cityscapes,Mapillary}. This supervised scheme has achieved great success for daytime images, but it scales badly to adverse conditions.  In this work, we focus on semantic segmentation at nighttime, both at the method level and the evaluation level. 

At the method level, this work adapts semantic segmentation models from daytime to nighttime, without annotations in the latter domain.
To this aim, we propose a new method called Guided Curriculum Model Adaptation (GCMA). The underpinnings of GCMA are threefold: power of time, power of place, and power of data. \textbf{Time}: environmental illumination changes continuously from daytime to nighttime. This enables adding intermediate domains between the two to smoothly transfer semantic knowledge. This idea is found to be effective in~\cite{SynRealDataFogECCV18,daytime:2:nighttime}; we extend it by adding two more modules. \textbf{Place}: images taken over different time but with the same 6D camera pose share a large portion of content. The shared content can be used to guide the knowledge transfer process from a favorable condition (daytime) to an adverse condition (nighttime). We formalize this observation and propose a solution for large-scale application. \textbf{Data}: GCMA takes advantage of the powerful image translation techniques to stylize large-scale real annotated daytime datasets to darker target domains in order to perform standard supervised learning.

The adversity of nighttime poses further challenges for perceptual tasks compared to daytime. The extracted features become corrupted due to visual hazards~\cite{cv:hazop} such as underexposure, noise, and motion blur. The degradation of affected input regions is often so intense that they are rendered \emph{indiscernible}, i.e.\ determining their semantic content is impossible even for humans. We term such regions as \emph{invalid} for the task of semantic segmentation. A robust model should predict with high \emph{uncertainty} on invalid regions while still being confident on valid (discernible) regions, and a sound evaluation framework should reward such behavior. The above requirement is particularly significant for safety-oriented applications such as autonomous cars, since having the vision system declare a prediction as invalid can help the downstream driving system avoid the fatal consequences of this prediction being false, e.g.\ when a pedestrian is missed.

To this end, we design a generic uncertainty-aware annotation and evaluation framework for semantic segmentation in adverse conditions which explicitly distinguishes invalid from valid regions of input images, and apply it to nighttime. On the annotation side, our novel protocol leverages privileged information in the form of daytime counterparts of the annotated nighttime scenes, which reveal a large portion of the content of invalid regions. This allows to reliably label invalid regions and to indeed \emph{include} invalid regions in the evaluation, contrary to existing semantic segmentation benchmarks~\cite{Cityscapes} which completely exclude them from evaluation. Moreover, apart from the standard class-level semantic annotation, each image is annotated with a mask which designates its invalid regions. On the evaluation side, we allow the \emph{invalid} label in predictions and adopt from~\cite{wilddash} the principle that for invalid pixels with legitimate semantic labels, both these labels and the \emph{invalid} label are considered correct predictions. However, this principle does not cover the case of valid regions. We address this by introducing the concept of false invalid predictions. This enables calculation of \emph{uncertainty-aware intersection-over-union (UIoU)}, a joint performance metric for valid and invalid regions which generalizes standard IoU, reducing to the latter when no invalid prediction exists. UIoU rewards predictions with confidence that is \emph{consistent} to human annotators, i.e.\ with higher confidence on valid regions than invalid ones, meeting the aforementioned requirement.

Finally, we present \emph{Dark Zurich}, a dataset of real images which contains corresponding images of the same driving scenes at daytime, twilight and nighttime. We use this dataset to feed real data to GCMA and to create a benchmark with 151 nighttime images for our uncertainty-aware evaluation. Our dataset and code are publicly available\footnote{\scriptsize{\url{https://trace.ethz.ch/projects/adverse/GCMA_UIoU}}}.

\section{Related Work} 
\label{sec:related} 

\PAR{Vision at Nighttime.}
Nighttime has attracted a lot of attention in the literature due to its ubiquitous nature. Several works pertain to human detection at nighttime, using FIR cameras~\cite{night:vision:pedestrian:05,pedestrian:detection:tracking:night:09}, visible light cameras~\cite{cnn:human:detection:nighttime:17}, or a combination of both~\cite{nighttime:pedestrian:detection:08}. In driving scenarios, a few methods have been proposed to detect cars~\cite{nighttime:object:proposal:18} and vehicles' rear lights~\cite{night:rear:lights:16}. Contrary to these domain-specific methods, previous work also includes both methods designed for robustness to illumination changes, by employing domain-invariant representations~\cite{road:detection:illumination:invariant,outdoor:transformation:labeling:iv15} or fusing information from complementary modalities and spectra~\cite{AdapNet:adverse:17}, and datasets with adverse illumination~\cite{Oxford,localization:benchmarking:adverse} for localization benchmarking.
A recent work~\cite{daytime:2:nighttime} on semantic nighttime segmentation shows that images captured at twilight are helpful for supervision transfer from daytime to nighttime. Our work is partially inspired by~\cite{daytime:2:nighttime} and extends it by proposing a guided curriculum adaptation framework which learns jointly from stylized images and unlabeled real images of increasing darkness and exploits scene correspondences.

\PAR{Domain Adaptation.}
Performance of semantic segmentation on daytime scenes has increased rapidly in recent years. As a consequence, attention is now turning to adaptation to adverse conditions~\cite{AdapNet:adverse:17,benchmark:sensor:adverse:weather:18,wulfmeier2017addressing,continuous:manifold:adaptation}. A case in point are recent efforts to adapt clear-weather models to fog~\cite{SFSU_synthetic,SynRealDataFogECCV18,CMAda:IJCV2019}, by using both labeled synthetic images and unlabeled real images of increasing fog density. This work instead focuses on the nighttime domain, which poses very different and---as we would claim---greater challenges than the foggy domain (e.g.\ artificial light sources casting very different illumination patterns at night). A major class of adaptation approaches, including~\cite{cyCADA,learning:synthetic:data:cvpr18,chen2018road,adapt:structured:output:cvpr18,incremental:adversarial:DA:18,conditional:GAN:adaptation,FCNs:adaptation,conservative:loss:adaptation,DCAN:adaptation,bidirectional:learning:adaptation}, involves adversarial confusion or feature alignment between domains.
The general concept of curriculum learning has been applied to domain adaptation by ordering tasks~\cite{curriculum:domain:adaptation:17} or target-domain pixels~\cite{self:training:adaptation}, while we order domains. Cross-domain correspondences as guidance have only been used very recently in~\cite{cross:season:correspondence}, which requires pixel-level matches, while we use more generic image-level correspondences.

\PAR{Semantic Segmentation Evaluation.}
Semantic segmentation evaluation is commonly performed with the IoU metric~\cite{pascal:2011}. Cityscapes~\cite{Cityscapes} introduced an instance-level IoU (iIoU) to remove the large-instance bias, as well as mean average precision for the task of instance segmentation. The two tasks have recently been unified into panoptic segmentation~\cite{panoptic:segmentation}, with a respective panoptic quality metric. The most closely related work to ours in this regard is WildDash~\cite{wilddash}, which uses standard IoU together with a fine-grained evaluation to measure the impact of visual hazards on performance. In contrast, we introduce UIoU, a new semantic segmentation metric that handles images with regions of uncertain semantic content and is suited for adverse conditions. Our uncertainty-aware evaluation is \emph{complementary} to uncertainty-aware methods such as~\cite{uncertainty:bayesian} that explicitly incorporate uncertainty in their model formulation and aims to promote the development of such methods, as UIoU rewards models that accurately capture heteroscedastic aleatoric uncertainty~\cite{uncertainty:bayesian} in the input images through the different treatment of invalid and valid regions.

\section{Guided Curriculum Model Adaptation} 
\label{sec:gcma}

\subsection{Problem Formulation}
\label{sec:gcma:general}
GCMA involves a source domain $\mathcal{S}$, an ultimate target domain $\mathcal{T}$, and an intermediate target domain $\dot{\mathcal{T}}$. In this work, $\mathcal{S}$ is daytime, $\mathcal{T}$ is nighttime, and $\dot{\mathcal{T}}$ is twilight time with an intermediate level of darkness between $\mathcal{S}$ and $\mathcal{T}$. GCMA adapts semantic segmentation models through this sequence of domains $(\mathcal{S}, \dot{\mathcal{T}}, \mathcal{T})$, which is sorted in ascending order with respect to level of darkness. The approach proceeds progressively and adapts the model from one domain in the sequence to the next. The knowledge is transferred through the domain sequence via this gradual adaptation process. The transfer is performed using two coupled branches: 1) learning from labeled synthetic stylized images and 2) learning from real data without annotations, to jointly leverage the assets of both. Stylized images inherit the human annotations of their original counterparts but contain unrealistic artifacts, whereas real images have less reliable pseudo-labels but are characterized by artifact-free textures.

Let us use $z \in \{1,2,3\}$ as the index in $(\mathcal{S}, \dot{\mathcal{T}}, \mathcal{T})$. 
Once the model for the current domain $z$ is trained, its knowledge can be distilled on unlabeled real data from $z$, and then used, along with a new version of synthetic data from the next domain $z+1$ to adapt the current model to $z+1$.

Before diving into the details, we first define all datasets used. 
The inputs for GCMA consist of: 1) a labeled daytime set with $M$ real images $\mathcal{D}^1_{lr}=\{(I_m^1,Y^1_m)\}_{m=1}^M$, e.g.\ Cityscapes~\cite{Cityscapes}, where $Y_m^{1}(i,j) \in \mathcal{C} = \{1, ..., C\}$ is the ground-truth label of pixel $(i,j)$ of $I_m^{1}$; 2) an unlabeled daytime set of $N_1$ images $\mathcal{D}^1_{ur}=\{I_n^1\}_{n=1}^{N_1}$; 3) an unlabeled twilight set of $N_2$ images $\mathcal{D}^2_{ur}=\{I_n^2\}_{n=1}^{N_2}$; and 4) an unlabeled nighttime set of $N_3$ images $\mathcal{D}_{ur}^3=\{I_n^3\}_{n=1}^{N_3}$. In order to perform knowledge transfer with annotated data, $\mathcal{D}_{lr}^1$ is rendered in the style of $\mathcal{D}^2_{ur}$ and $\mathcal{D}_{ur}^3$. We use CycleGAN~\cite{cycleGAN} to perform this style transfer, leading to two more sets: $\mathcal{D}^2_{ls}=\{(\bar{I}_m^2,Y^1_m)\}_{m=1}^M$ and $\mathcal{D}^3_{ls}=\{(\bar{I}_m^3,Y^1_m)\}_{m=1}^M$, where $\bar{I}_m^2$ and $\bar{I}_m^3$ are the stylized twilight and nighttime version of $I_m^1$ respectively, and labels are copied. 
For $z=1$, the semantic segmentation model $\phi^1$ is trained directly on $\mathcal{D}^1_{lr}$.
In order to perform knowledge transfer with unlabeled data, pseudo-labels for all three unlabeled real datasets need to be generated. The pseudo-labels for $\mathcal{D}_{ur}^1$ are generated using the model $\phi^1$ via $\hat{Y}_n^1=\phi^1(I_n^1)$. 
For $z>1$, training $\phi^z$ and generating $\hat{Y}_m^z$ is performed progressively as GCMA proceeds, as is detailed in Sec.~\ref{sec:gcma:learning}. All six datasets are summarized in Table~\ref{tab:GCMA:notations}.

\setlength{\tabcolsep}{1pt}
\begin{table}
\caption{The training sets used in GCMA. $I$ indicates an image and $Y$ its label map; $\bar{I}$ is a synthetic image and $\hat{Y}$ a pseudo-label map. See the text for details.}
    \label{tab:GCMA:notations}
    \centering
    \footnotesize
    \begin{tabular}{lccc}
 \toprule
 &  \multicolumn{2}{c}{Labeled}  & Unlabeled   \\
 & Real & Synthetic & Real  \\
 1. Daytime & $\{(I_m^1,Y^1_m)\}_{m=1}^M$ & &  $\{(I_n^1,\hat{Y}^1_n)\}_{n=1}^{N_1}$  \\
 2. Twilight time & & $\{(\bar{I}_m^2,Y^1_m)\}_{m=1}^M$ &   $\{(I_n^2,\hat{Y}^2_n)\}_{n=1}^{N_2}$  \\
 3. Nighttime & & $\{(\bar{I}_m^3,Y^1_m)\}_{m=1}^M$ &  $\{(I_n^3,\hat{Y}^3_n)\}_{n=1}^{N_3}$ \\ 
 \bottomrule
\end{tabular}
\end{table}

\subsubsection{Guided Curriculum Model Adaptation} 
\label{sec:gcma:learning}
Since the method proceeds in an iterative manner, we present the algorithmic details only for a single adaptation step from $z-1$ to $z$. The presented algorithm is straightforward to generalize to multiple intermediate target domains.
In order to adapt the semantic segmentation model $\phi^{z-1}$ from the previous domain $z-1$ to the current domain $z$, we generate synthetic stylized data in domain $z$: $\mathcal{D}_{ls}^z$.   

For real unlabeled images, since no human annotations are available, we rely on a strategy of self-learning or curriculum learning. Our motivating assumption is that objects are generally easier to recognize in lighter conditions, so the tasks are solved in ascending order with respect to the level of darkness and the easier, solved tasks are used to re-train the model to further solve the harder tasks. This is in line with the concept of curriculum learning~\cite{curriculum:learning}.  
In particular, the model $\phi^{z-1}$ for domain $z-1$ can be applied to the unlabeled real images of domain $z-1$ to generate supervisory labels for training $\phi^{z}$. Specifically, the dataset of real images with pseudo-labels for adaptation to domain $z$ is $\mathcal{D}_{ur}^{z-1}=\{(I_n^{z-1}, \hat{Y}_n^{z-1})\}_{n=1}^{N_{z-1}}$, where $\hat{Y}_n^{z-1}$ denotes the predicted labels of image $I_n^{z-1}$. A simple way to get these labels is by directly feeding $I_n^{z-1}$ to $\phi^{z-1}$, similar to the approach of~\cite{SynRealDataFogECCV18,CMAda:IJCV2019} for the case of fog. This choice, however, suffers from accumulation of substantial errors in the prediction of $\phi^{z-1}$ into the subsequent training step if domain $z-1$ is not the daytime domain. We instead propose a method to refine these errors by using \emph{guidance} from the semantics of a daytime image $I_n^{1}$ that \emph{corresponds} to $I_n^{z-1}$, i.e.\ depicts roughly the same scene as $I_n^{z-1}$ (the difference in the camera pose is small):
\begin{equation} \label{eq:guided:prediction} 
\hat{Y}_n^{z-1} = G\left(\phi^{z-1}(I_n^{z-1}), I_n^{z-1}, \phi^{1}(I_{A_{z-1 \rightarrow 1}(n)}^{1})\right),
\end{equation}
where $G$ is a guidance function which will be defined in Sec.~\ref{sec:gcma:guidance} and $z-1 > 1$. $A_{z-1 \rightarrow 1}(n)$ is the correspondence function giving the index of the daytime image that corresponds to $I_n^{z-1}$.

Once we have the two training sets $\mathcal{D}^{z-1}_{ur}$ (with labels inferred through \eqref{eq:guided:prediction}) and $\mathcal{D}^z_{ls}$, learning $\phi^z$ is performed by optimizing a loss function that involves both datasets:
\begin{equation} \label{eq:ssl}
\min_{\phi^z}  \bigg(\sum_{\substack{(I, Y)  \\ \in \mathcal{D}_{ls}^z}} L(\phi^z(I),Y) + \mu \sum_{\substack{(I, \hat{Y}) \\ \in \mathcal{D}_{ur}^{z-1}}} L(\phi^z(I), \hat{Y}) \bigg),
\end{equation}
where $L(.,.)$ is the cross entropy loss and $\mu$ is a hyper-parameter balancing the contribution of the two datasets.

In order to leverage the \emph{place} prior at large scale to improve predictions through the guided label refinement defined in \eqref{eq:guided:prediction}, specific aligned datasets need to be compiled. With this aim, we collected the \emph{Dark Zurich} dataset by driving several laps in disjoint areas of Zurich; each lap was driven multiple times during the same day, starting from daytime through twilight to nighttime. The recordings include GPS readings and are split into three sets: daytime, twilight and nighttime (cf.\ Sec.~\ref{sec:dark:zurich}). Since different drives of the same lap correspond to the same route, the camera orientation at a certain point of the lap is similar across all drives. We implement the correspondence function $A_{z\rightarrow1}$ that assigns to each image in domain $z$ its daytime counterpart using a GPS-based nearest neighbor assignment. The method presented in Sec.~\ref{sec:gcma:guidance} carefully handles the effects of misalignment and dynamic objects in paired images.

\subsection{Guided Segmentation Refinement}
\label{sec:gcma:guidance}

\begin{figure*}
    \centering
    \subfloat[Dark image $I^z$]{\includegraphics[width=0.24\textwidth]{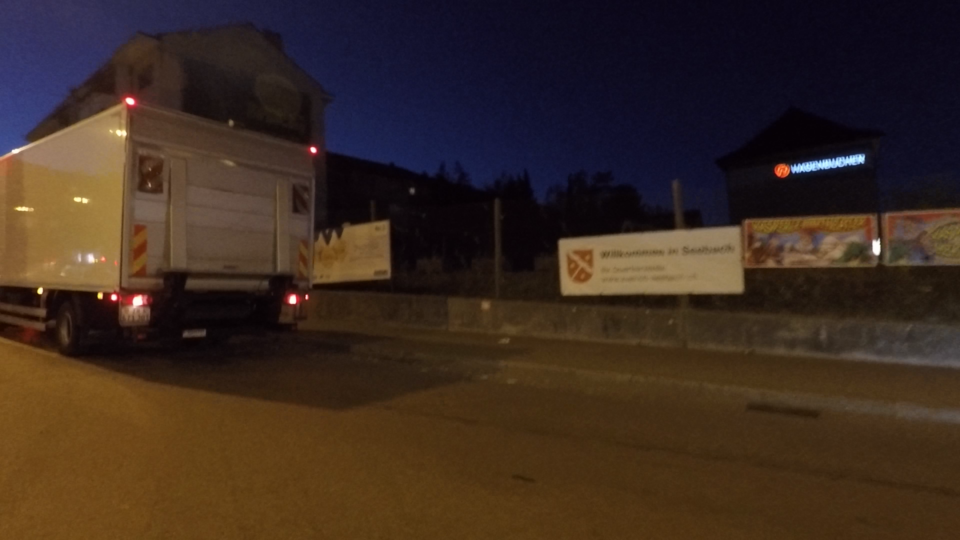}}\label{fig:refinement:twilight}
    \hfil
    \subfloat[Daytime image $I^{1}$]{\includegraphics[width=0.24\textwidth]{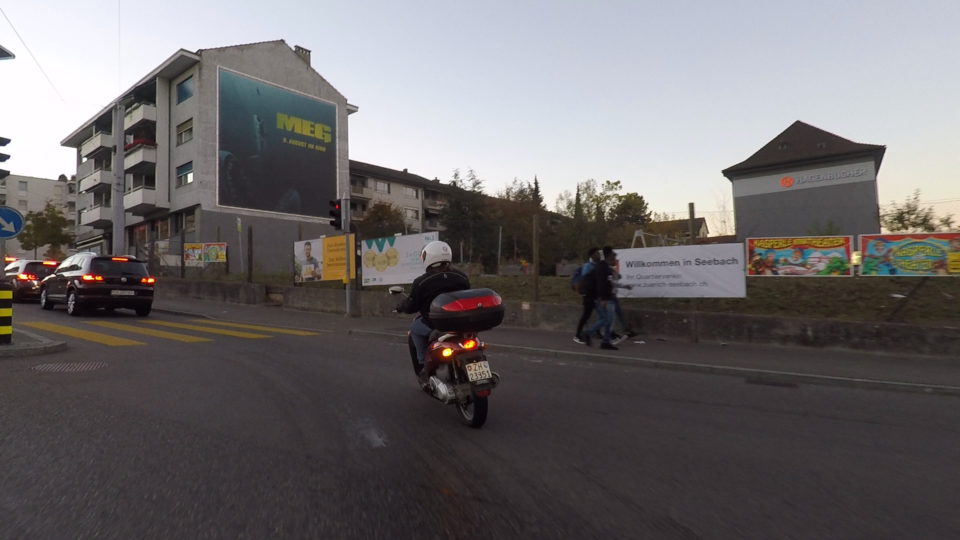}}\label{fig:refinement:day}
    \hfil
    \subfloat[Initial prediction $\mathbf{S}^z$ for $I^z$]{\includegraphics[width=0.24\textwidth]{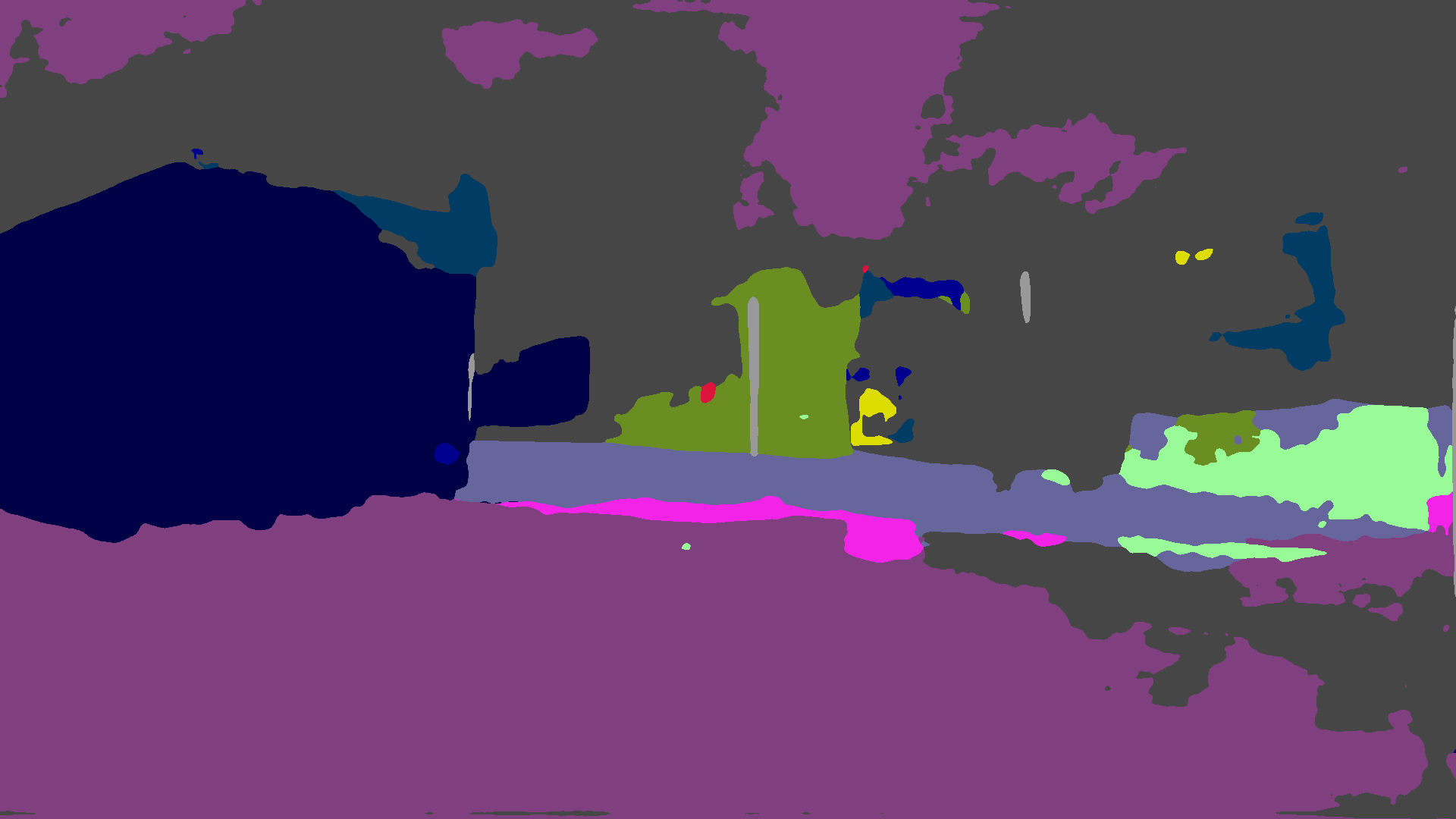}}\label{fig:refinement:init}
    \hfil
    \subfloat[Our refined prediction $\hat{\mathbf{S}}^{z}$ for $I^z$]{\includegraphics[width=0.24\textwidth]{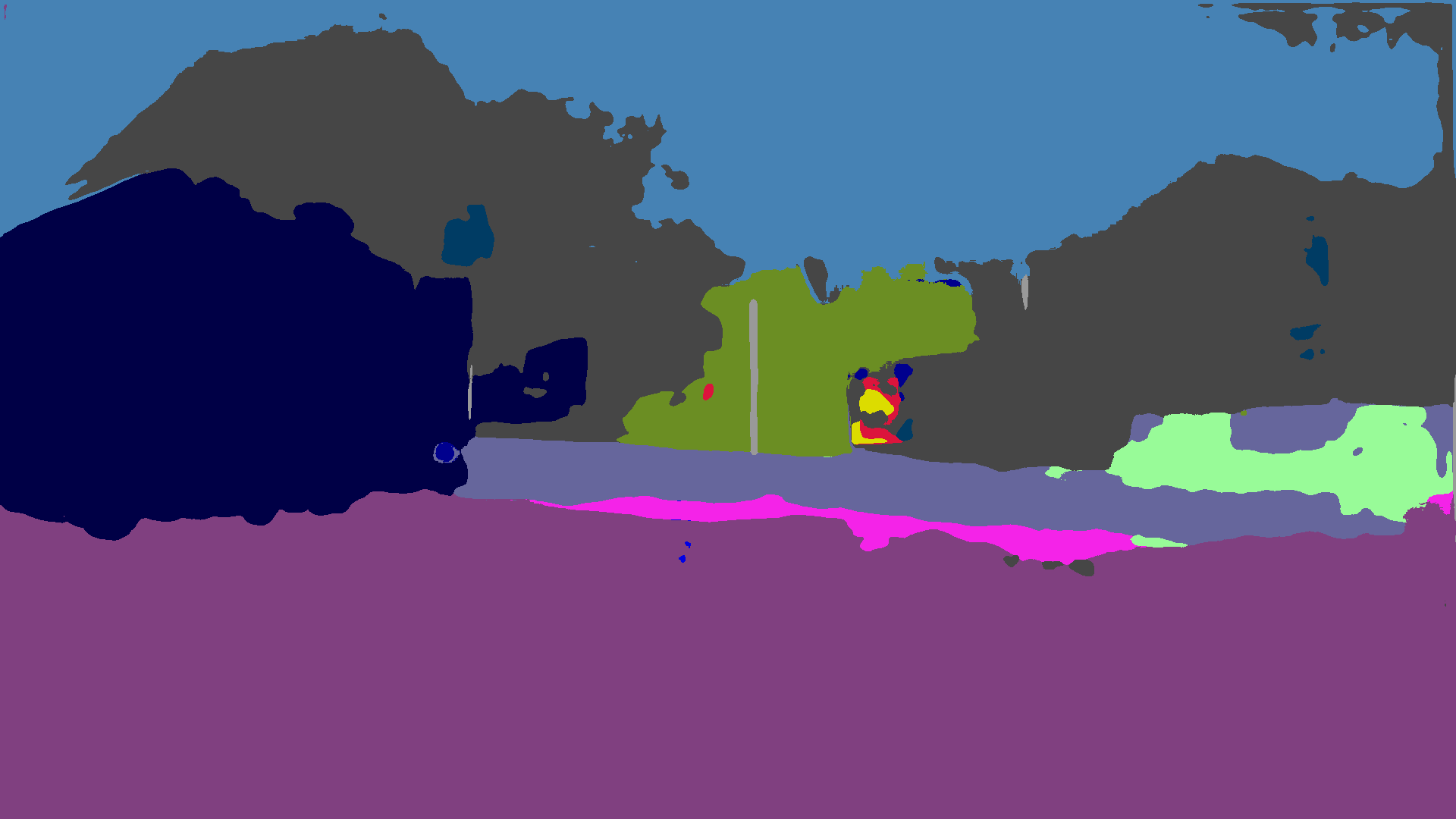}}\label{fig:refinement:ours}
    \caption{Example pair of corresponding images from \emph{Dark Zurich}, initial prediction for the dark image and our refined prediction.}
    \label{fig:refinement}
\end{figure*}

In the following presentation of our guided segmentation refinement for dark images using corresponding daytime images, we drop for brevity the subscript which was used to indicate this correspondence. The guidance function $G$ which models our refinement approach and was introduced in a  general form in \eqref{eq:guided:prediction} can be written more specifically as
\begin{equation} \label{eq:guidance:explicit}
G\left(\phi^z(I^z), I^z, \phi^1(I^1)\right) = R\left(\phi^z(I^z),\,B(\phi^1(I^1), I^z)\right),
\end{equation}
i.e.\ as the composition of a cross bilateral filter $B$ on the daytime predictions, which aligns them to the dark image, with a fusion function $R$, which adaptively combines the aligned daytime predictions with the initial dark image predictions to refine the latter.

\subsubsection{Cross Bilateral Filter for Prediction Alignment}
\label{sec:gcma:guidance:bilateral}

The correspondences between real images that are used in GCMA are not perfect, in the sense that they are not aligned at a pixel-accurate level. Therefore, to leverage the prediction for the daytime image $I^{1}$ as \emph{guidance} for refining the respective prediction for the dark image $I^{z}$, it is necessary to first align the former prediction to $I^{z}$. To this end, we operate on \emph{soft} predictions and define a cross bilateral filter on the initial soft prediction map $\mathbf{S}^1 = \phi^{1}(I^1)$ which uses the color of the dark image $I^{z}$ as reference:
\begin{align}
&\tilde{\mathbf{S}}^{1}(\mathbf{p}) \nonumber\\
&{=}\;\frac{\displaystyle\sum_{\mathbf{q} \in \mathcal{N}(\mathbf{p})} G_{\sigma_s}(\vecnorm{\mathbf{q} - \mathbf{p}}) G_{\sigma_r}(\vecnorm{I^{z}(\mathbf{q}) - I^{z}(\mathbf{p})}) \mathbf{S}^{1}(\mathbf{q}) }{ \displaystyle\sum_{\mathbf{q} \in \mathcal{N}(\mathbf{p})} G_{\sigma_s}(\vecnorm{\mathbf{q} - \mathbf{p}}) G_{\sigma_r}(\vecnorm{I^{z}(\mathbf{q}) - I^{z}(\mathbf{p})}) }. \label{eq:cross:bilateral}
\end{align}
In \eqref{eq:cross:bilateral}, $\mathbf{p}$ and $\mathbf{q}$ denote pixel positions, $\mathcal{N}(\mathbf{p})$ is the neighborhood of $\mathbf{p}$, $G_{\sigma_s}$ is the spatial-domain Gaussian kernel and $G_{\sigma_r}$ is the color-domain kernel. The definition of the filter implies that only pixels $\mathbf{q}$ with similar color to the examined pixel $\mathbf{p}$ in the dark image $I^{z}$ contribute to the output $\tilde{\mathbf{S}}^{1}(\mathbf{p})$, which shifts salient edges in the initial daytime prediction to their correct position in the dark image. For the color-domain kernel, we use the CIELAB version of $I^{z}$, as it is more appropriate for measuring color similarity~\cite{bilateral:grid}. We set the spatial parameter $\sigma_s$ to $80$ to account for large misalignment, and $\sigma_r$ to $10$ following~\cite{bilateral:grid,SynRealDataFogECCV18}.

\subsubsection{Confidence-Adaptive Prediction Fusion}
\label{sec:gcma:guidance:fusion}

The final step in our refinement approach is to fuse the aligned prediction $\tilde{\mathbf{S}}^{1}$ for $I^{1}$ with the initial prediction $\mathbf{S}^{z} = \phi^z(I^{z})$ for $I^{z}$ in order to obtain the refined prediction $\hat{\mathbf{S}}^{z}$, the hard version of which is subsequently used in training. We propose an adaptive fusion scheme, which uses the \emph{confidence} associated with the two predictions at each pixel to weigh their contribution in the output and addresses disagreements due to dynamic content by properly adjusting the fusion weights. Let us denote the confidence of the aligned prediction $\tilde{\mathbf{S}}^{1}$ for $I^{1}$ at pixel $\mathbf{p}$ by $F^{1}(\mathbf{p}) = \max_{c \in \mathcal{C}} \tilde{S}_c^1(\mathbf{p})$ and respectively the confidence of the initial prediction $\mathbf{S}^{z}$ for $I^{z}$ by $F^{z}(\mathbf{p})$. Our confidence-adaptive fusion is then defined as
\begin{equation} \label{eq:fusion}
\hat{\mathbf{S}}^{z} = \frac{F^{z}}{F^{z} + \alpha F^{1}} \mathbf{S}^{z} + \frac{\alpha F^{1}}{F^{z} + \alpha F^{1}} \tilde{\mathbf{S}}^{1},
\end{equation}
where $0 < \alpha = \alpha(\mathbf{p}) \leq 1$ may vary and we have completely dropped the pixel argument $\mathbf{p}$ for brevity. In this way, we allow the daytime image prediction to have a greater effect on the output at regions of the dark image which were not easy for model $\phi^z$ to classify, while preserving the initial prediction $\mathbf{S}^{z}$ at lighter regions of the dark image where $\mathbf{S}^{z}$ is more reliable.

Our fusion distinguishes between dynamic and static scene content by regulating $\alpha$. In particular, $\alpha$ downweights $\tilde{\mathbf{S}}^{1}$ to induce a preference towards $\mathbf{S}^{z}$ when both predictions have high confidence. However, apart from imperfect alignment, the two scenes also differ due to dynamic content. Intuitively, the prediction of a dynamic object in the daytime image should be assigned an even lower weight in case the corresponding prediction in the dark image does not agree, since this object might only be present in the former scene. More formally, we denote the subset of $\mathcal{C}$ that includes dynamic classes by $\mathcal{C}_d$ and define
\begin{align}
&\alpha(\mathbf{p}) \nonumber\\
&{=}\;\left\{
\begin{array}{cl}
    \alpha_l, & \text{if } c_1 = \arg\displaystyle\max_{c \in \mathcal{C}} \tilde{S}_c^1(\mathbf{p}) \in \mathcal{C}_d \text{ and } S_{c_1}^z(\mathbf{p}) \leq \eta \\
     & \text{ or } c_2 = \arg\displaystyle\max_{c \in \mathcal{C}} S_c^z(\mathbf{p}) \in \mathcal{C}_d \text{ and } \tilde{S}_{c_2}^1(\mathbf{p}) \leq \eta, \\
    \alpha_h & \text{otherwise.}
\end{array}
\right.
\label{eq:alpha}
\end{align}
In our experiments, we manually tune $\alpha_l = 0.3$, $\alpha_h = 0.6$ and $\eta = 0.2$ on a couple of training images (no grid search). A result of our guided refinement is shown in Fig.~\ref{fig:refinement}.

\section{Uncertainty-Aware Evaluation}
\label{sec:evaluation} 

Images taken under adverse conditions such as nighttime contain invalid regions, i.e.\ regions with indiscernible semantic content. Invalid regions are closely related to the concept of negative test cases which was considered in~\cite{wilddash}. However, invalid regions constitute intra-image entities and can co-exist with valid regions in the same image, whereas a negative test case refers to an entire image that should be treated as invalid. We build upon the evaluation of~\cite{wilddash} for negative test cases and generalize it to be applied uniformly to all images in the evaluation set, whether they contain invalid regions or not. Our annotation and evaluation framework includes invalid regions in the set of evaluated pixels, but treats them differently from valid regions to account for the high uncertainty of their content. In the following, we elaborate on the generation of ground-truth annotations using privileged information through the day-night correspondences of our dataset and present our UIoU metric.

\subsection{Annotation with Privileged Information}
\label{sec:evaluation:annotation}


\begin{figure*}
    \centering
    \subfloat[Input image $I$]{\includegraphics[width=0.24\textwidth]{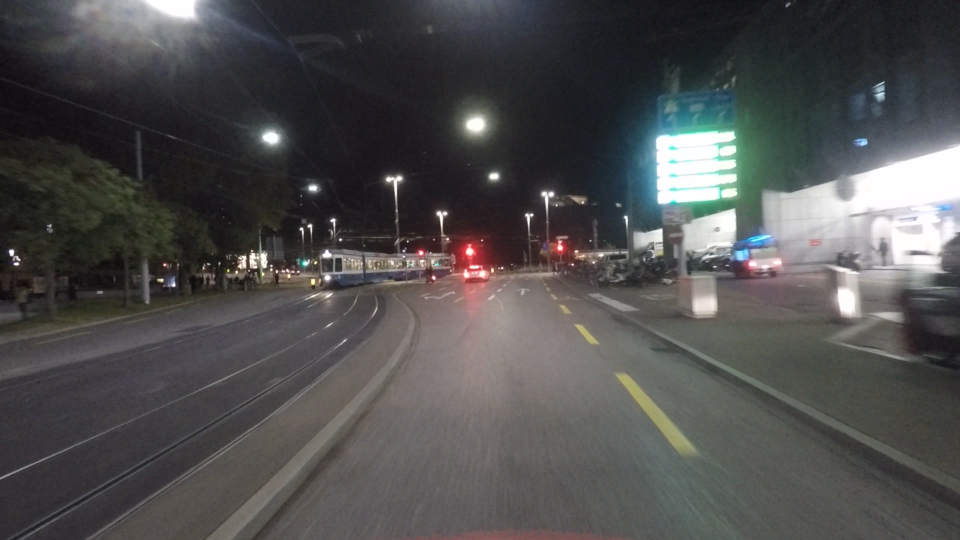}}\label{fig:annotation:input}
    \hfil
    \subfloat[Auxiliary image $I^{\prime}$]{\includegraphics[width=0.24\textwidth]{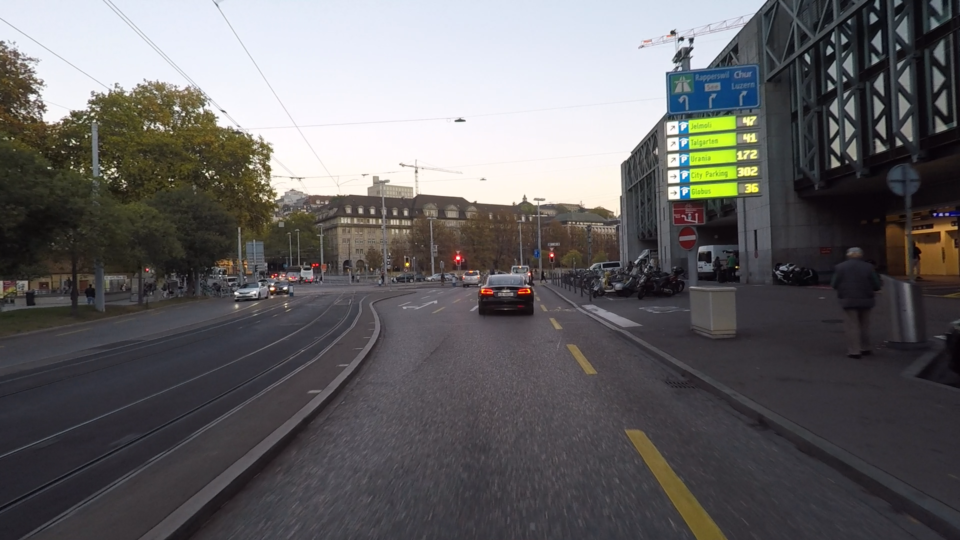}}\label{fig:annotation:auxiliary}
    \hfil
    \subfloat[GT invalid mask $J$]{\includegraphics[width=0.24\textwidth]{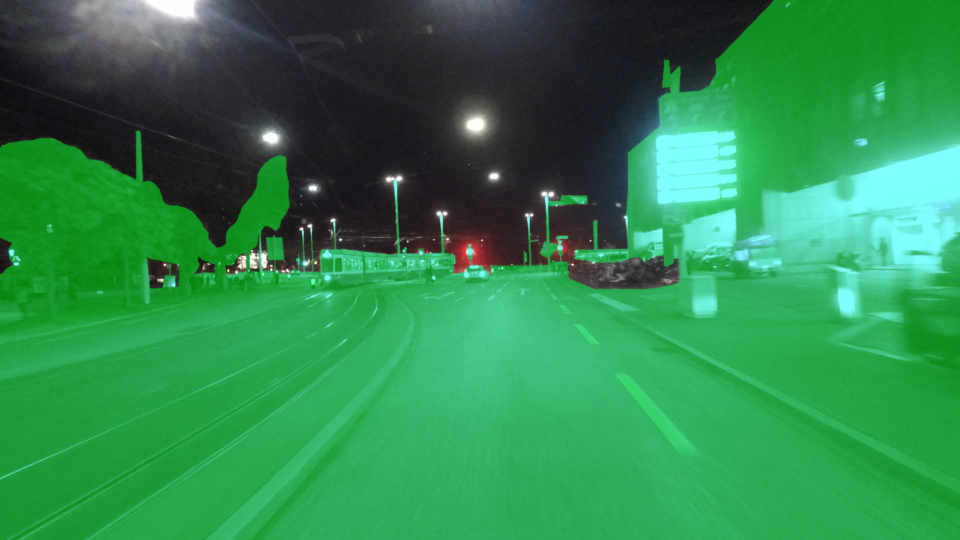}}\label{fig:annotation:invalid}
    \hfil
    \subfloat[GT semantic labeling $H$]{\includegraphics[width=0.24\textwidth]{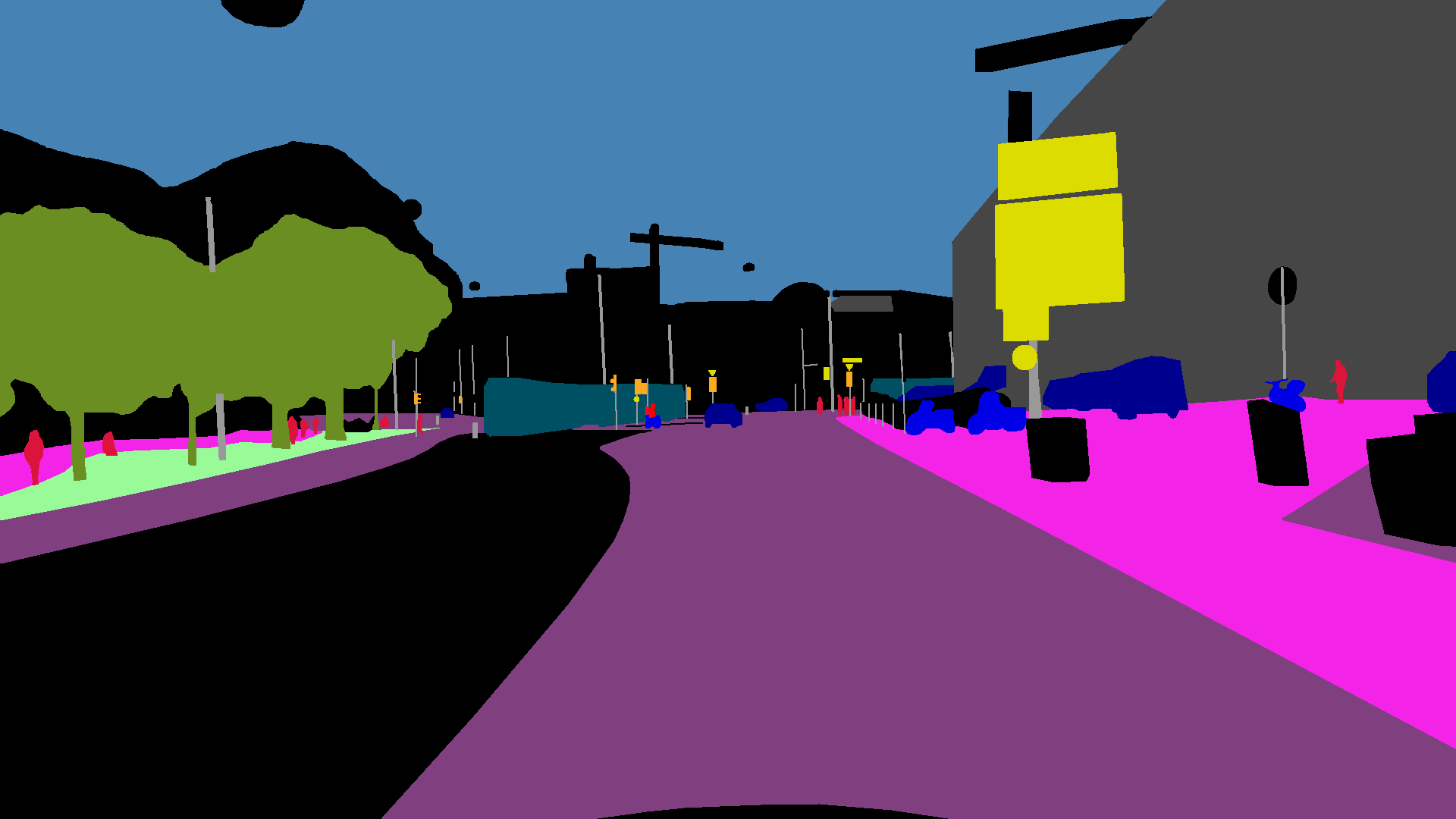}}\label{fig:annotation:gt}
    \caption{Example input images from \emph{Dark Zurich-test} and output annotations with our protocol. Valid pixels in $J$ are marked green.}
    \label{fig:annotation}
\end{figure*}

For each image $I$, the annotation process involves two steps: 1) creation of the ground-truth invalid mask $J$, and 2) creation of the ground-truth semantic labeling $H$.

For the semantic labels, we consider a predefined set $\mathcal{C}$ of $C$ classes, which is equal to the set of Cityscapes~\cite{Cityscapes} evaluation classes ($C = 19$). The annotator is first presented only with $I$ and is asked to mark the valid regions in it as the regions which she can unquestionably assign to one of the $C$ classes or declare as not belonging to any of them. The result of this step is the invalid mask $J$, which is set to 0 at valid pixels and 1 at invalid pixels.

Secondly, the annotator is asked to mark the semantic labels of $I$, only that this time she also has access to an \emph{auxiliary} image $I^{\prime}$. This latter image has been captured with roughly the same 6D camera pose as $I$ but under more favorable conditions. In our dataset, $I^{\prime}$ is captured at daytime whereas $I$ is captured at nighttime. The large overlap of static scene content between the two images allows the annotator to label certain regions in $H$ with a legitimate semantic label from $\mathcal{C}$, even though the same regions have been annotated as invalid (and are kept as such) in $J$. This allows joint evaluation on valid and invalid regions, as it creates regions which can accept both the \emph{invalid} label and the ground-truth label from $\mathcal{C}$ as correct predictions. Due to the imperfect match of the camera poses for $I$ and $I^{\prime}$, the labeling of invalid regions in $H$ is done conservatively, marking a coarse boundary which may leave unlabeled zones around the true semantic boundaries in $I$, so that no pixel is assigned a wrong label. The parts of $I$ which remain indiscernible even after inspection of $I^{\prime}$ are left unlabeled in $H$. These parts as well as instances of classes outside $\mathcal{C}$ are not considered during evaluation. We illustrate a visual example of our annotation inputs and outputs in Fig.~\ref{fig:annotation}.

\subsection{Uncertainty-Aware Predictions}


The semantic segmentation prediction that is fed to our evaluation is expected to include pixels labeled as \emph{invalid}. Instead of defining a separate, explicit \emph{invalid} class, which would potentially require the creation of new training data to incorporate this class, we allow a more flexible approach for soft predictions with the original set of semantic classes by using a \emph{confidence threshold}, which affords an evaluation curve for our UIoU metric by varying this threshold.

In particular, we assume that the evaluated method outputs an intermediate soft prediction $\mathbf{S}(\mathbf{p})$ at each pixel $\mathbf{p}$ as a probability distribution among the $C$ classes, which is subsequently converted to a hard assignment by outputting the class $\tilde{H}(\mathbf{p}) = \arg\max_{c \in \mathcal{C}}\{S_c(\mathbf{p})\}$ with the highest probability. In this case, $S_{\tilde{H}(\mathbf{p})}(\mathbf{p}) \in [1/C,\,1]$ is the effective confidence associated with the prediction. This assumption is not very restrictive, as most recent semantic segmentation methods are based on CNNs with a softmax layer that outputs such soft predictions.

The final evaluated output $\hat{H}$ is computed based on a free parameter $\theta \in [1/C,\,1]$ which acts as a confidence threshold by invalidating those pixels where the confidence of the prediction is lower than $\theta$, i.e.\ $\hat{H}(\mathbf{p}) = 
\tilde{H}(\mathbf{p})$ if $S_{\tilde{H}(\mathbf{p})}(\mathbf{p}) \geq \theta$ and \emph{invalid} otherwise. Increasing $\theta$ results in more pixels being predicted as \emph{invalid}. This approach is motivated by the fact that ground-truth invalid regions are identified during annotation by the uncertainty of their semantic content, which implies that a model should ideally place lower confidence (equivalently higher uncertainty) in predictions on invalid regions than on valid ones, so that the former get invalidated for lower values of $\theta$ than the latter. The formulation of our UIoU metric rewards this behavior as we shall see next. Note that our evaluation does not strictly require soft predictions, as UIoU can be normally computed for fixed, hard predictions $\hat{H}$.

\subsection{UIoU}


We propose UIoU as a generalization of the standard IoU metric for evaluation of semantic segmentation predictions which may contain pixels labeled as \emph{invalid}. UIoU reduces to standard IoU if no pixel is predicted to be invalid, e.g.\ when $\theta = 1/C$.

The calculation of UIoU for class $c$ involves five sets of pixels, which are listed along with their symbols: true positives (TP), false positives (FP), false negatives (FN), true invalids (TI), and false invalids (FI). Based on the ground-truth invalid masks $J$, the ground-truth semantic labelings $H$ and the predicted labels $\hat{H}$ for the set of evaluation images, these five sets are defined as follows:
\begin{align}
\text{TP} &= \{\mathbf{p}: H(\mathbf{p}) = \hat{H}(\mathbf{p}) = c\}, \label{eq:tp}\\
\text{FP} &= \{\mathbf{p}: H(\mathbf{p}) \neq c \text{ and } \hat{H}(\mathbf{p}) = c\}, \label{eq:fp}\\
\text{FN} &= \{\mathbf{p}: H(\mathbf{p}) = c \text{ and } \hat{H}(\mathbf{p}) \notin \{c,\,\text{\emph{invalid}}\}\}, \label{eq:fn}\\
\text{TI} &= \{\mathbf{p}: H(\mathbf{p}) = c \text{ and } \hat{H}(\mathbf{p}) = \text{\emph{invalid} and } J(\mathbf{p}) = 1\}, \label{eq:ti}\\
\text{FI} &= \{\mathbf{p}: H(\mathbf{p}) = c \text{ and } \hat{H}(\mathbf{p}) = \text{\emph{invalid} and } J(\mathbf{p}) = 0\}. \label{eq:fi}
\end{align}
UIoU for class $c$ is then defined as
\begin{equation} \label{eq:uiou}
\text{UIoU} = \frac{|\text{TP}| + |\text{TI}|}{|\text{TP}| + |\text{TI}| + |\text{FP}| + |\text{FN}| + |\text{FI}|}.
\end{equation}
Note that a true invalid prediction results in equal reward to predicting the correct semantic label of the pixel. Moreover, an invalid prediction does not come at no cost: it incurs the same penalty on valid pixels as predicting an incorrect label.

When dealing with multiple classes, we modify our notation to $\text{UIoU}^{(c)}$ (similarly for the five sets of pixels related to class $c$), which we avoided in the previous definitions to reduce clutter. The overall semantic segmentation performance on the evaluation set is reported as the mean UIoU over all $C$ classes. By varying the confidence threshold $\theta$ and using the respective output, we obtain a parametric expression $\text{UIoU}(\theta)$. When $\theta = 1/C$, no pixel is predicted as invalid and thus $\text{UIoU}(1/C) = \text{IoU}$.

We motivate the usage of UIoU instead of standard IoU in case the test set includes ground-truth invalid masks by showing in Th.~\ref{thm:UIoU:greater:iou} that UIoU is guaranteed to be larger than IoU for some $\theta > 1/C$ under the assumption that predictions on invalid regions are associated with lower confidence than those on valid regions, which lies in the heart of our evaluation framework. The proof is in Appendix~\ref{supp:sec:proof}.

\begin{thm} \label{thm:UIoU:greater:iou}
Assume that there exist $\theta_1$, $\theta_2$ such that $\theta_1 < \theta_2$, $\forall p: J(p) = 1 \Rightarrow S_{\tilde{H}(p)}(p) \leq \theta_1$ and $J(p) = 0 \Rightarrow S_{\tilde{H}(p)}(p) \geq \theta_2$. If we additionally assume that $\exists p \in \text{\emph{FN}}^{(c)}(1/C) \cup \text{\emph{FP}}^{(c)}(1/C): J(p) = 1$, then $\text{\emph{IoU}}^{(c)} < \text{\emph{UIoU}}^{(c)}(\theta_1)$.
\end{thm}

\section{The Dark Zurich Dataset}
\label{sec:dark:zurich}

\begin{table}[!tb]
    \caption{Comparison of \emph{Dark Zurich} against related datasets with nighttime semantic annotations. ``Night annot.'': annotated nighttime images, ``Invalid'': can invalid regions get legitimate labels?}
    \label{table:datasets:comparison}
    \centering
    \footnotesize
    \setlength\tabcolsep{1.2pt}
    \begin{tabular}{lccccc}
    \toprule
    Dataset & Night annot. & Classes & Reliable GT & Fine GT & Invalid \\
    \midrule
    WildDash~\cite{wilddash} & 13 & 19 & \yes & \yes & \no \\
    Raincouver~\cite{raincouver} & 95 & 3 & \yes & \no & \no \\
    BDD100K~\cite{BDD100K} & 345 & 19 & \no & \yes & \no \\
    Nighttime Driving~\cite{daytime:2:nighttime} & 50 & 19 & \yes & \no & \no \\
    Dark Zurich & 151 & 19 & \yes & \yes & \yes \\
    \bottomrule
    \end{tabular}
\end{table}

\emph{Dark Zurich} was recorded in Zurich using a 1080p GoPro Hero 5 camera, mounted on top of the front windshield of a car. The collection protocol with multiple drives of several laps to establish correspondences is detailed in Sec.~\ref{sec:gcma}.

We split \emph{Dark Zurich} and reserve one lap for testing. The rest of the laps remain unlabeled and are used for training. They comprise 3041 daytime, 2920 twilight and 2416 nighttime images extracted at 1 fps, which are named \emph{Dark Zurich}-\{\emph{day}, \emph{twilight}, \emph{night}\} respectively and correspond to the three sets in the rightmost column of Table~\ref{tab:GCMA:notations}. From the testing night lap, we extract one image every 50m or 20s, whichever comes first, and assign to it the corresponding daytime image to serve as the auxiliary image $I^{\prime}$ in our annotation (cf.\ Sec.~\ref{sec:evaluation:annotation}). We annotate 151 nighttime images with fine pixel-level Cityscapes labels and invalid masks following our protocol and name this set \emph{Dark Zurich-test}. In total, 272.2M pixels have been annotated with semantic labels and 56.7M of these pixels are marked as invalid. We validate the quality of our annotations by having 20 images annotated twice by different subjects and measuring consistency. 93.5\% of the labeled pixels are consistent in the semantic annotations and respectively 95\% in the invalid masks.
We compare to existing annotated nighttime sets in Table~\ref{table:datasets:comparison}, noting that most large-scale sets for road scene parsing, such as Cityscapes~\cite{Cityscapes} and Mapillary Vistas~\cite{Mapillary}, contain few or no nighttime scenes. Nighttime Driving~\cite{daytime:2:nighttime} and Raincouver~\cite{raincouver} only include \emph{coarse} annotations. \emph{Dark Zurich-test} contains ten times more nighttime images than WildDash~\cite{wilddash}---the only other dataset with \emph{reliable} fine nighttime annotations. Detailed inspection showed that $\sim$70\% of the 345 densely annotated nighttime images of BDD100K~\cite{BDD100K} contain severe labeling errors which render them unsuitable for evaluation, especially in dark regions we treat as invalid (e.g.\ \emph{sky} is often mislabeled as \emph{building}). Our annotation protocol helps avoid such errors by properly defining invalid regions and using daytime images to aid annotation, and \emph{Dark Zurich-test} is an initial high-quality benchmark to promote our uncertainty-aware evaluation.

\section{Results}
\label{sec:exp}

\begin{figure*}[!tb]
    \centering
    \subfloat{\includegraphics[width=0.195\textwidth]{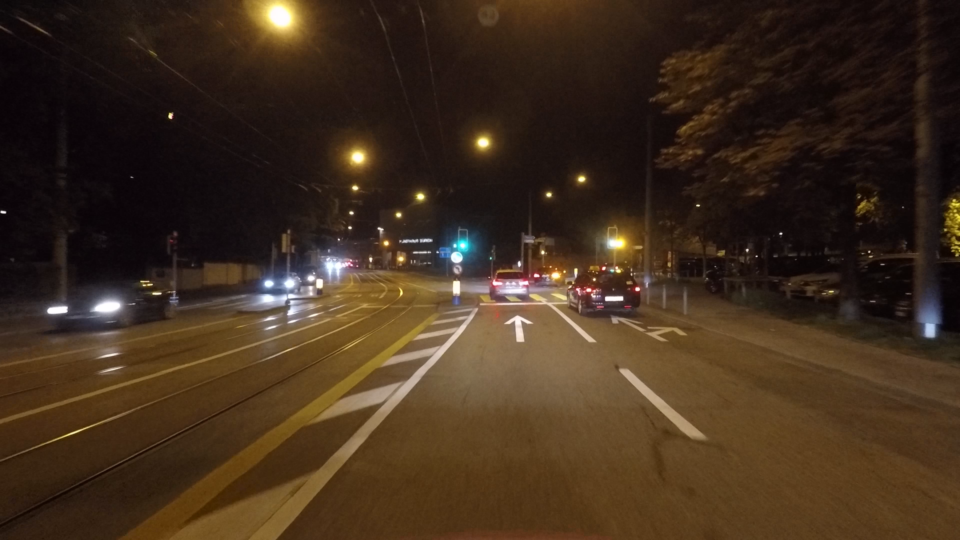}}
    \hfil
    \subfloat{\includegraphics[width=0.195\textwidth]{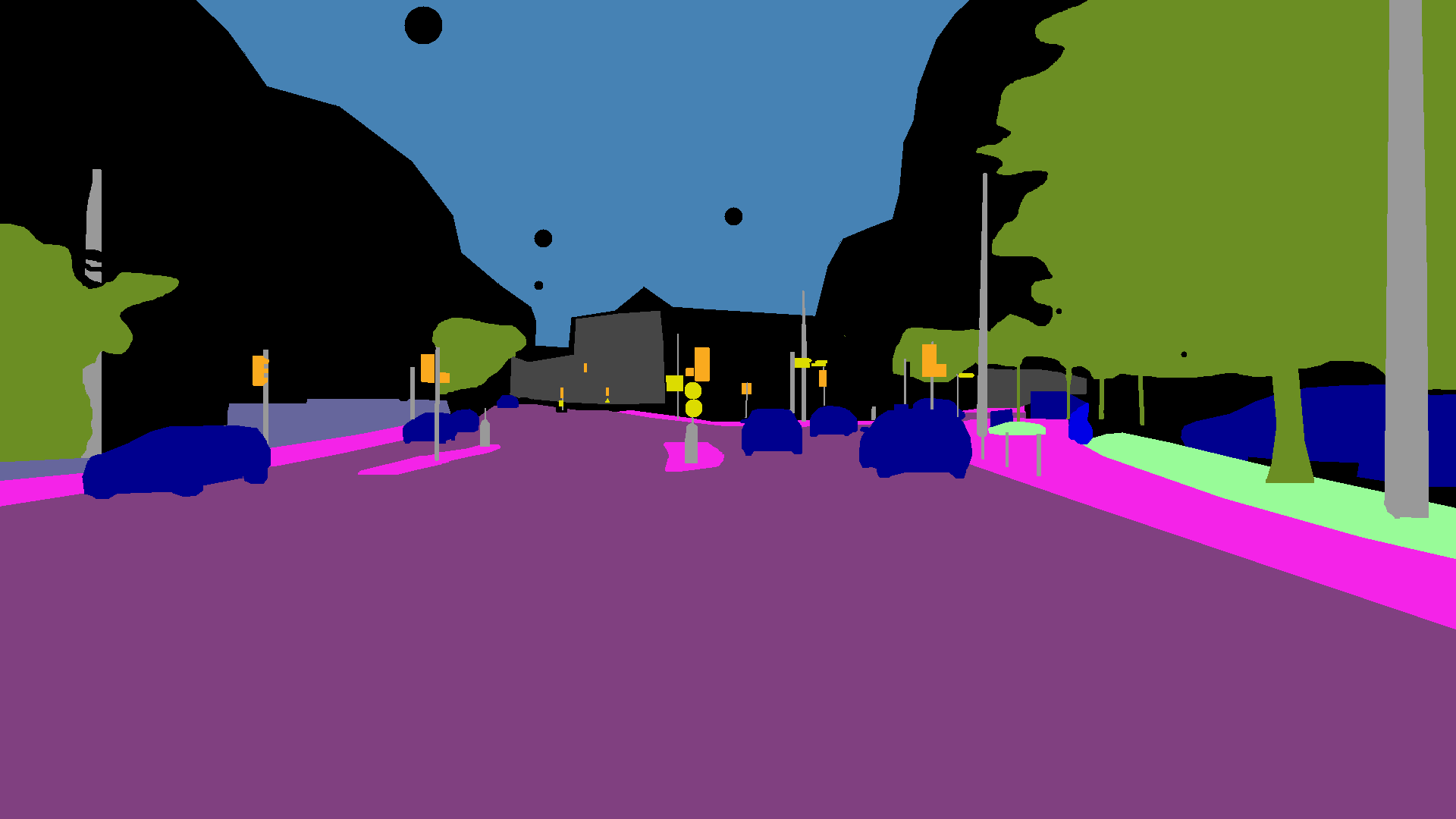}}
    \hfil
    \subfloat{\includegraphics[width=0.195\textwidth]{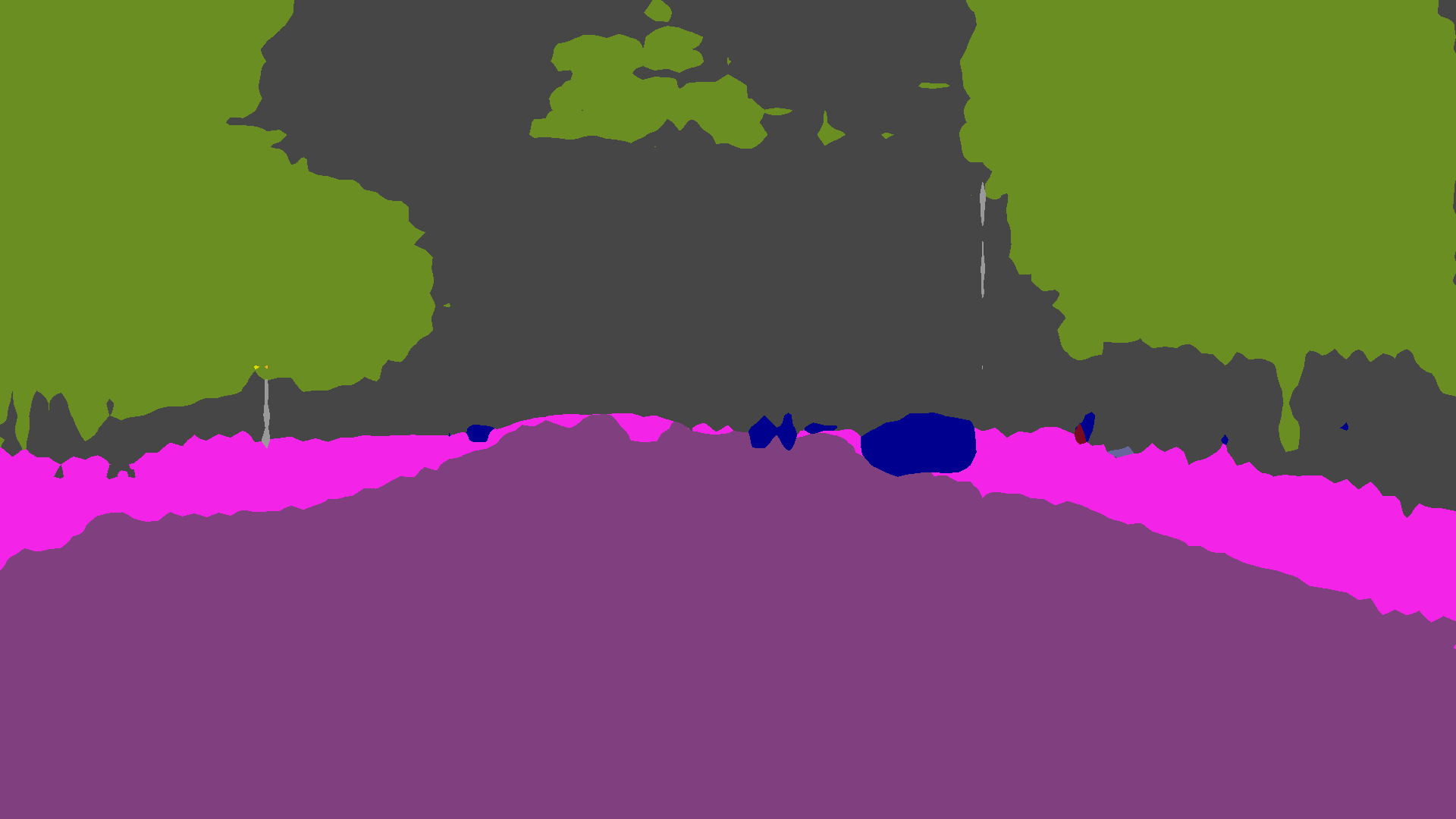}}
    \hfil
    \subfloat{\includegraphics[width=0.195\textwidth]{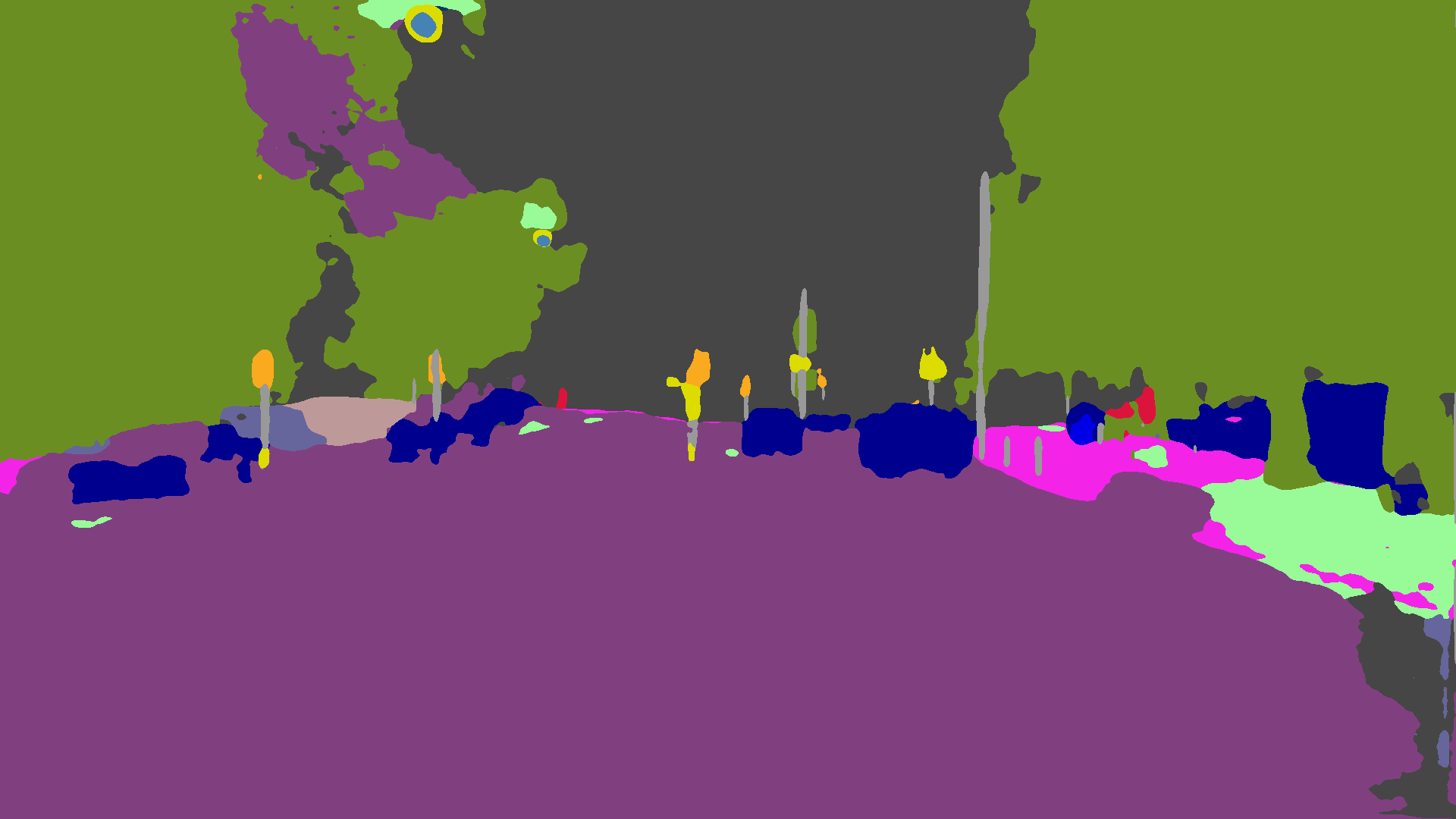}}
    \hfil
    \subfloat{\includegraphics[width=0.195\textwidth]{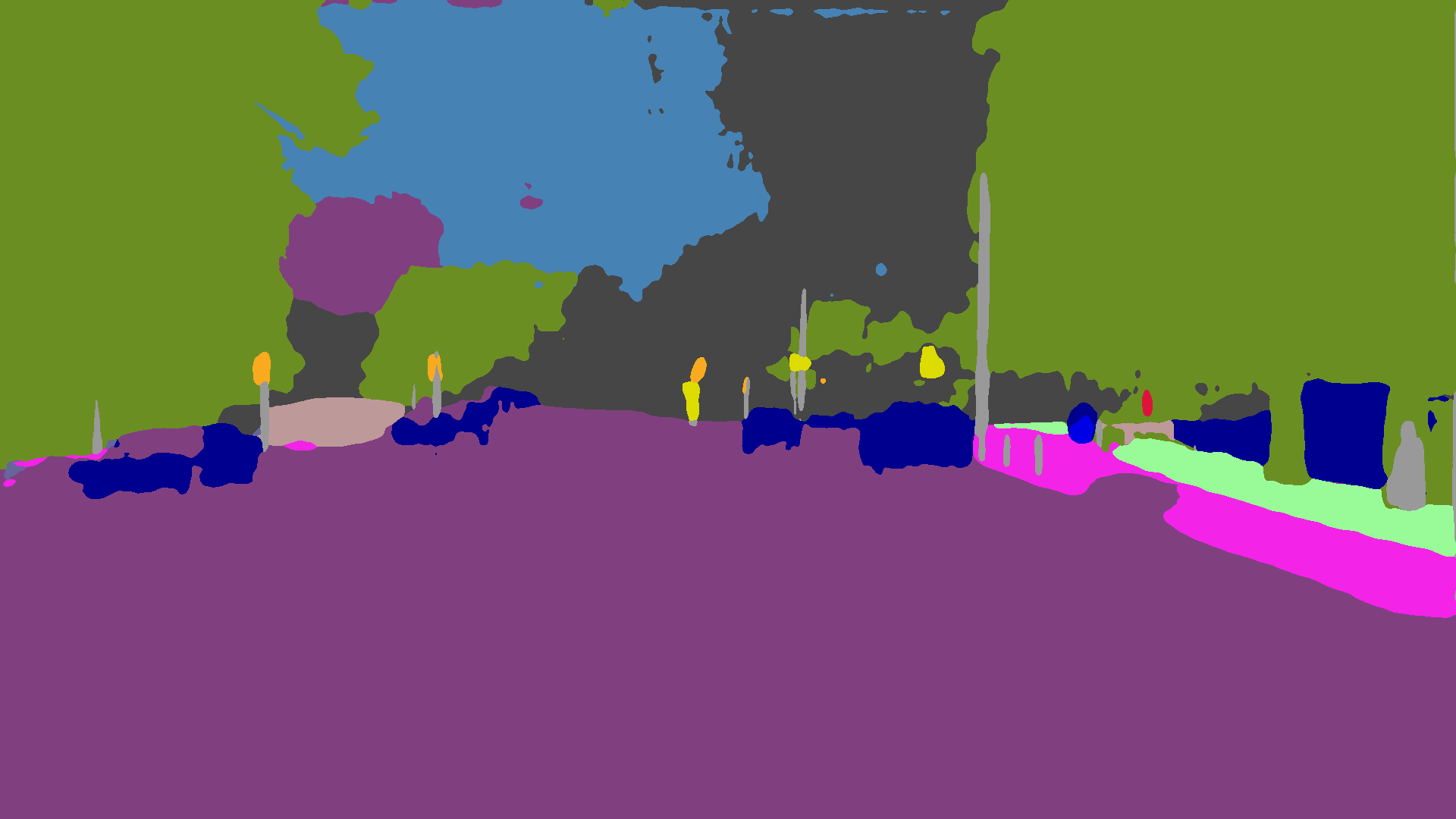}}
    \\
    \vspace{-0.3cm}
    \subfloat{\includegraphics[width=0.195\textwidth]{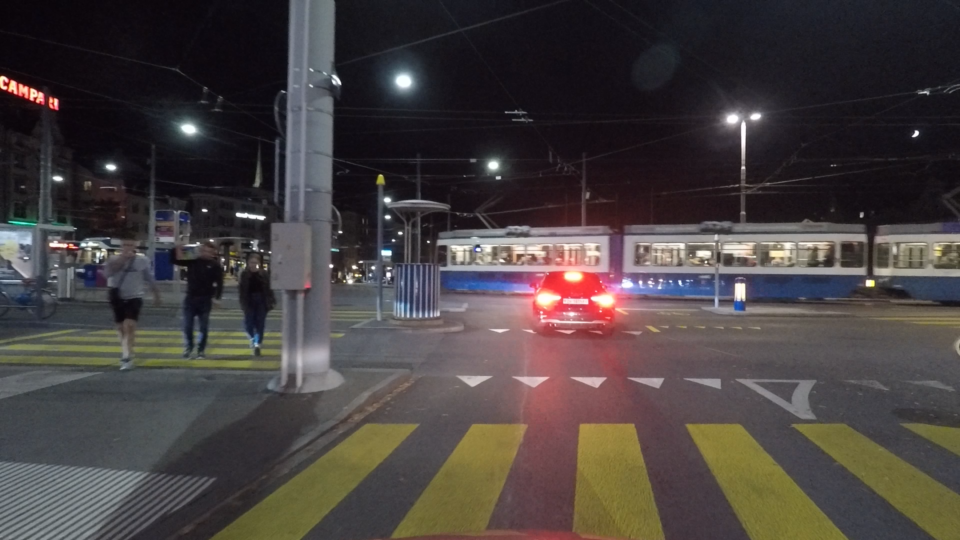}}
    \hfil
    \subfloat{\includegraphics[width=0.195\textwidth]{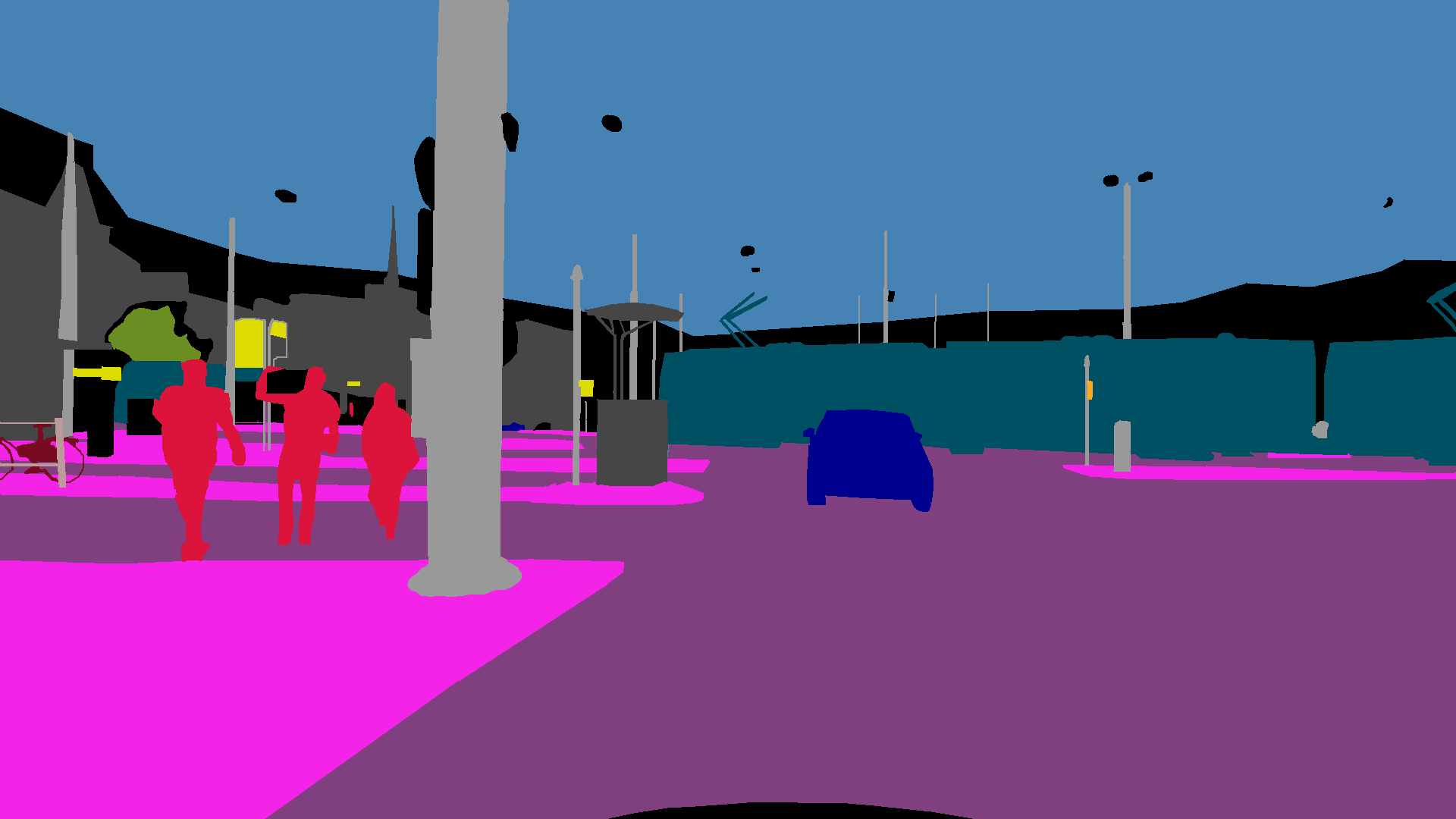}}
    \hfil
    \subfloat{\includegraphics[width=0.195\textwidth]{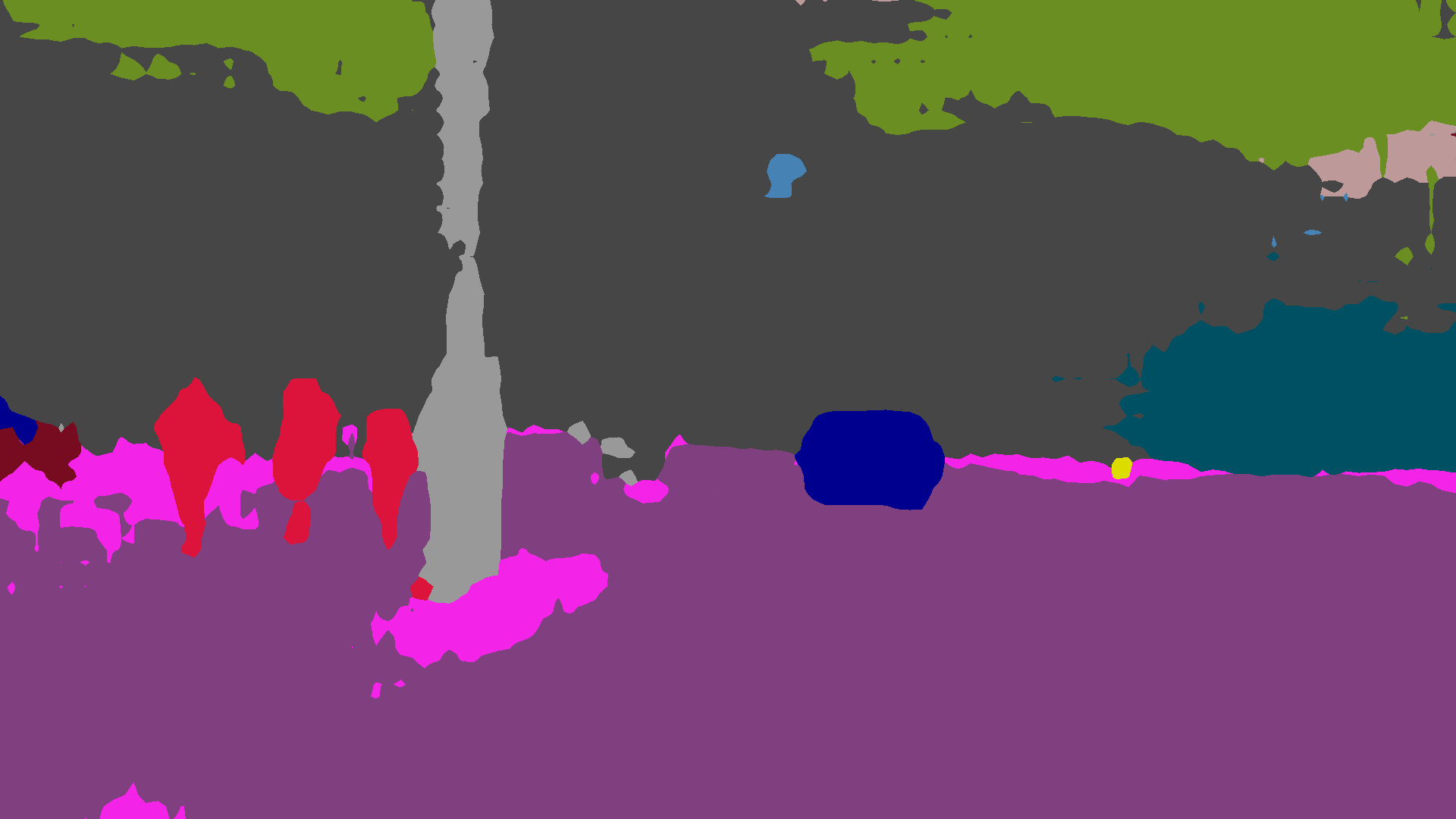}}
    \hfil
    \subfloat{\includegraphics[width=0.195\textwidth]{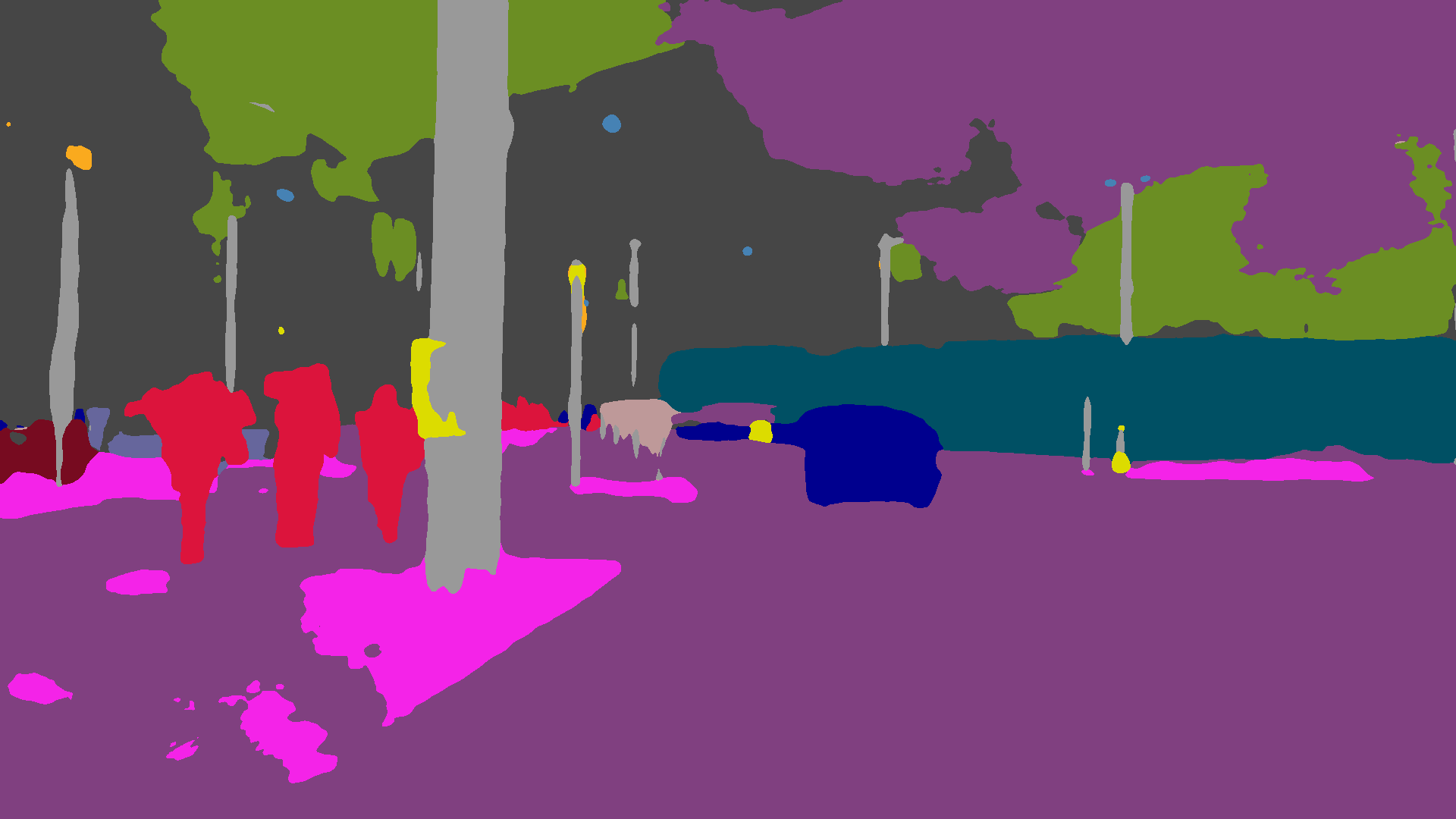}}
    \hfil
    \subfloat{\includegraphics[width=0.195\textwidth]{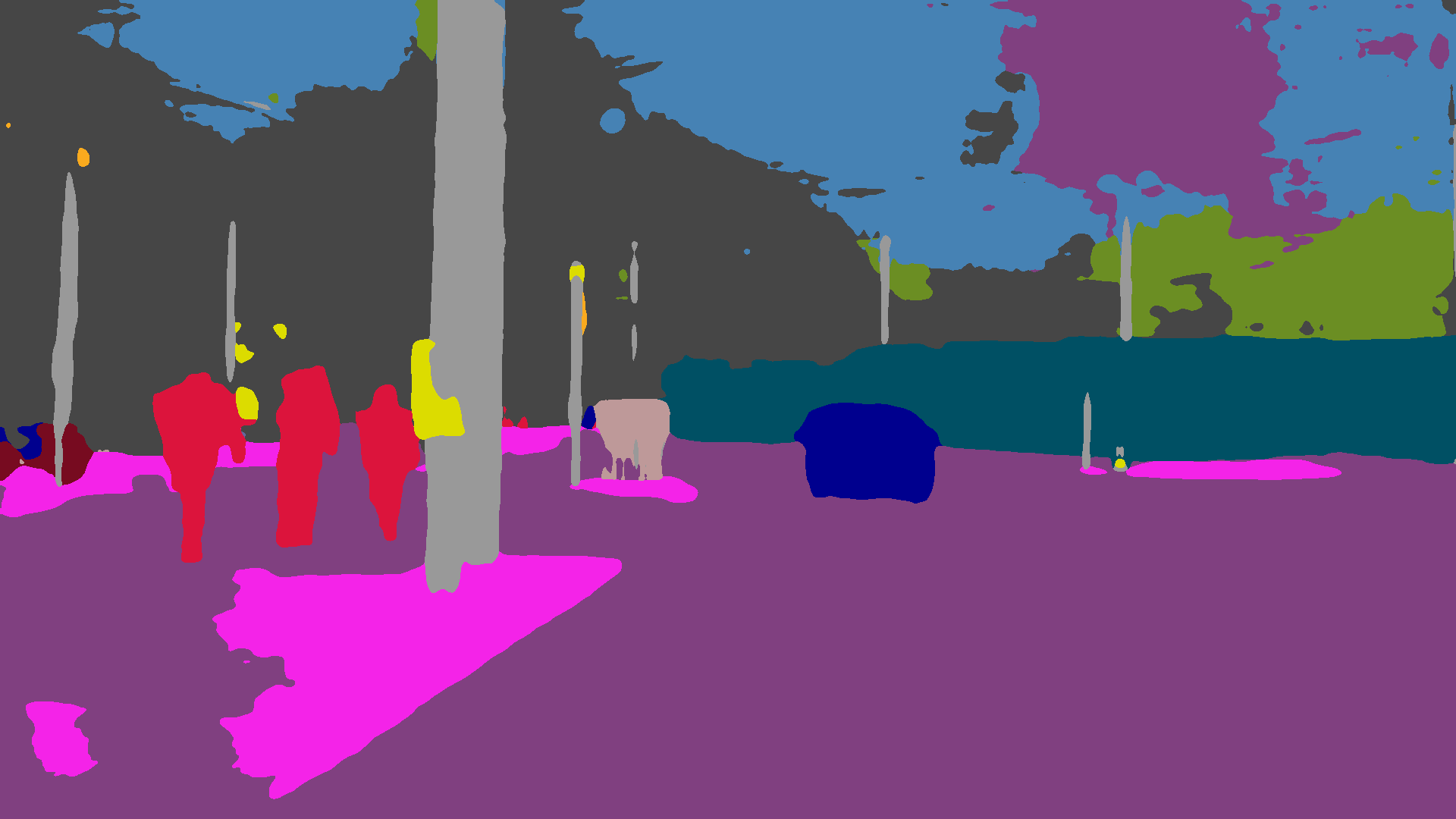}}
    \\
    \vspace{-0.3cm}
    \addtocounter{subfigure}{-10}
    \subfloat[Image]{\includegraphics[width=0.195\textwidth]{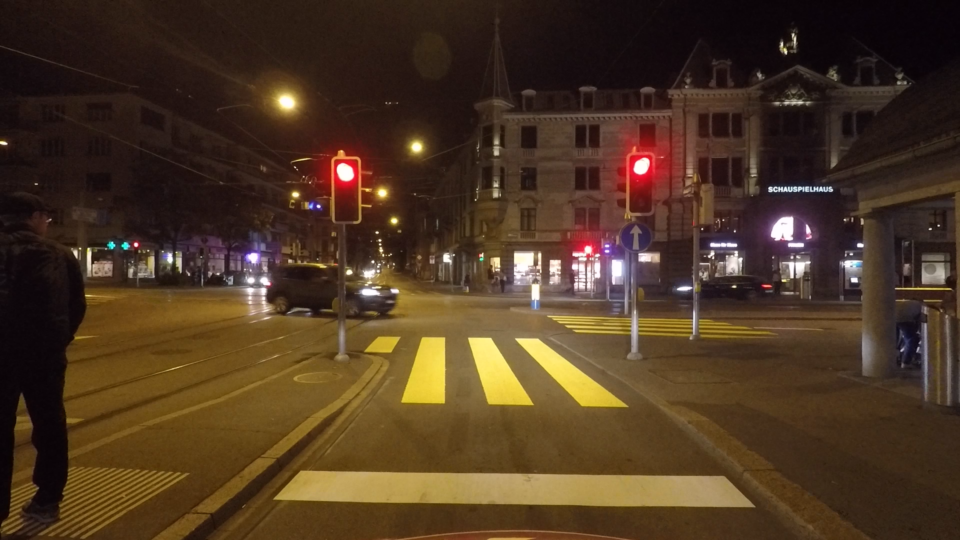}}
    \hfil
    \subfloat[Semantic GT]{\includegraphics[width=0.195\textwidth]{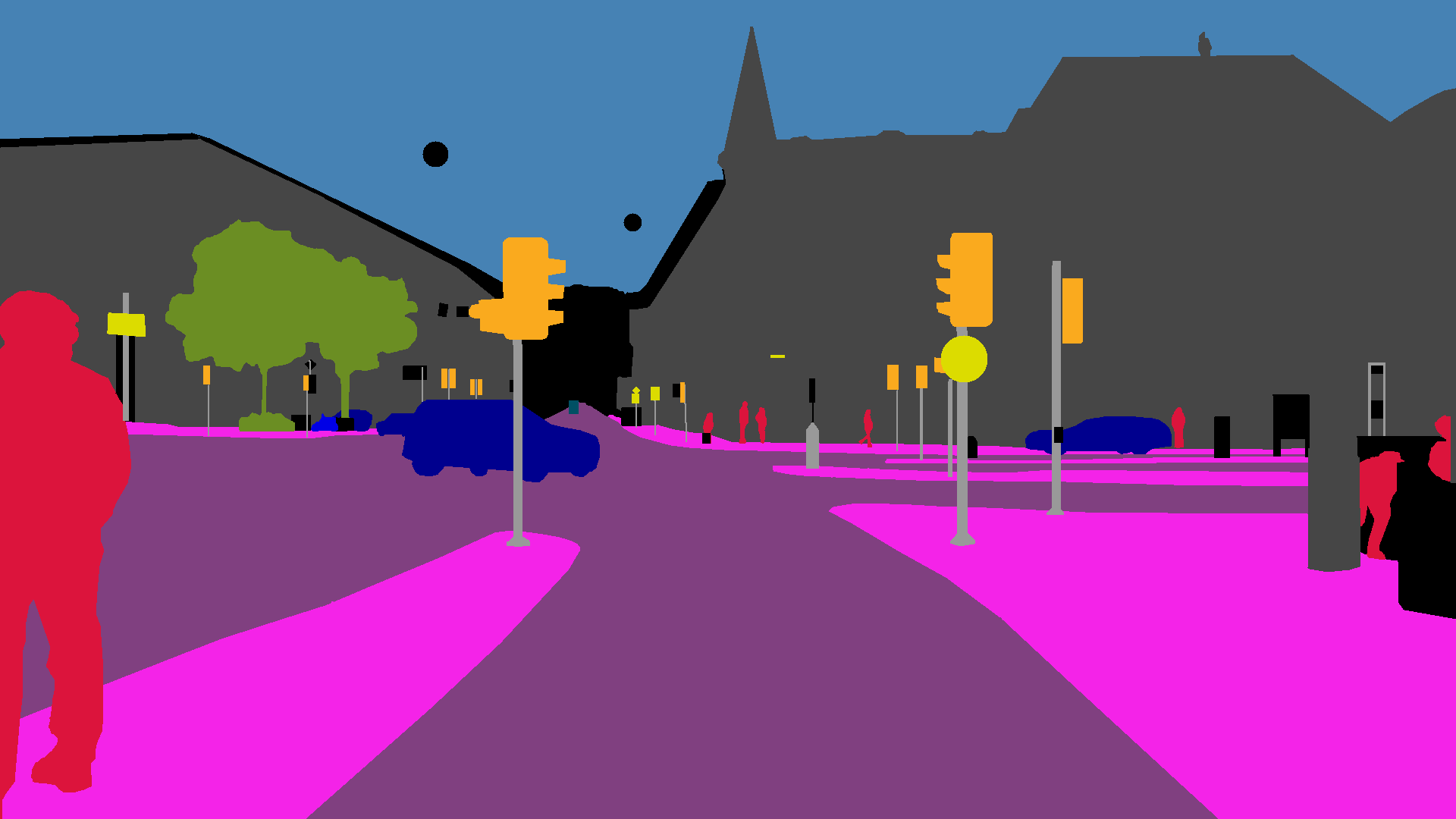}}
    \hfil
    \subfloat[AdaptSegNet~\cite{adapt:structured:output:cvpr18}]{\includegraphics[width=0.195\textwidth]{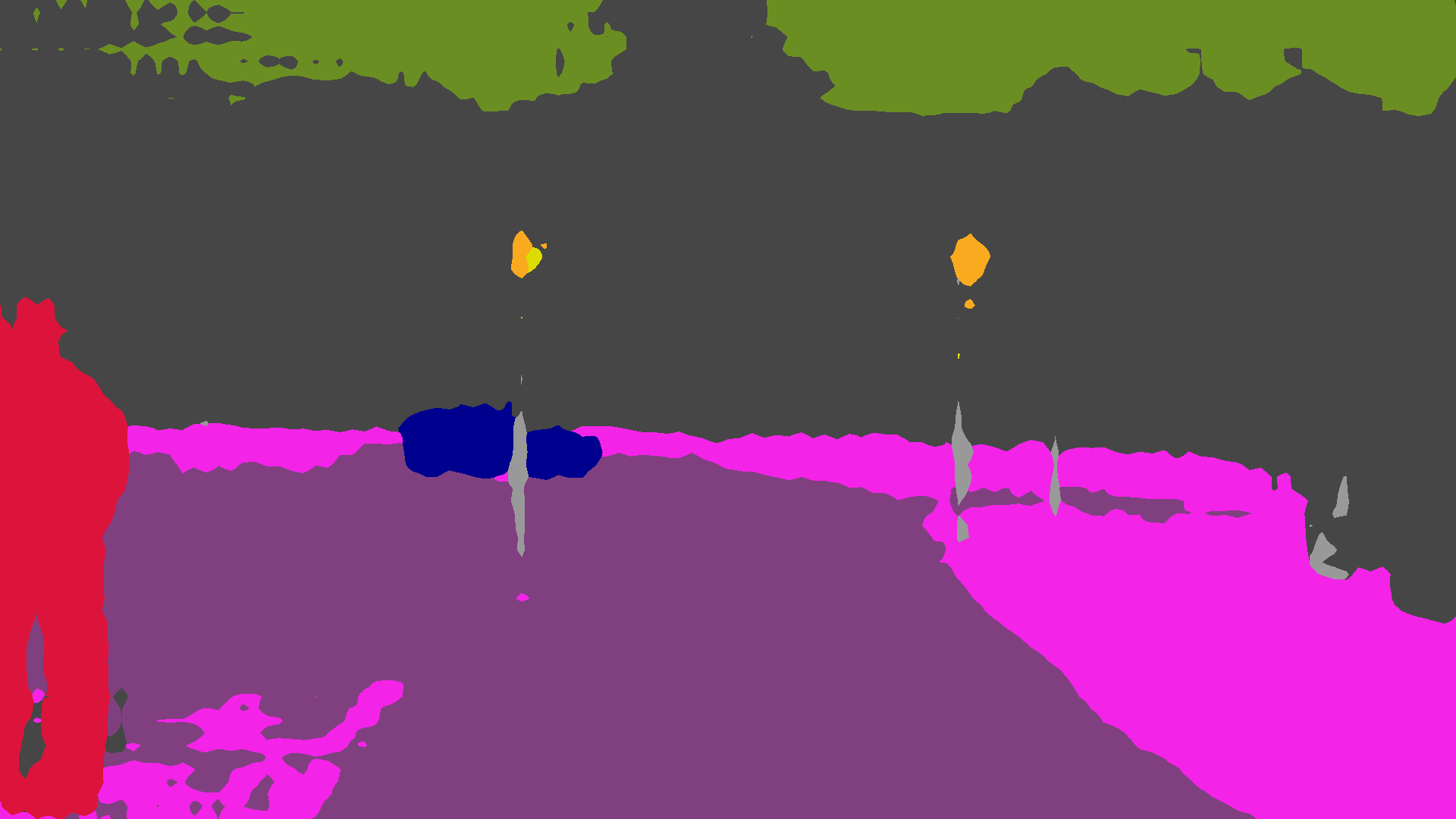}}
    \hfil
    \subfloat[DMAda~\cite{daytime:2:nighttime}]{\includegraphics[width=0.195\textwidth]{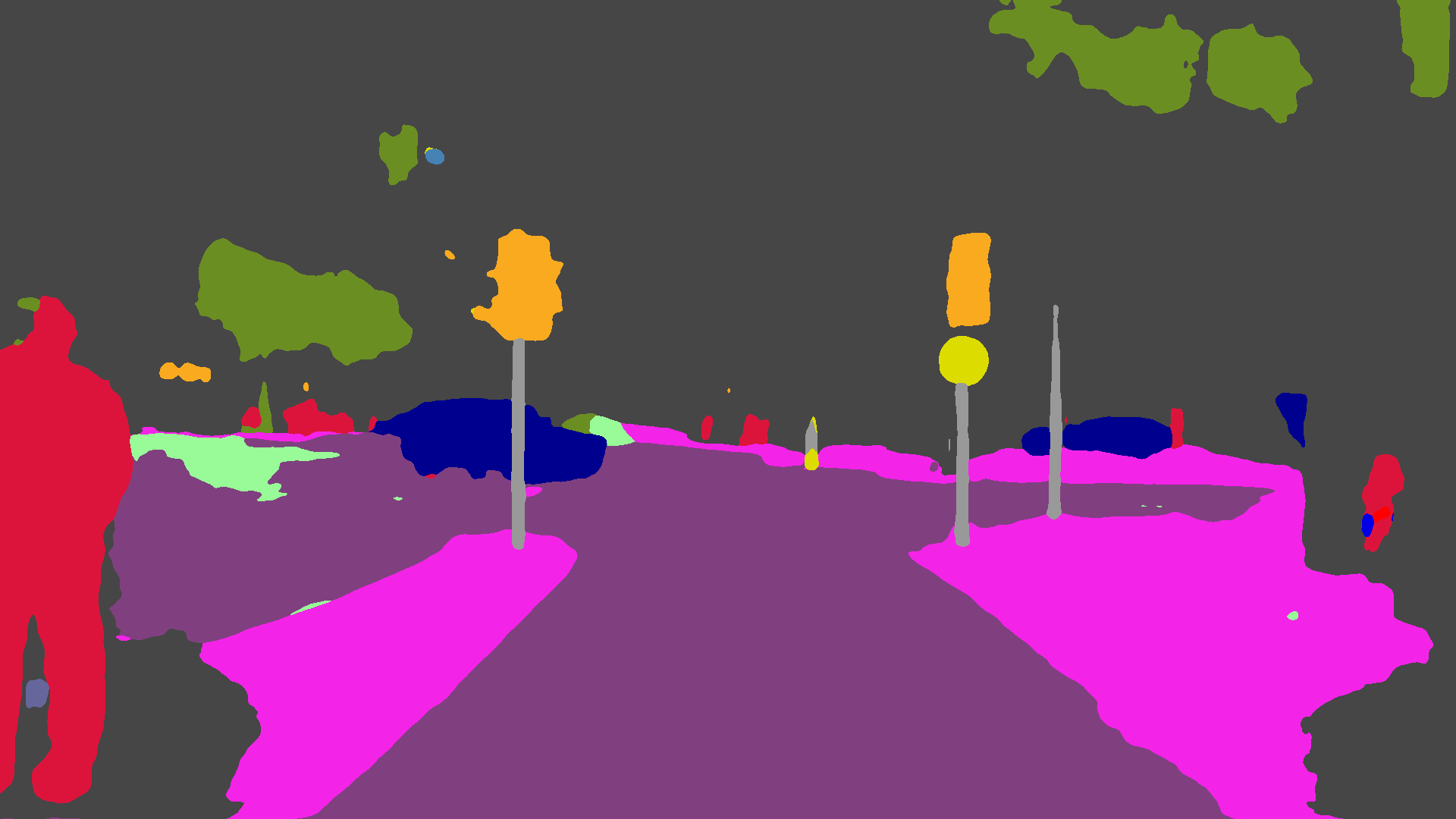}}
    \hfil
    \subfloat[GCMA (Ours)]{\includegraphics[width=0.195\textwidth]{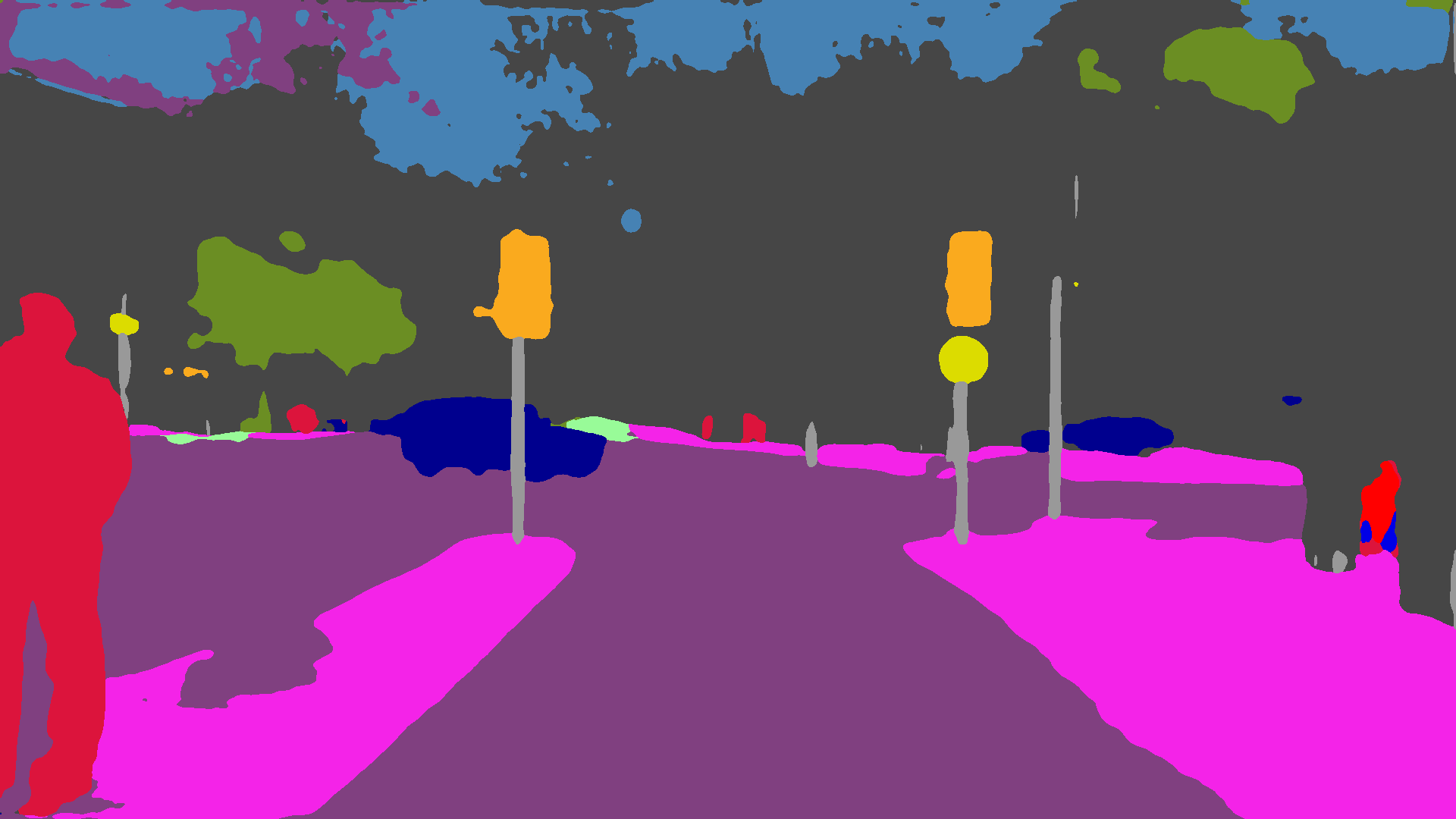}}
    \caption{Qualitative semantic segmentation results on \emph{Dark Zurich-test}. ``AdaptSegNet'' adapts from Cityscapes to \emph{Dark Zurich-night}.}
    \label{fig:sem:seg}
\end{figure*}

\begin{table*}[!tb]
  \caption{Comparison on \emph{Dark Zurich-test}. AdaptSegNet-Cityscapes$\to$\emph{DZ-night} denotes adaptation from Cityscapes to \emph{Dark Zurich-night}.}
  \label{table:exp:main:dark_zurich}
  \centering
  \setlength\tabcolsep{1.9pt}
  \footnotesize
  \begin{tabular}{lcccccccccccccccccccc}
  \toprule
  Method & \ver{road} & \ver{sidew.} & \ver{build.} & \ver{wall} & \ver{fence} & \ver{pole} & \ver{light} & \ver{sign} & \ver{veget.} & \ver{terrain} & \ver{sky} & \ver{person} & \ver{rider} & \ver{car} & \ver{truck} & \ver{bus} & \ver{train} & \ver{motorc.} & \ver{bicycle} & mIoU\\
  \midrule
  RefineNet~\cite{refinenet} & 68.8&23.2&46.8&20.8&12.6&29.8&30.4&26.9&43.1&14.3&0.3&36.9&49.7&63.6&6.8&\best{0.2}&24.0&33.6&9.3 & 28.5\\
  AdaptSegNet-Cityscapes~\cite{adapt:structured:output:cvpr18} & 79.0 & 21.8 & 53.0 & 13.3 & 11.2 & 22.5 & 20.2 & 22.1 & 43.5 & 10.4 & 18.0 & 37.4 & 33.8 & 64.1 & 6.4 & 0.0 & \best{52.3} & 30.4 & 7.4     & 28.8\\
  \midrule
  AdaptSegNet-Cityscapes$\to$\emph{DZ-night}~\cite{adapt:structured:output:cvpr18} & \best{86.1} & 44.2 & 55.1 & \best{22.2} & 4.8 & 21.1 & 5.6 & 16.7 & 37.2 & 8.4 & 1.2 & 35.9 & 26.7 & 68.2 & \best{45.1} & 0.0 & 50.1 & 33.9 & 15.6       & 30.4\\
  DMAda~\cite{daytime:2:nighttime} & 75.5&29.1&48.6&21.3&14.3&34.3&36.8&29.9&49.4&13.8&0.4&43.3&\best{50.2}&69.4&18.4&0.0&27.6&\best{34.9}&11.9             & 32.1\\
  Ours: GCMA & 81.7&\best{46.9}&\best{58.8}&22.0&\best{20.0}&\best{41.2}&\best{40.5}&\best{41.6}&\best{64.8}&\best{31.0}&\best{32.1}&\best{53.5}&47.5&\best{75.5}&39.2&0.0&49.6&30.7&\best{21.0}             & \best{42.0}\\
  \bottomrule
  \\
  \end{tabular}
\end{table*}

\begin{table}[!tb]
  \caption{Comparison on Nighttime Driving~\cite{daytime:2:nighttime}. Read as Table~\ref{table:exp:main:dark_zurich}.}
  \label{table:exp:main:nighttime_driving}
  \centering
  \setlength\tabcolsep{4pt}
  \footnotesize
  \begin{tabular}{lc}
  \toprule
  Method & mIoU (\%)\\
  \midrule
  RefineNet~\cite{refinenet} & 31.5\\
  AdaptSegNet-Cityscapes~\cite{adapt:structured:output:cvpr18} & 32.6\\
  \midrule
  AdaptSegNet-Cityscapes$\to$\emph{DZ-night}~\cite{adapt:structured:output:cvpr18} & 34.5\\
  DMAda~\cite{daytime:2:nighttime} & 36.1\\
  Ours: GCMA & \best{45.6}\\
  \bottomrule
  \\
  \end{tabular}
\end{table}

Our architecture of choice for implementing GCMA is RefineNet~\cite{refinenet}. We use the publicly available \emph{RefineNet-res101-Cityscapes} model, trained on Cityscapes, as the baseline model to be adapted to nighttime. Throughout our experiments, we train this model with a constant learning rate of $5 \times 10^{-5}$ on mini-batches of size 1.

\PAR{Comparison to Other Adaptation Methods.}
Our first experiment compares GCMA to state-of-the-art approaches for adaptation of semantic segmentation models to nighttime. To obtain the synthetic labeled datasets for GCMA, we stylize Cityscapes to twilight using a CycleGAN model that is trained to translate Cityscapes to \emph{Dark Zurich-twilight} (respectively to nighttime with \emph{Dark Zurich-night}). The real training datasets for GCMA are \emph{Dark Zurich-day}, instantiating $\mathcal{D}^1_{ur}$, and \emph{Dark Zurich-twilight}, instantiating $\mathcal{D}^2_{ur}$. Each adaptation step comprises 30k SGD iterations and uses $\mu = 1$. For the second step, we apply our guided refinement to the labels of \emph{Dark Zurich-twilight} that are predicted by model $\phi^2$ fine-tuned in the first step, using the correspondences of \emph{Dark Zurich-twilight} to \emph{Dark Zurich-day}.

We evaluate GCMA on \emph{Dark Zurich-test} against the state-of-the-art adaptation approaches AdaptSegNet~\cite{adapt:structured:output:cvpr18} and DMAda~\cite{daytime:2:nighttime} and report standard IoU performance in Table~\ref{table:exp:main:dark_zurich}, including \emph{invalid} pixels which are assigned a legitimate semantic label in the evaluation. We have trained AdaptSegNet to adapt from Cityscapes to \emph{Dark Zurich-night}. For fair comparison, we also report the performance of the respective baseline Cityscapes models for each method. RefineNet is the common baseline of GCMA and DMAda. GCMA significantly outperforms the other methods for most classes and achieves a substantial 10\% improvement in the overall mIoU score against the second-best method. The improvement with GCMA is pronounced for classes which usually appear dark at nighttime, such as \emph{sky}, \emph{vegetation}, \emph{terrain} and \emph{person}, indicating that our method successfully handles large domain shifts from its source daytime domain. These findings are supported by visually assessing the predictions of the compared methods, as in the examples of Fig.~\ref{fig:sem:seg}. We repeat the above comparison on Nighttime Driving~\cite{daytime:2:nighttime} in Table~\ref{table:exp:main:nighttime_driving} and show that GCMA generalizes very well to different datasets.

\begin{table}[!tb]
  \caption{Ablations of GCMA on \emph{Dark Zurich-test}, reporting mIoU.}
  \label{table:exp:ablation:dark_zurich}
  \centering
  \setlength\tabcolsep{4pt}
  \footnotesize
  \begin{tabular}{lc}
  \toprule
  Daytime baseline: RefineNet~\cite{refinenet} & 28.5\%\\
  +direct CycleGAN adapt.\ (w/o real, w/o curriculum) & 37.1\%\\
  +GCMA w/o guided refinement & 39.4\%\\
  +GCMA w/ guided refinement & \best{42.0\%}\\
  \bottomrule
  \\
  \end{tabular}
\end{table}

\PAR{Ablation Study for GCMA.}
We measure the individual effect of the main components of GCMA in Table~\ref{table:exp:ablation:dark_zurich} by evaluating its ablated versions on \emph{Dark Zurich-test}. Direct adaptation to nighttime in a single step using only Cityscapes images stylized as nighttime with CycleGAN is a strong baseline, due to the reliable ground-truth labels that accompany the stylized Cityscapes, its high diversity and the limited artifacts of CycleGAN-based translation. Adding our real images to the training algorithm and applying our two-stage curriculum significantly improves upon this baseline. Finally, our guided segmentation refinement in the second step of GCMA brings an additional 2.6\% benefit, as it corrects a lot of errors in the pseudo-labels of the real twilight images, which helps compute more reliable gradients from the corrected loss during the subsequent training.

\begin{figure}[!tb]
    \centering
    \includegraphics[clip,width=\linewidth,trim=25mm 100mm 25mm 100mm]{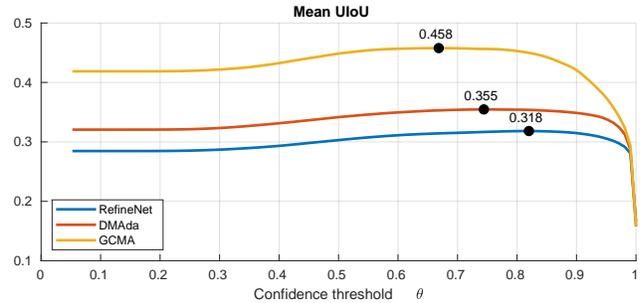}
    \caption{Uncertainty-aware evaluation of RefineNet~\cite{refinenet}, DMAda~\cite{daytime:2:nighttime} and GCMA on \emph{Dark Zurich-test}. We evaluate mean UIoU across the entire range $[1/C,\,1]$ of confidence threshold $\theta$. For each method, the point at which mean UIoU is maximized is marked black and labeled with this maximum mean UIoU value.}
    \label{fig:exp:uiou}
\end{figure}

\PAR{Comparisons with UIoU.}
In Fig.~\ref{fig:exp:uiou}, we use our novel UIoU metric to evaluate GCMA against DMAda and our baseline RefineNet model on \emph{Dark Zurich-test} for varying confidence threshold $\theta$ and plot the resulting mean UIoU$(\theta)$ curves. Note that standard mean IoU can be read out from the leftmost point of each curve. First, our expectation based on Th.~\ref{thm:UIoU:greater:iou} is confirmed for all methods, i.e.\ maximum UIoU values over the range of $\theta$ are larger than IoU by ca.\ 3\%. This implies that on \emph{Dark Zurich-test}, these models generally have lower confidence on invalid regions than valid ones. Second, the comparative performance of the methods is the same across all values of $\theta$ ---GCMA substantially outperforms the other two---which shows that UIoU is generally consistent with standard IoU and is a suitable substitute of the latter in adverse settings where declaring the input as invalid is relevant.

\section{Conclusion}
\label{sec:conclusion}

In this paper, we have introduced GCMA, a method to gradually adapt semantic segmentation models from daytime to nighttime with stylized data and unlabeled real data of increasing darkness, as well as UIoU, a novel evaluation metric for semantic segmentation designed for images with indiscernible content. We have also presented \emph{Dark Zurich}, a large-scale dataset of real scenes captured at multiple times of day with cross-time-of-day correspondences, and annotated 151 nighttime scenes of it with a new protocol which enables our evaluation. Detailed evaluation with standard IoU on real nighttime sets demonstrates the merit of GCMA, which substantially improves upon competing state-of-the-art methods. Finally, evaluation on our benchmark with UIoU shows that invalidating predictions is useful when the input includes ambiguous content.

\PAR{Acknowledgements.}
This work is funded by Toyota Motor Europe via the research project TRACE-Z\"urich. We thank Simon Hecker for his advice on decoding GoPro GPS data.

{\small
\bibliographystyle{ieee}
\bibliography{refs}
}

\appendix
\clearpage
\pagenumbering{roman}

\section{Proof of Theorem 1}
\label{supp:sec:proof}

\begin{proof}
For brevity in the proof, we drop the class superscript $(c)$ which is used in the statement of the theorem.

Firstly, we draw an association between pixel sets related to the standard $\text{IoU} = \text{UIoU}(1/C)$ and their counterparts for UIoU defined in \eqref{eq:tp}--\eqref{eq:fi}. In particular, the following holds true:
\begin{multline} \label{eq:proof:true:positive:false:negative:iou}
|\text{TP}(1/C)| + |\text{FN}(1/C)| \\
= |\text{TP}(\theta)| + |\text{FN}(\theta)| + |\text{TI}(\theta)| + |\text{FI}(\theta)|,\,\forall \theta \in [1/C,\,1].
\end{multline}
The first assumption of Th.~\ref{thm:UIoU:greater:iou} implies that $\text{FI}(\theta_1) = \emptyset$, because $\forall \theta < \theta_2$ (including $\theta_1$) there exists no false invalid pixel for the examined class. Thus, applying \eqref{eq:proof:true:positive:false:negative:iou} for $\theta = \theta_1$ leads to
\begin{equation} \label{eq:proof:true:positive:iou:theta1}
|\text{TP}(1/C)| \\
= |\text{TP}(\theta_1)| + |\text{TI}(\theta_1)| + |\text{FN}(\theta_1)| - |\text{FN}(1/C)|.
\end{equation}

Secondly, we plug the proposition of the first assumption of the theorem into the proposition of the second assumption to obtain
\begin{equation} \label{eq:proof:nonempty:improved:init}
\left(\text{FN}(1/C) \cup \text{FP}(1/C)\right) \setminus \left(\text{FN}(\theta_1) \cup \text{FP}(\theta_1)\right) \neq \emptyset.
\end{equation}
We further elaborate on \eqref{eq:proof:nonempty:improved:init} by observing that $\text{FN}(1/C) \cap \text{FP}(1/C) = \emptyset$, $\text{FN}(\theta_1) \subseteq \text{FN}(1/C)$ and $\text{FP}(\theta_1) \subseteq \text{FP}(1/C)$ to arrive at
\begin{equation} \label{eq:proof:nonempty:improved}
(|\text{FN}(1/C)| - |\text{FN}(\theta_1)|) + (|\text{FP}(1/C)| - |\text{FP}(\theta_1)|) > 0.
\end{equation}

Both terms on the left-hand side of \eqref{eq:proof:nonempty:improved} are nonnegative based on our previous observations, while at the same time \eqref{eq:proof:nonempty:improved} implies that at least one of the two is strictly positive. To complete the proof, we distinguish between the two corresponding cases.

\begin{figure*}[!tb]
    \centering
    \subfloat{\includegraphics[width=0.195\textwidth]{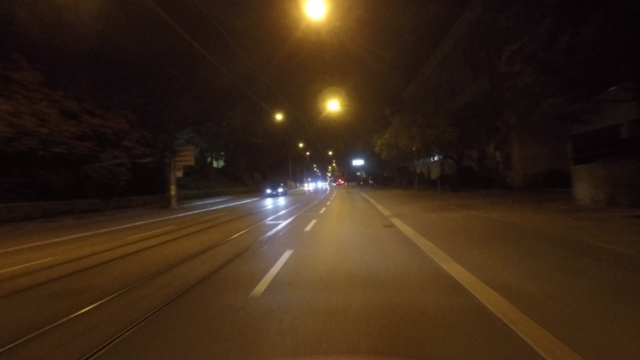}}
    \hfil
    \subfloat{\includegraphics[width=0.195\textwidth]{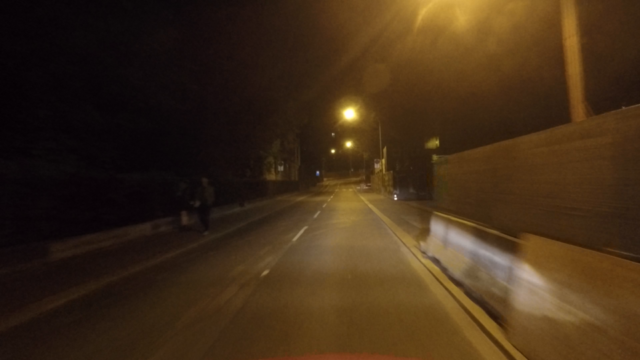}}
    \hfil
    \subfloat{\includegraphics[width=0.195\textwidth]{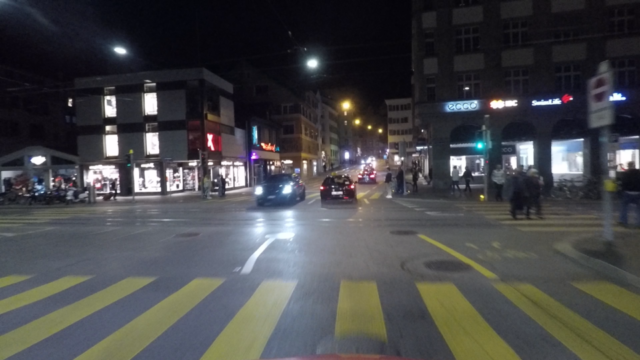}}
    \hfil
    \subfloat{\includegraphics[width=0.195\textwidth]{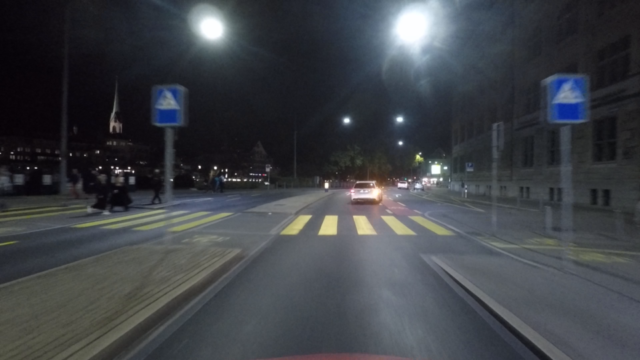}}
    \hfil
    \subfloat{\includegraphics[width=0.195\textwidth]{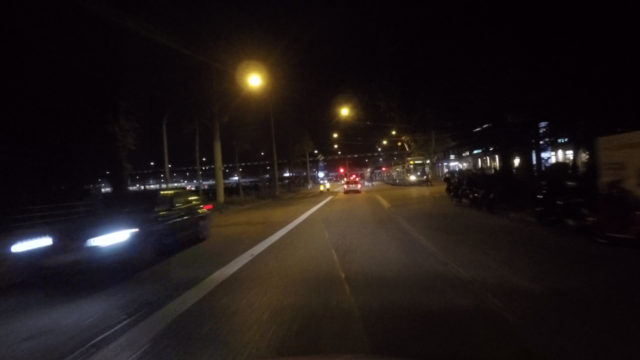}}
    \\
    \vspace{-0.3cm}
    \subfloat{\includegraphics[width=0.195\textwidth]{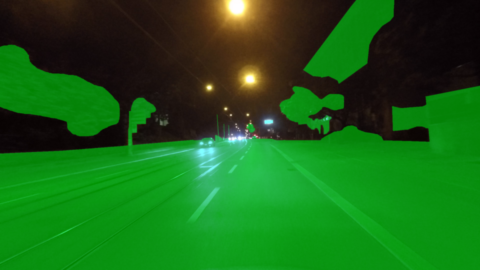}}
    \hfil
    \subfloat{\includegraphics[width=0.195\textwidth]{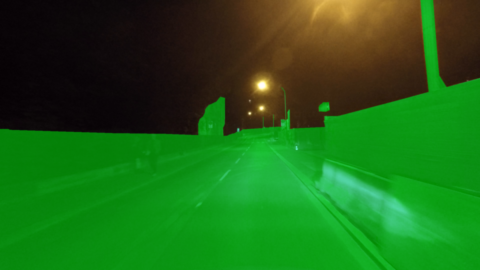}}
    \hfil
    \subfloat{\includegraphics[width=0.195\textwidth]{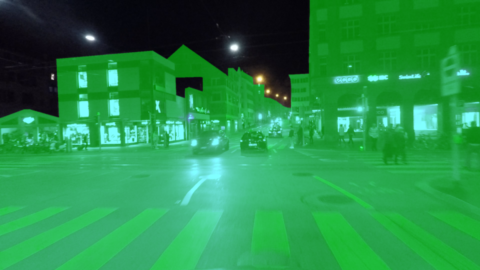}}
    \hfil
    \subfloat{\includegraphics[width=0.195\textwidth]{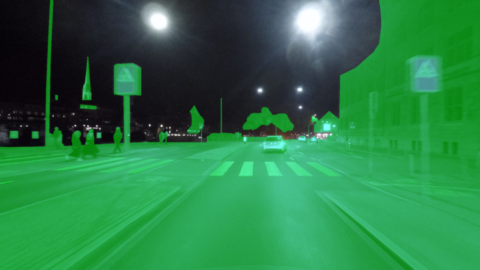}}
    \hfil
    \subfloat{\includegraphics[width=0.195\textwidth]{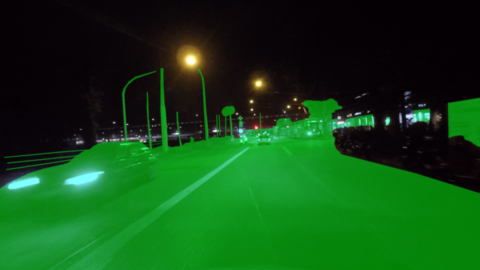}}
    \\
    \vspace{-0.3cm}
    \subfloat{\includegraphics[width=0.195\textwidth]{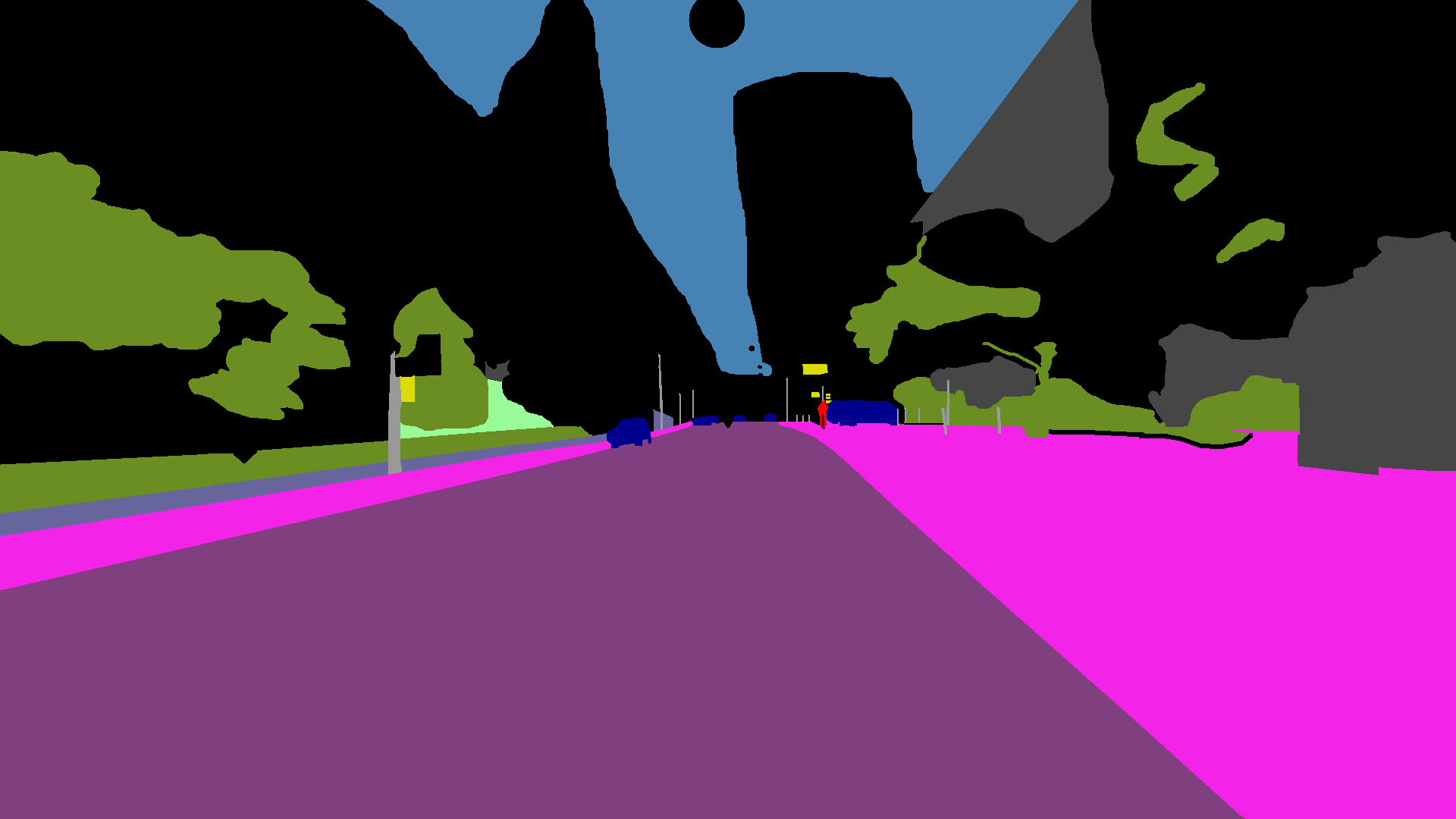}}
    \hfil
    \subfloat{\includegraphics[width=0.195\textwidth]{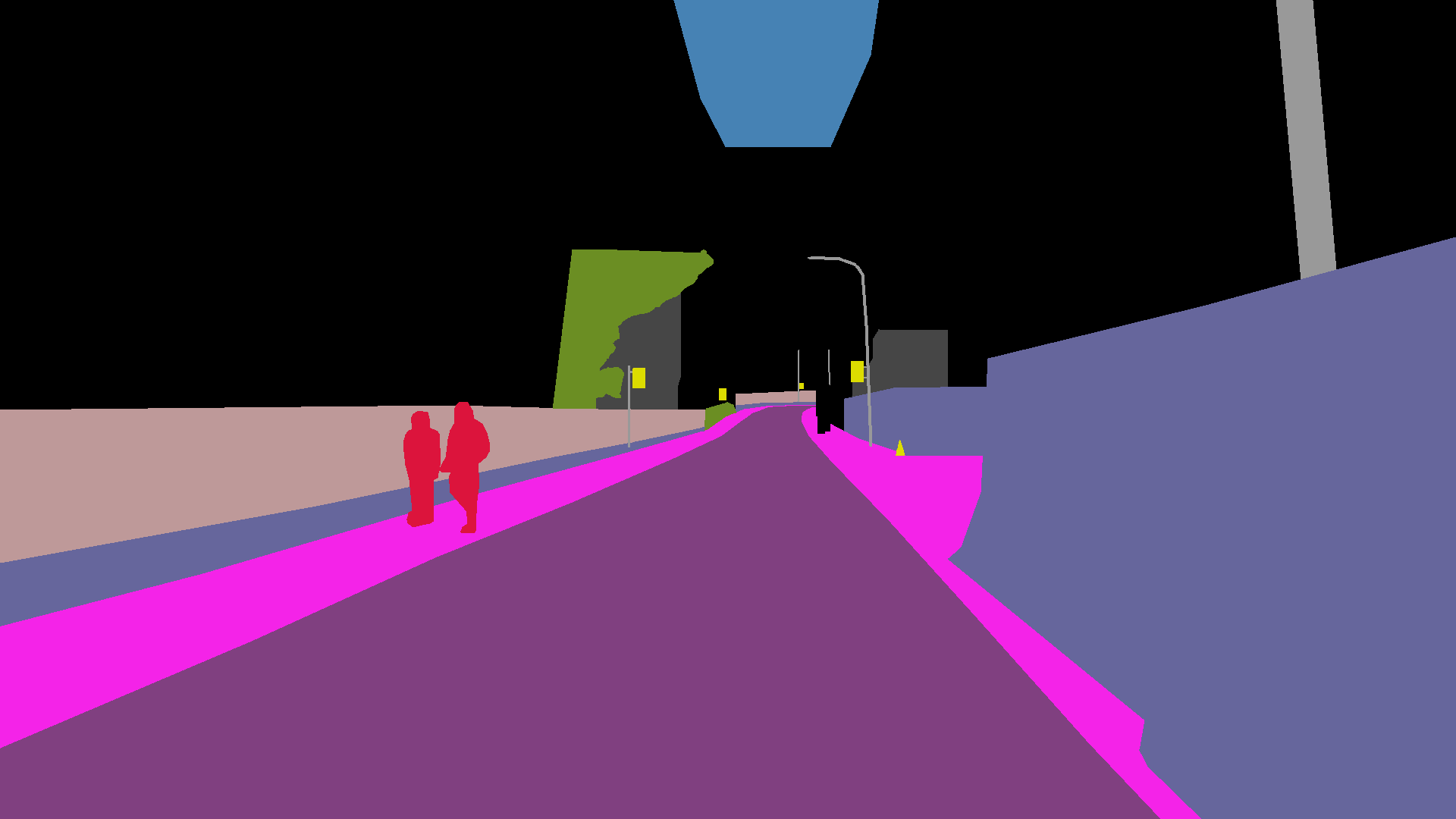}}
    \hfil
    \subfloat{\includegraphics[width=0.195\textwidth]{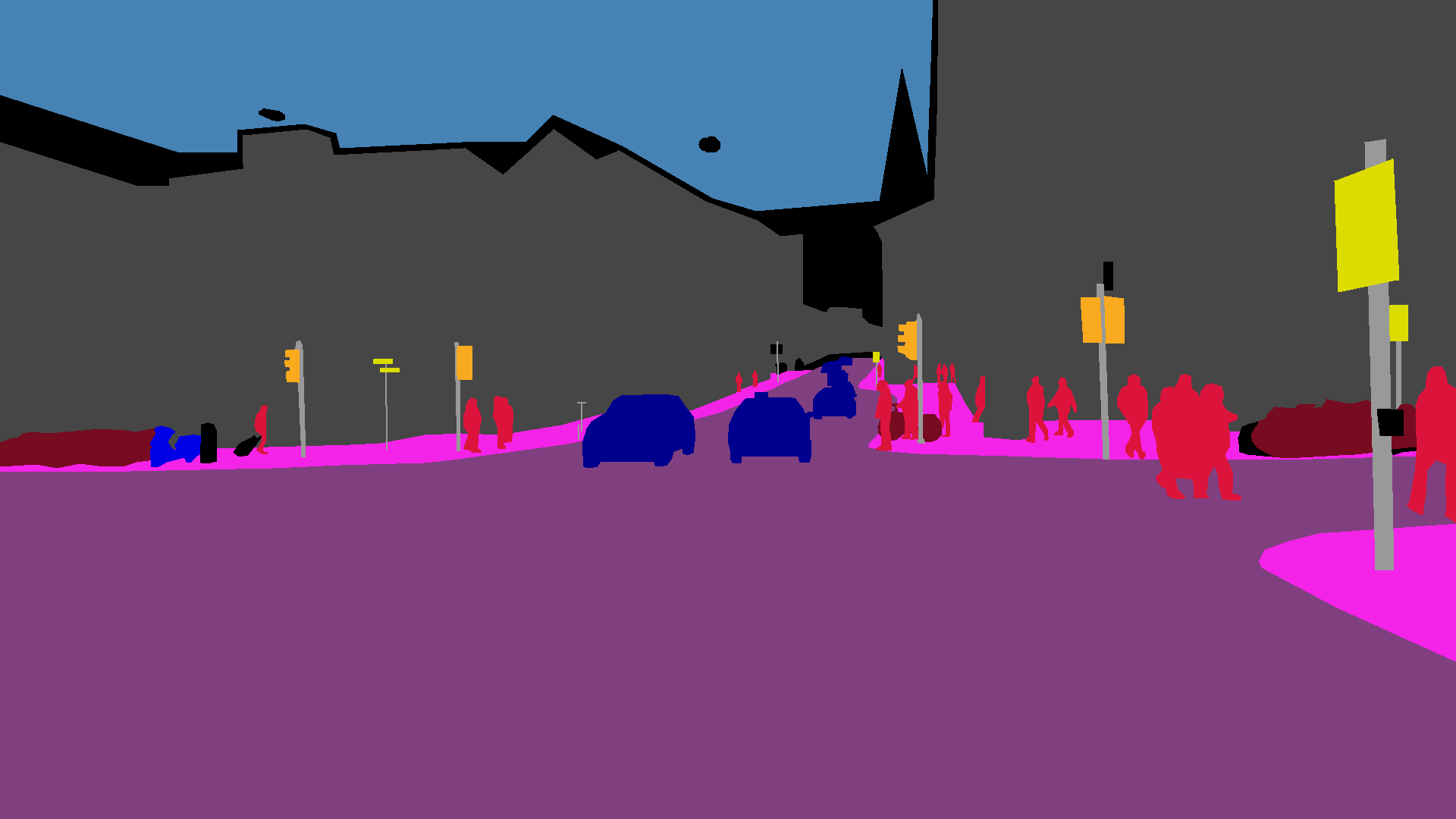}}
    \hfil
    \subfloat{\includegraphics[width=0.195\textwidth]{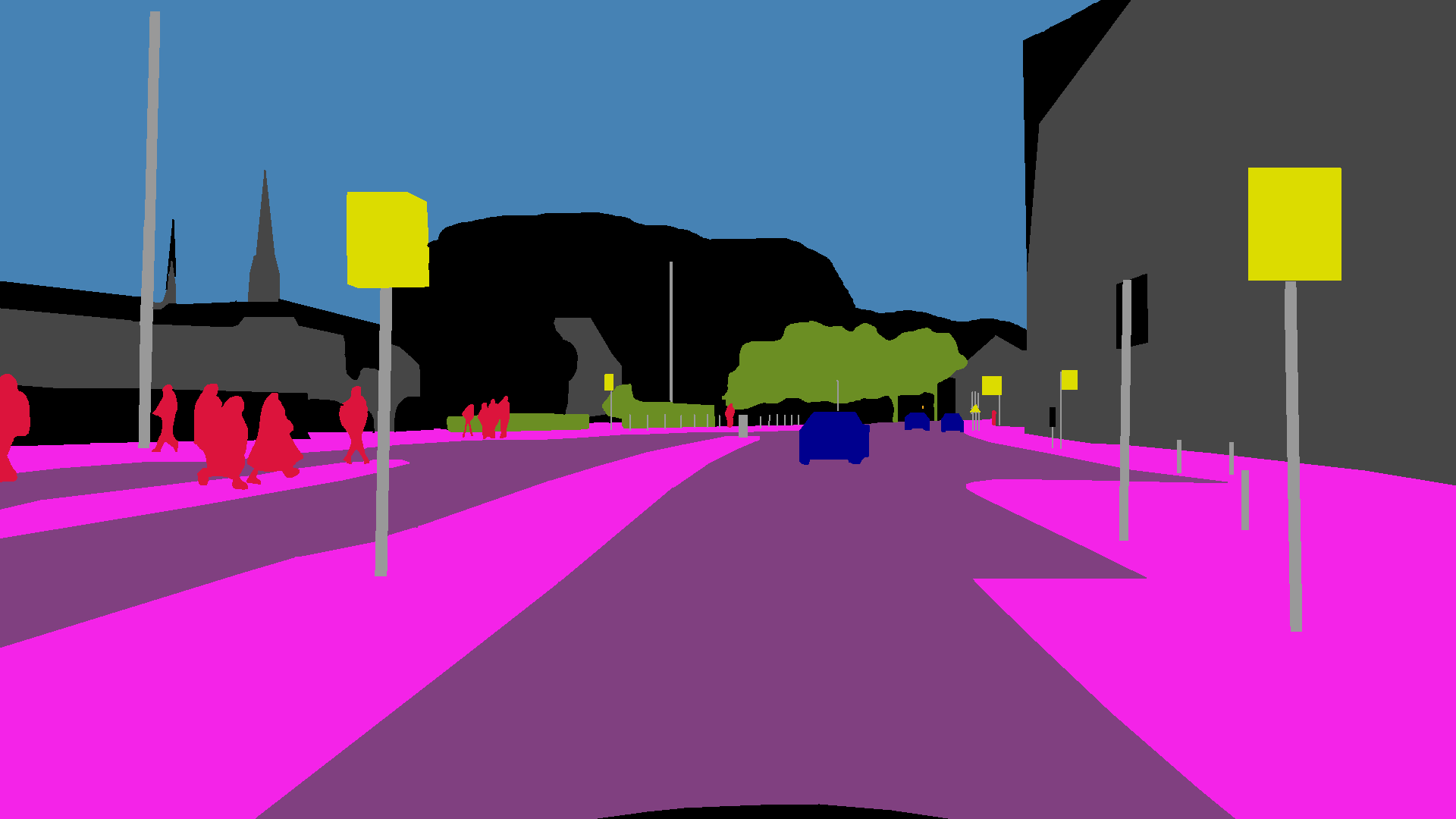}}
    \hfil
    \subfloat{\includegraphics[width=0.195\textwidth]{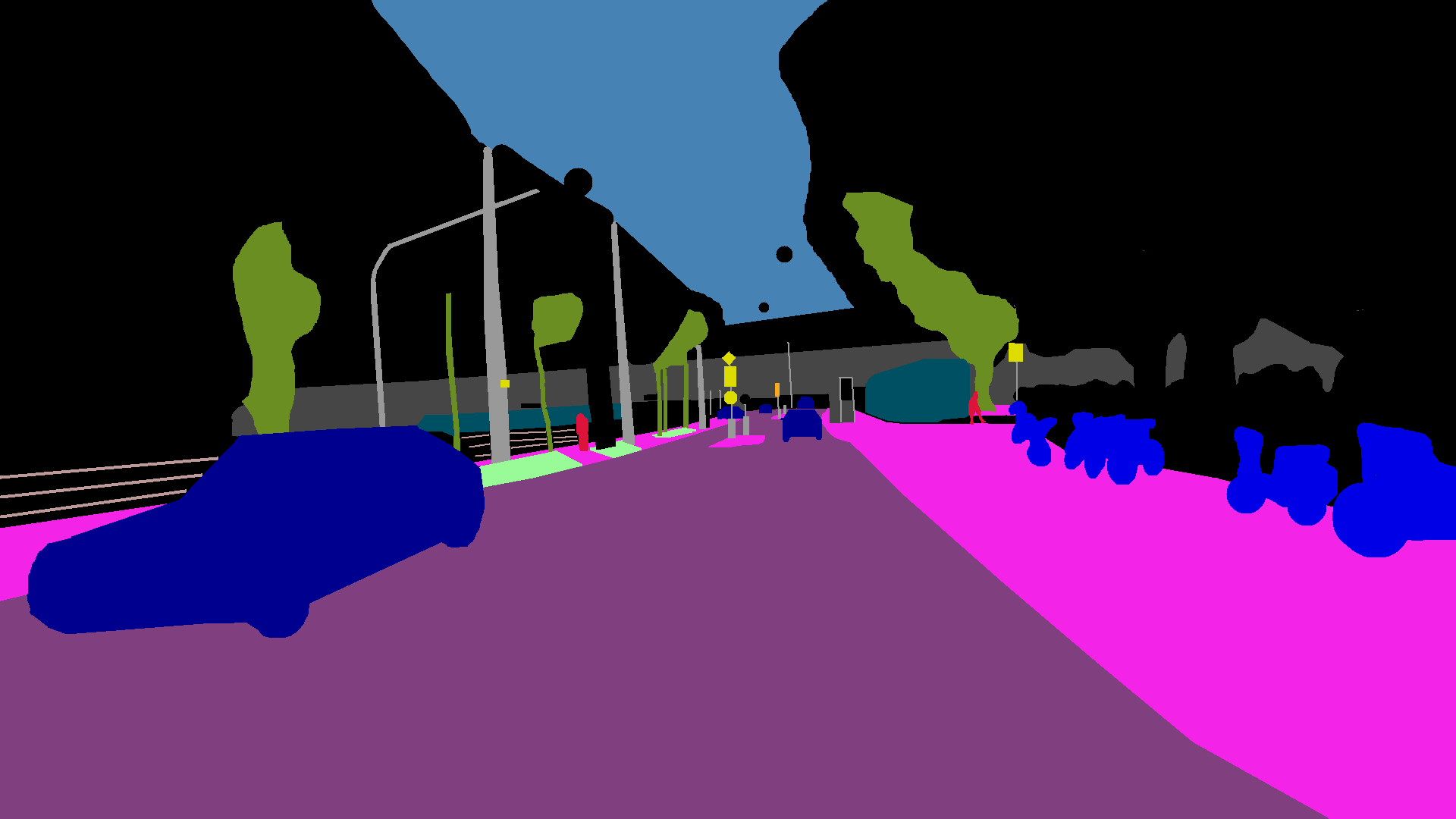}}
    \\
    \vspace{-0.3cm}
    \subfloat{\includegraphics[width=0.195\textwidth]{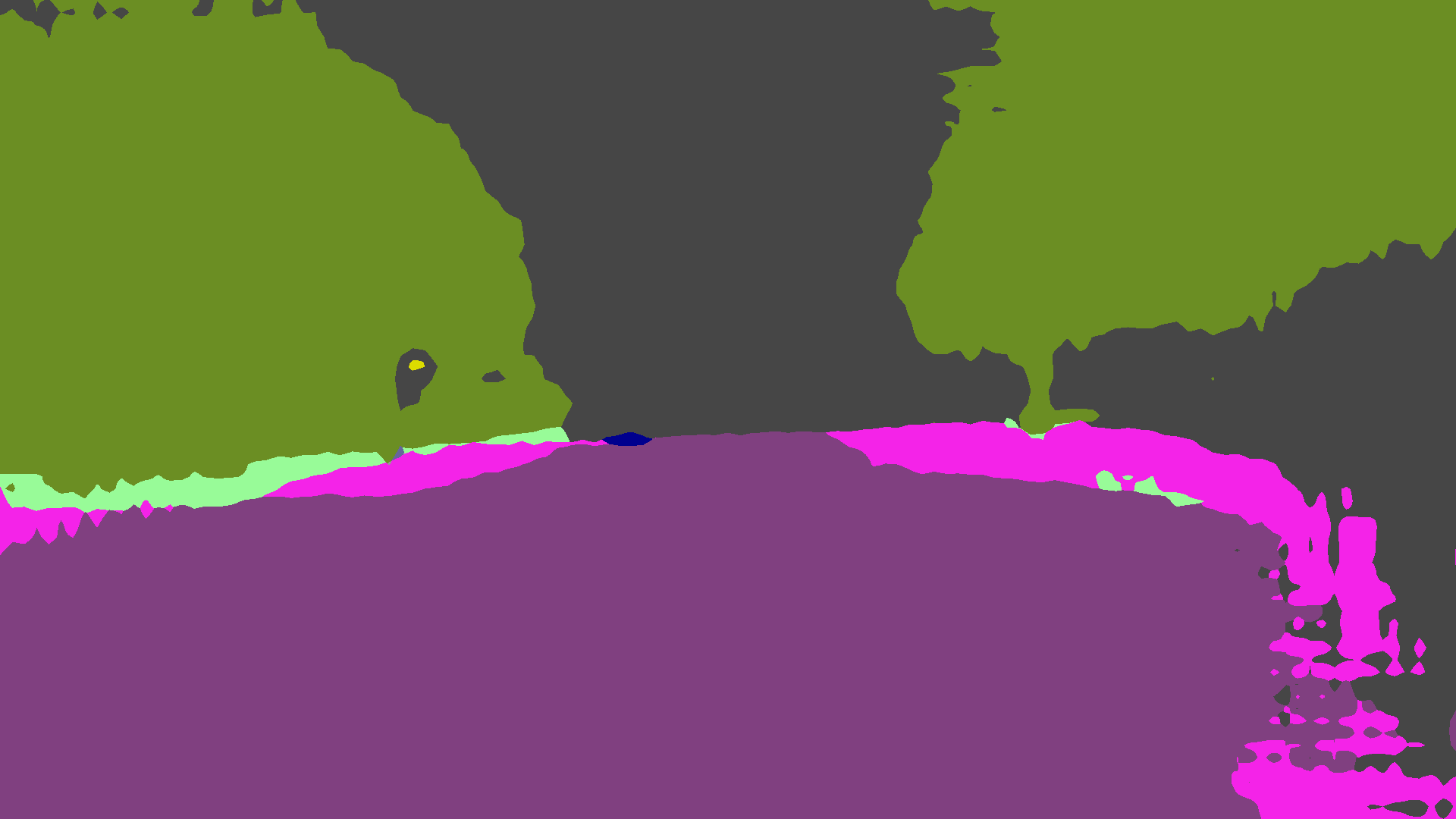}}
    \hfil
    \subfloat{\includegraphics[width=0.195\textwidth]{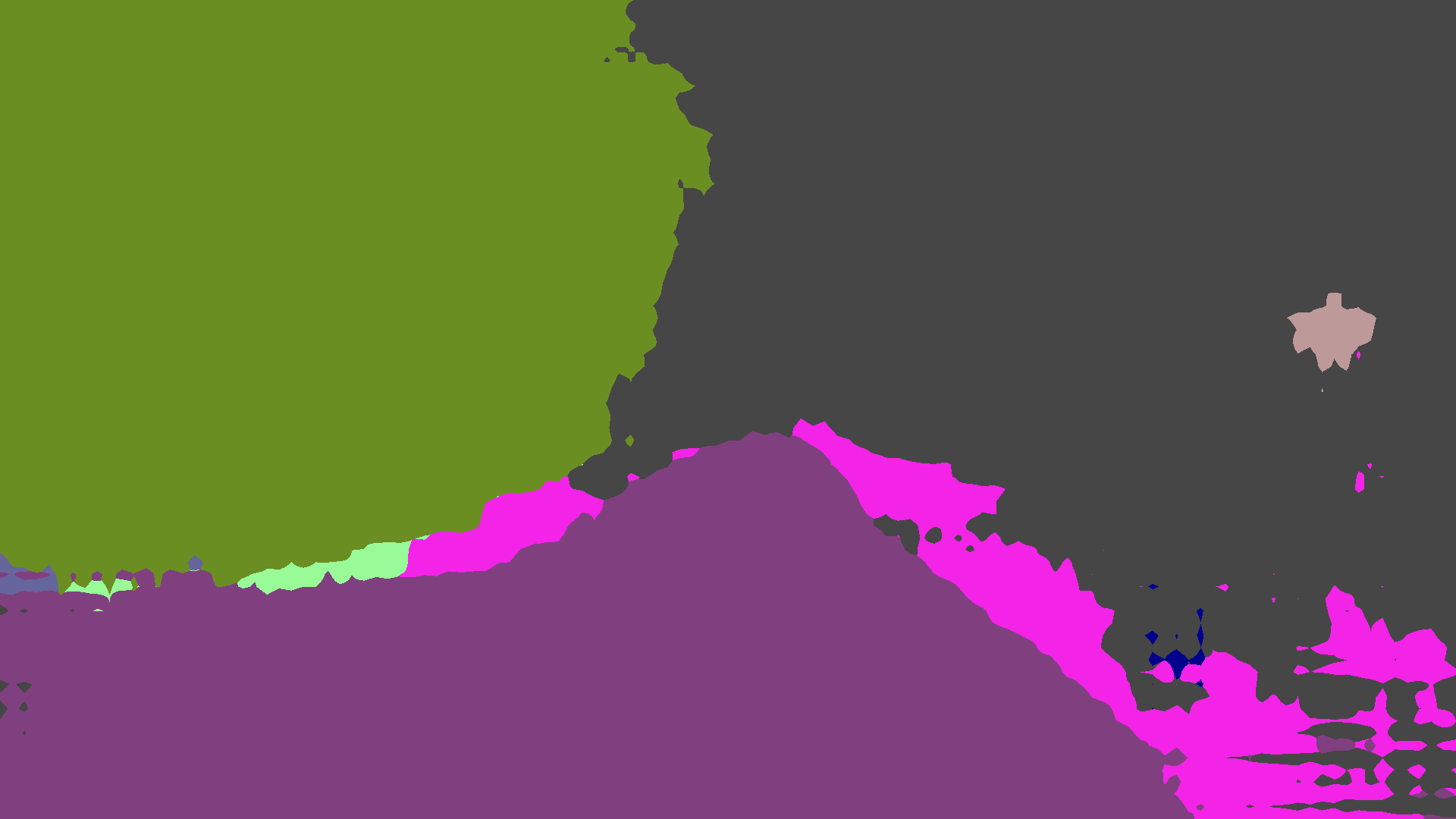}}
    \hfil
    \subfloat{\includegraphics[width=0.195\textwidth]{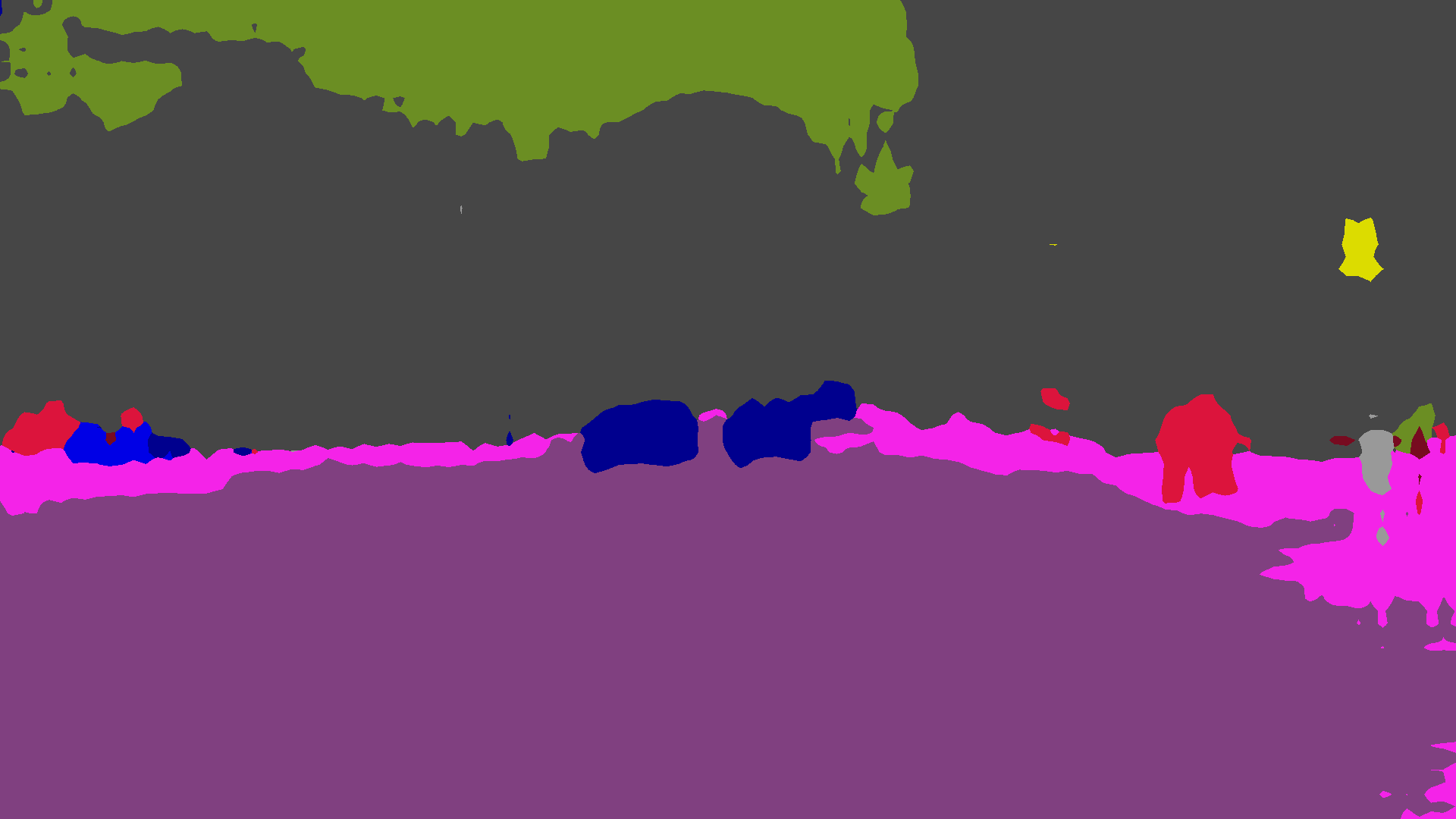}}
    \hfil
    \subfloat{\includegraphics[width=0.195\textwidth]{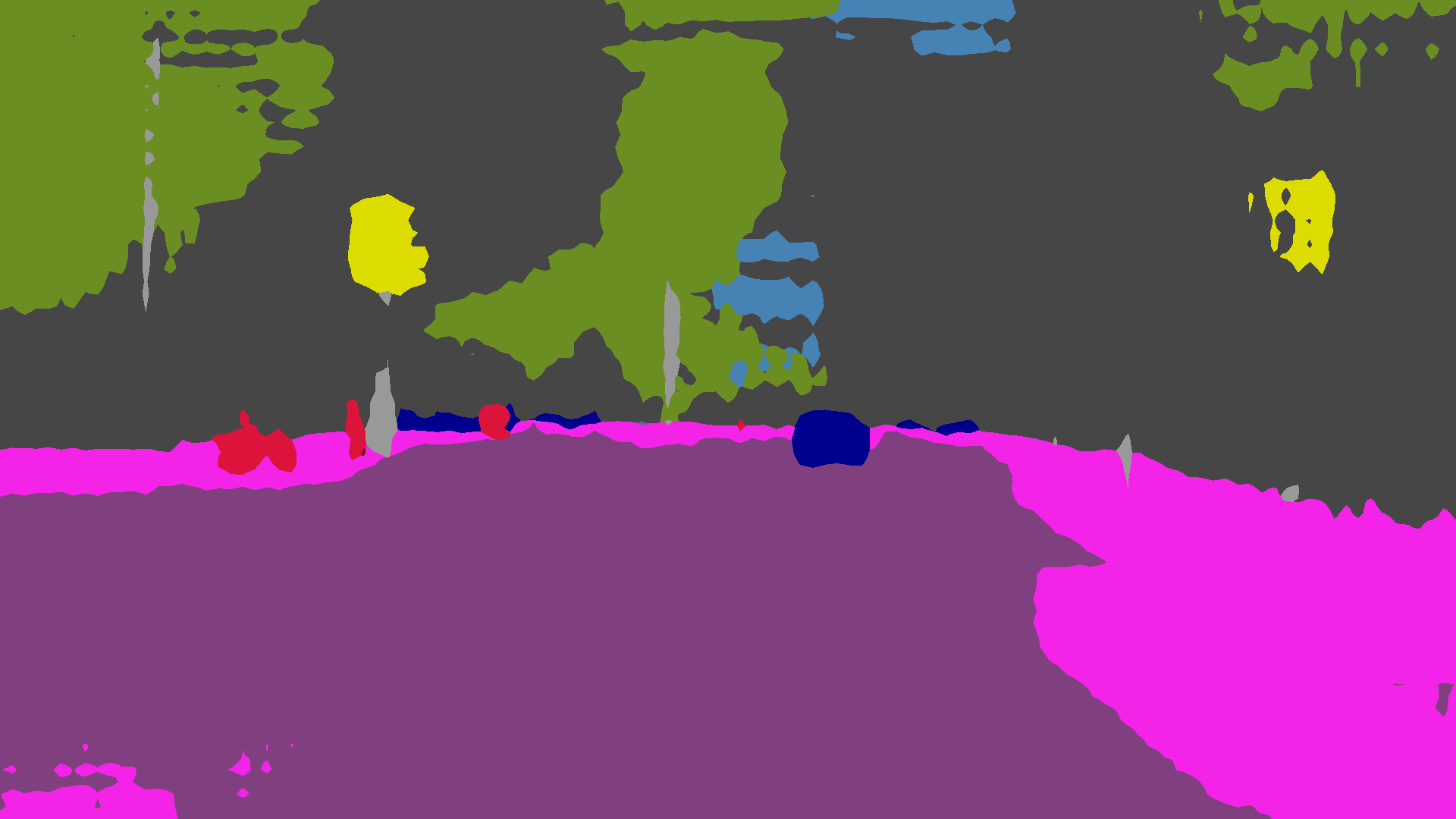}}
    \hfil
    \subfloat{\includegraphics[width=0.195\textwidth]{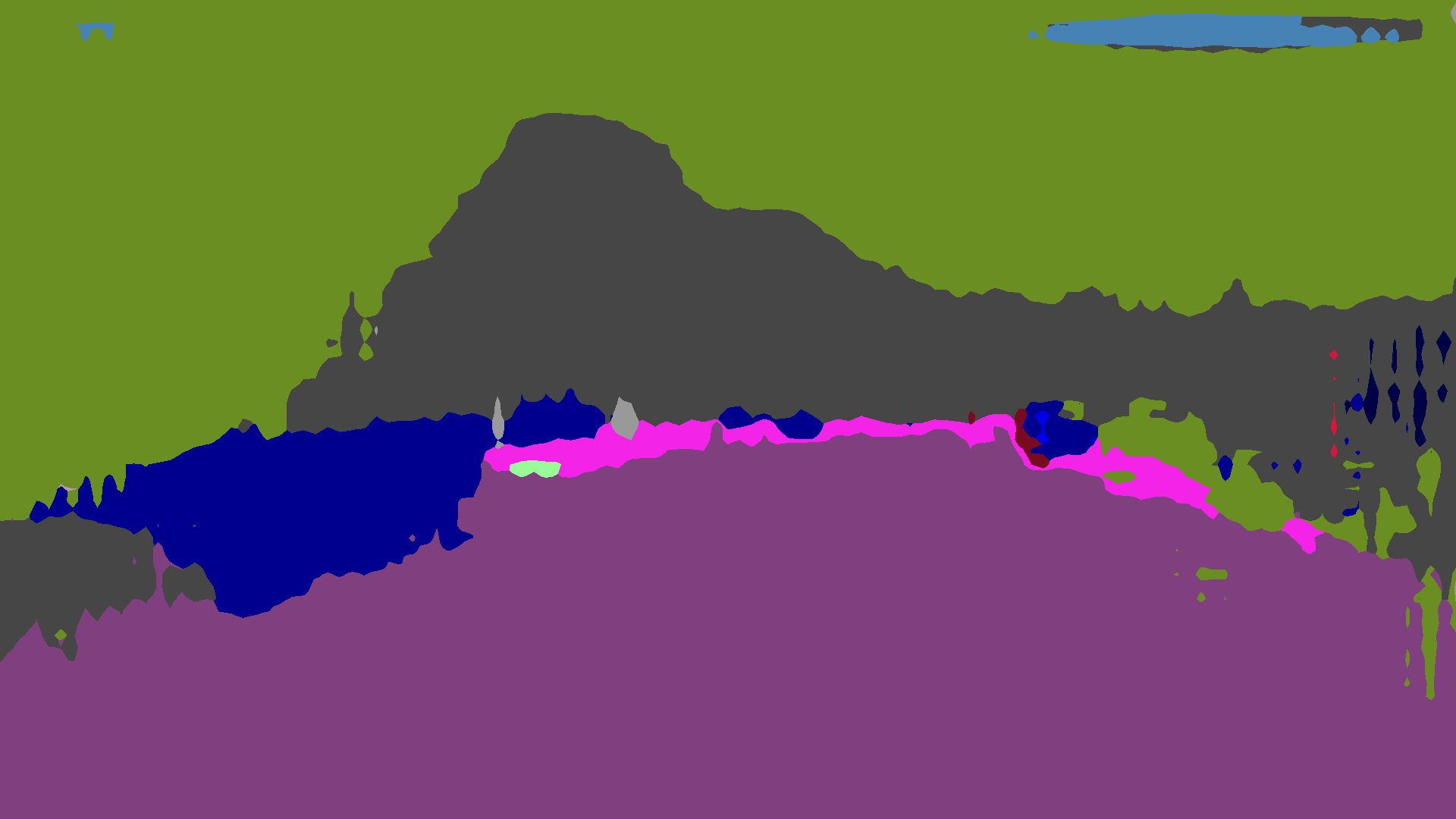}}
    \\
    \vspace{-0.3cm}
    \subfloat{\includegraphics[width=0.195\textwidth]{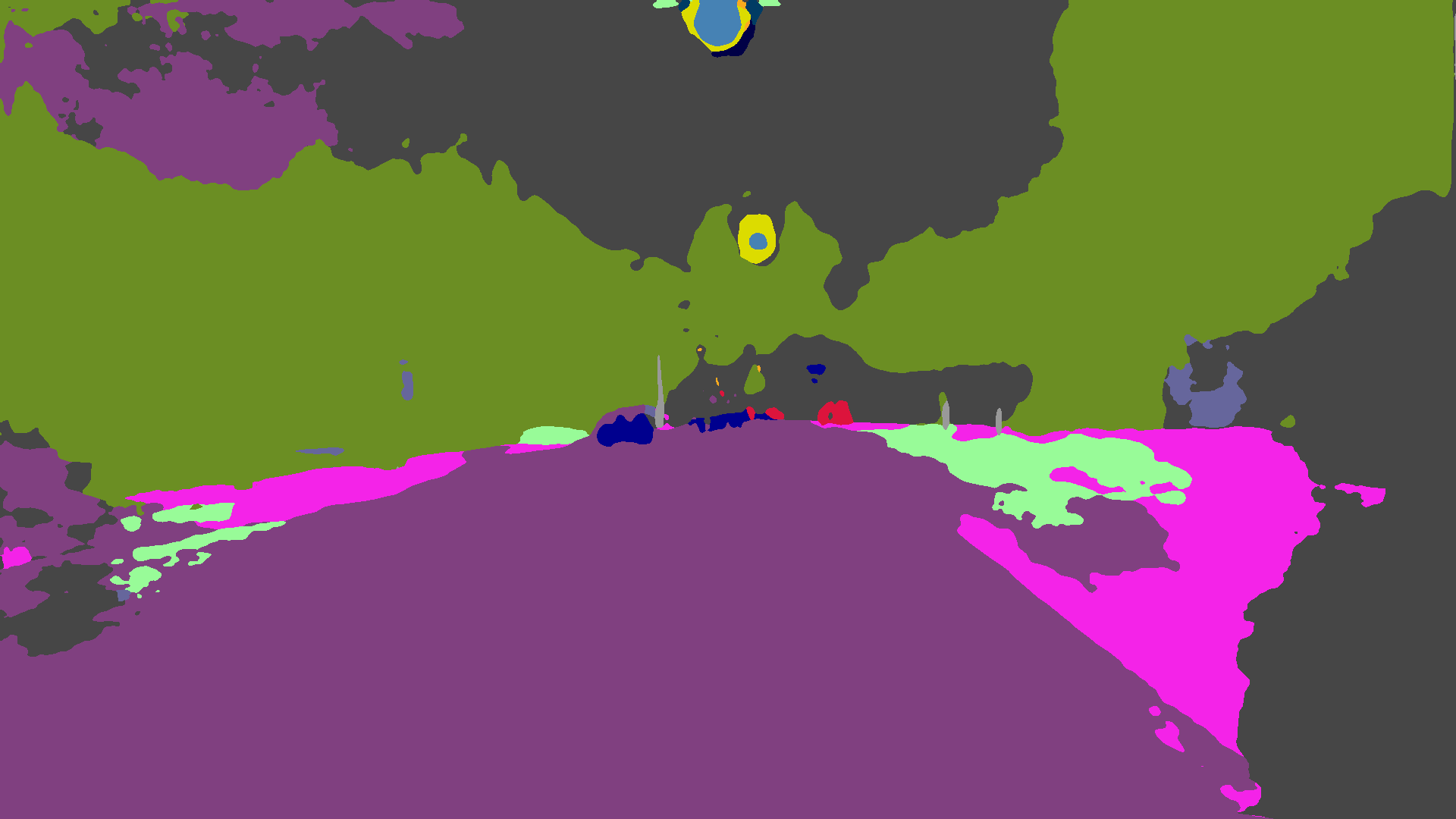}}
    \hfil
    \subfloat{\includegraphics[width=0.195\textwidth]{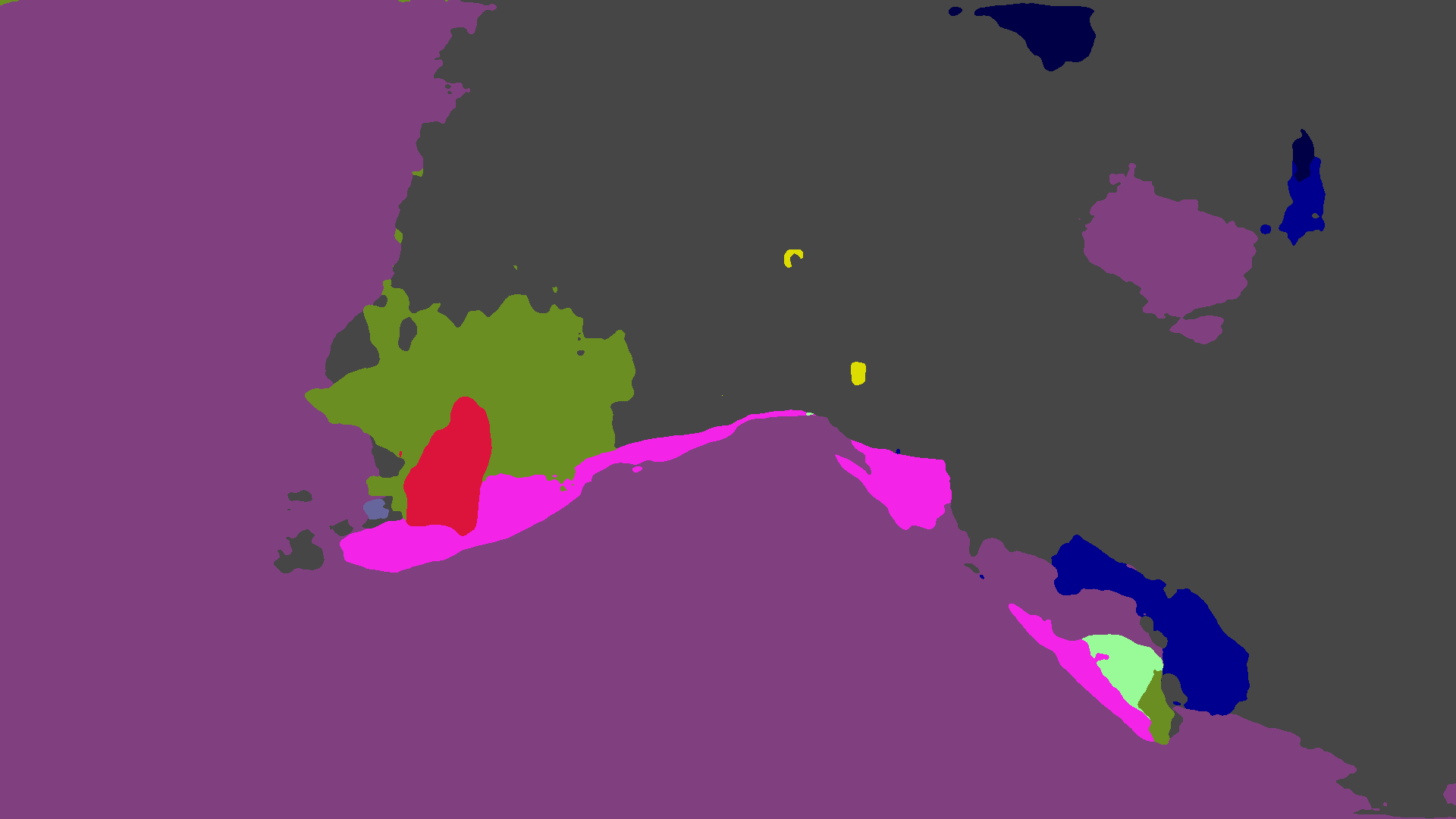}}
    \hfil
    \subfloat{\includegraphics[width=0.195\textwidth]{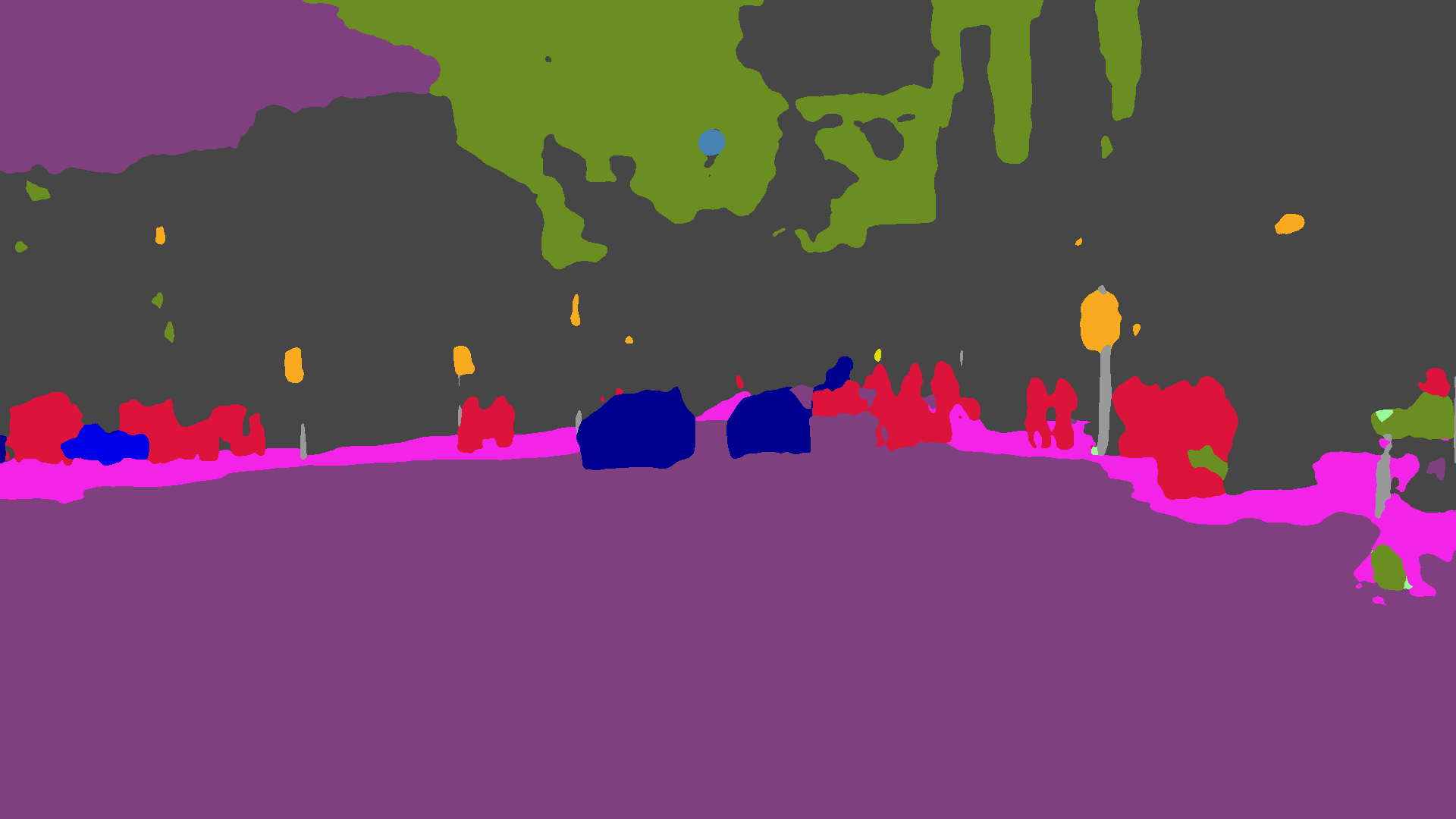}}
    \hfil
    \subfloat{\includegraphics[width=0.195\textwidth]{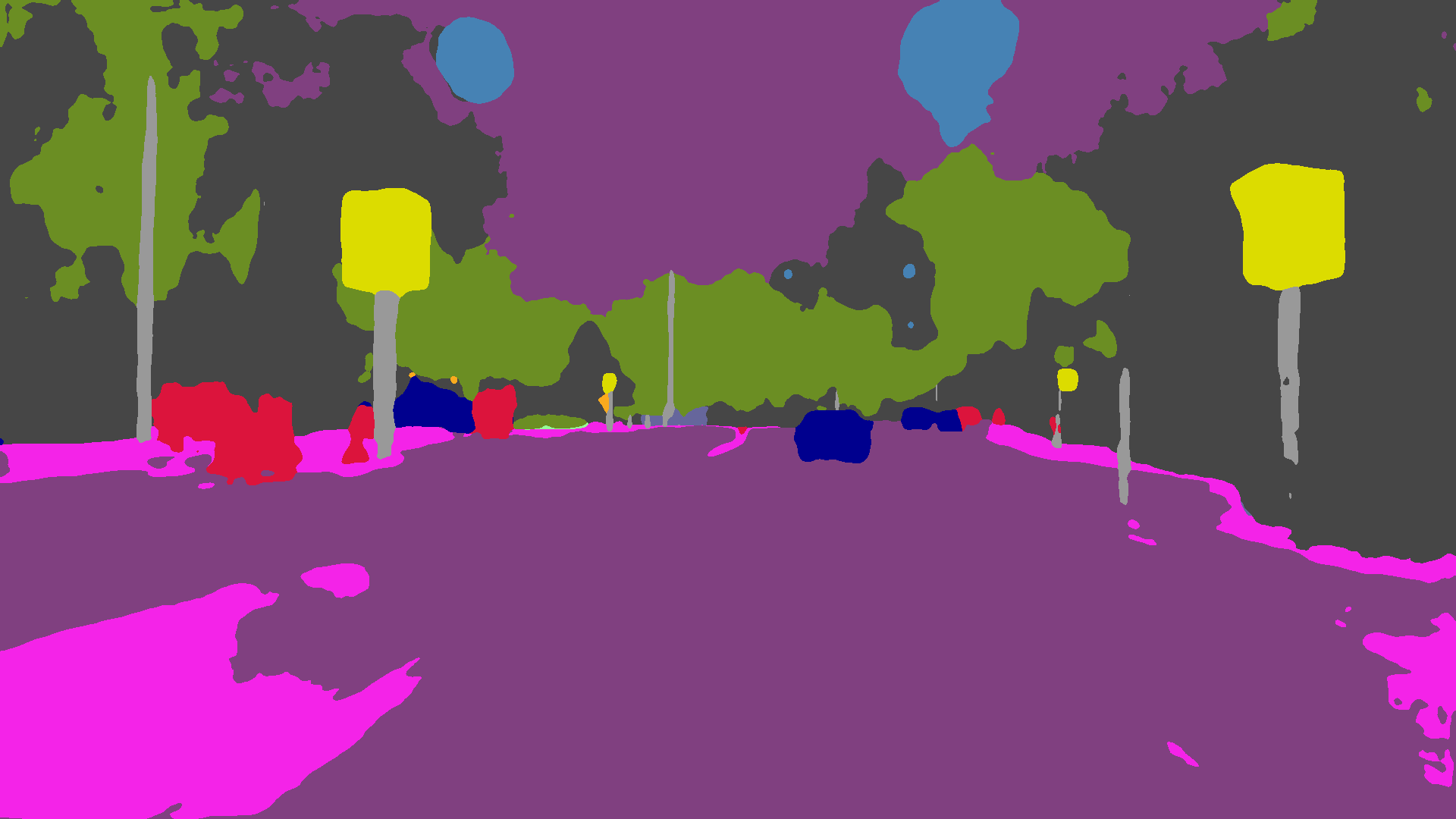}}
    \hfil
    \subfloat{\includegraphics[width=0.195\textwidth]{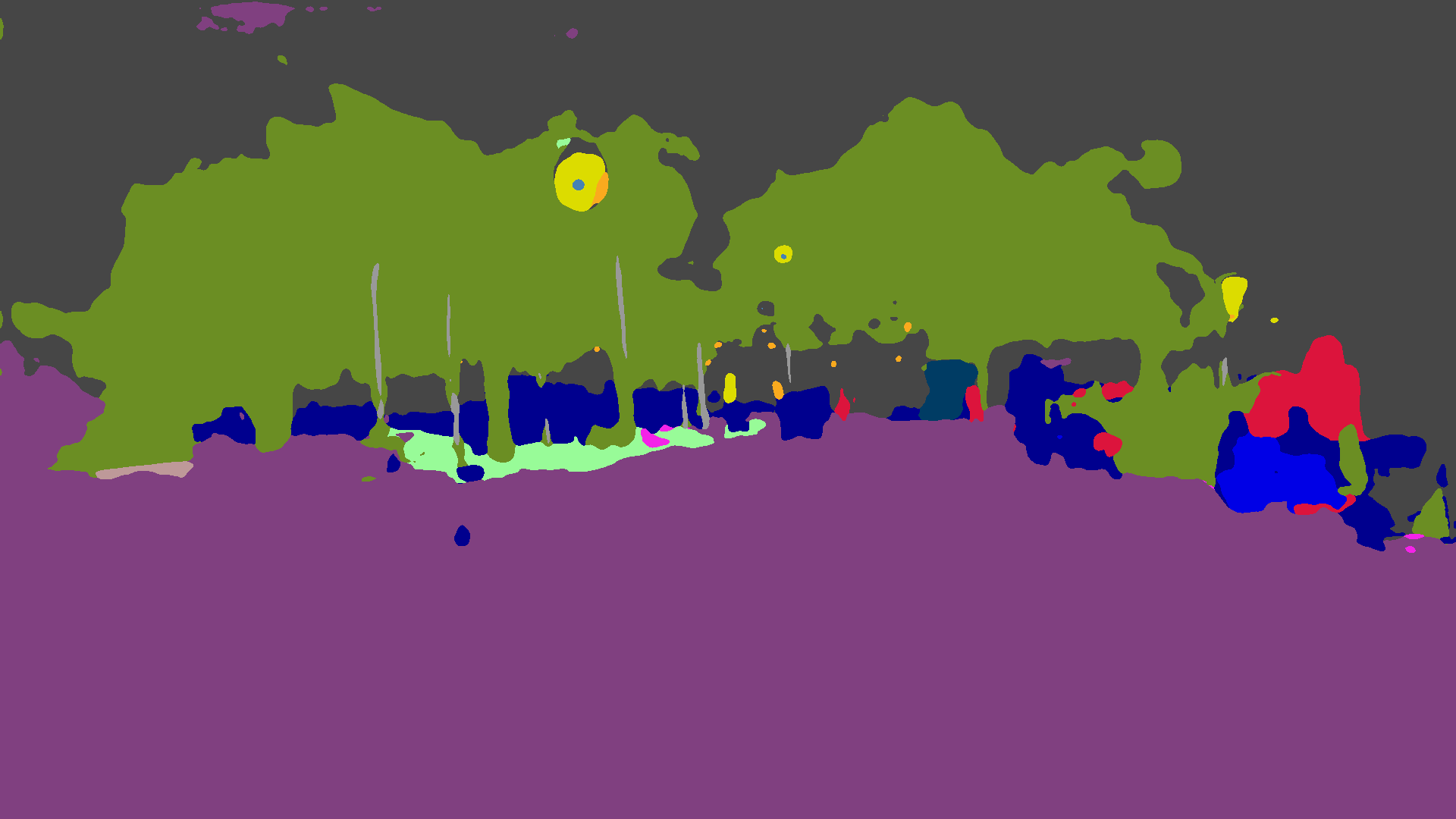}}
    \\
    \vspace{-0.3cm}
    \subfloat{\includegraphics[width=0.195\textwidth]{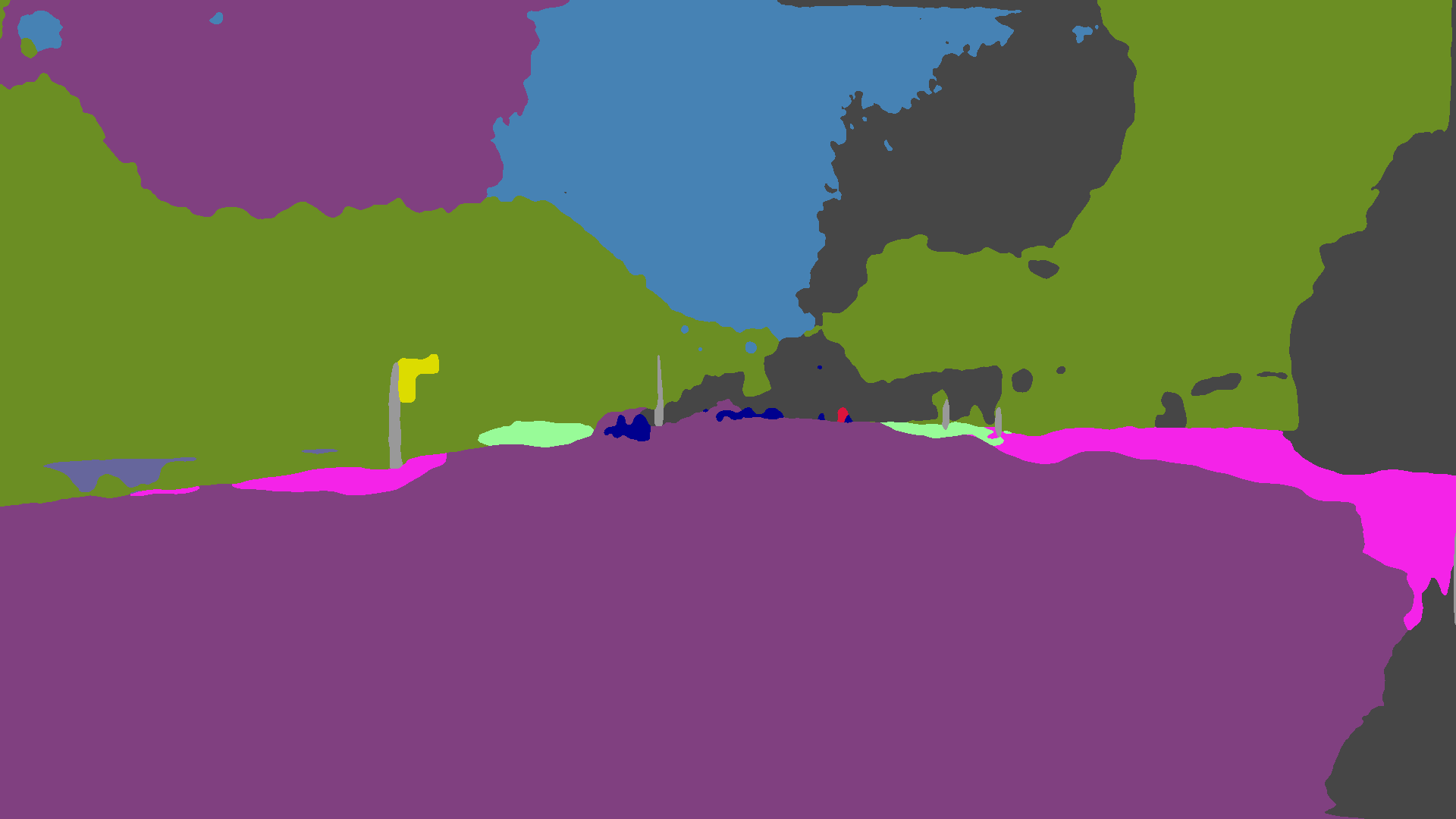}}
    \hfil
    \subfloat{\includegraphics[width=0.195\textwidth]{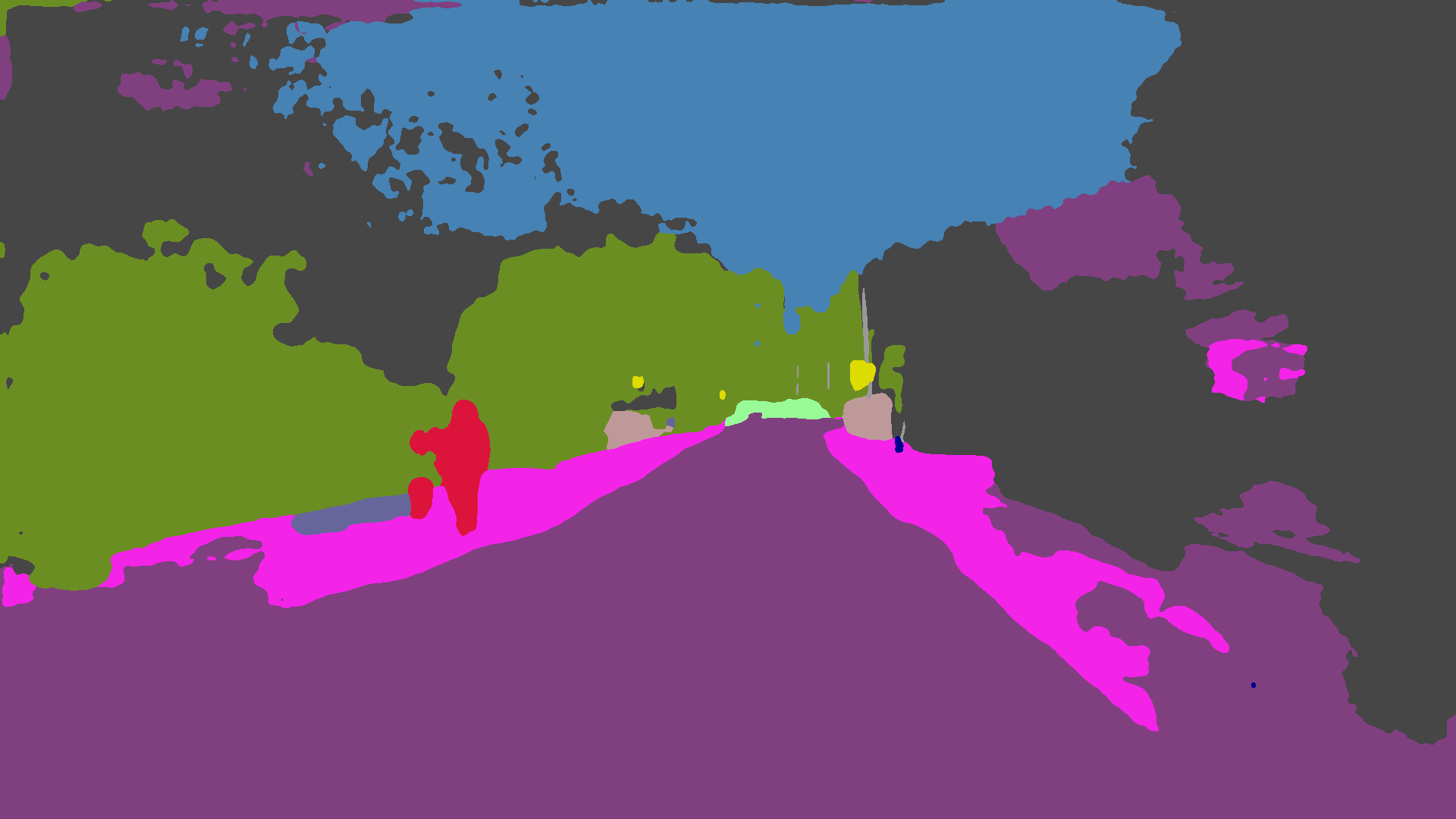}}
    \hfil
    \subfloat{\includegraphics[width=0.195\textwidth]{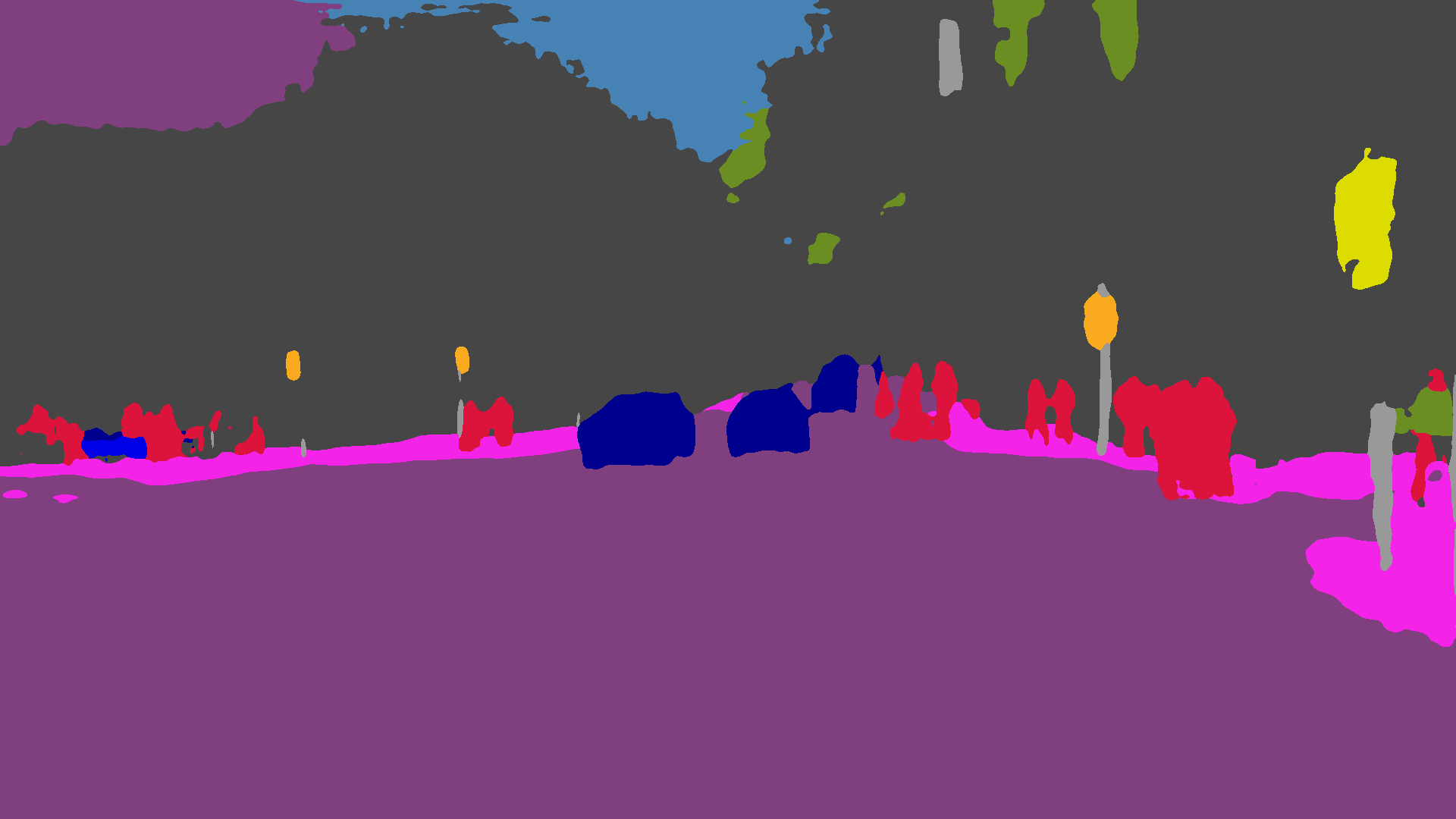}}
    \hfil
    \subfloat{\includegraphics[width=0.195\textwidth]{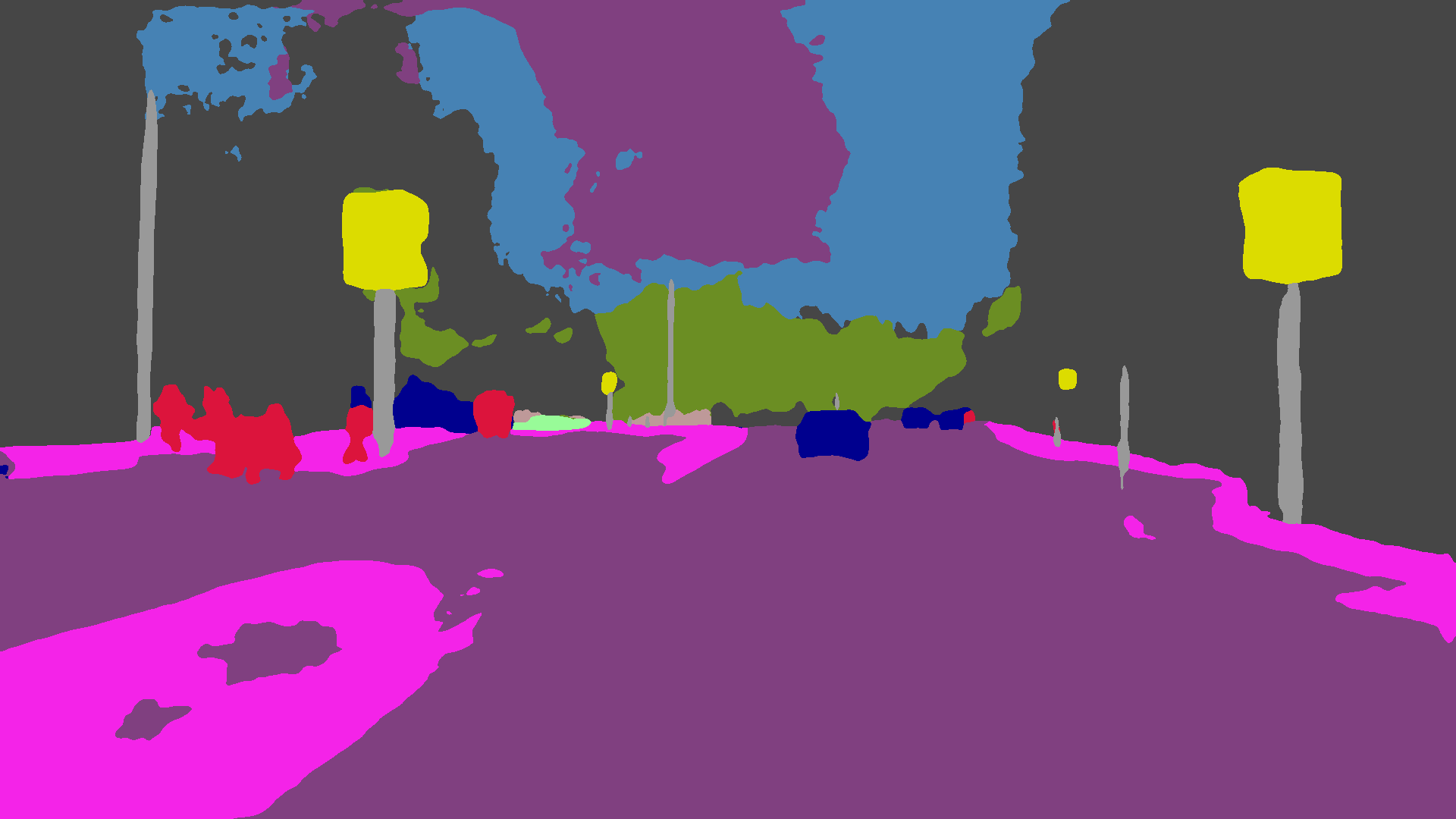}}
    \hfil
    \subfloat{\includegraphics[width=0.195\textwidth]{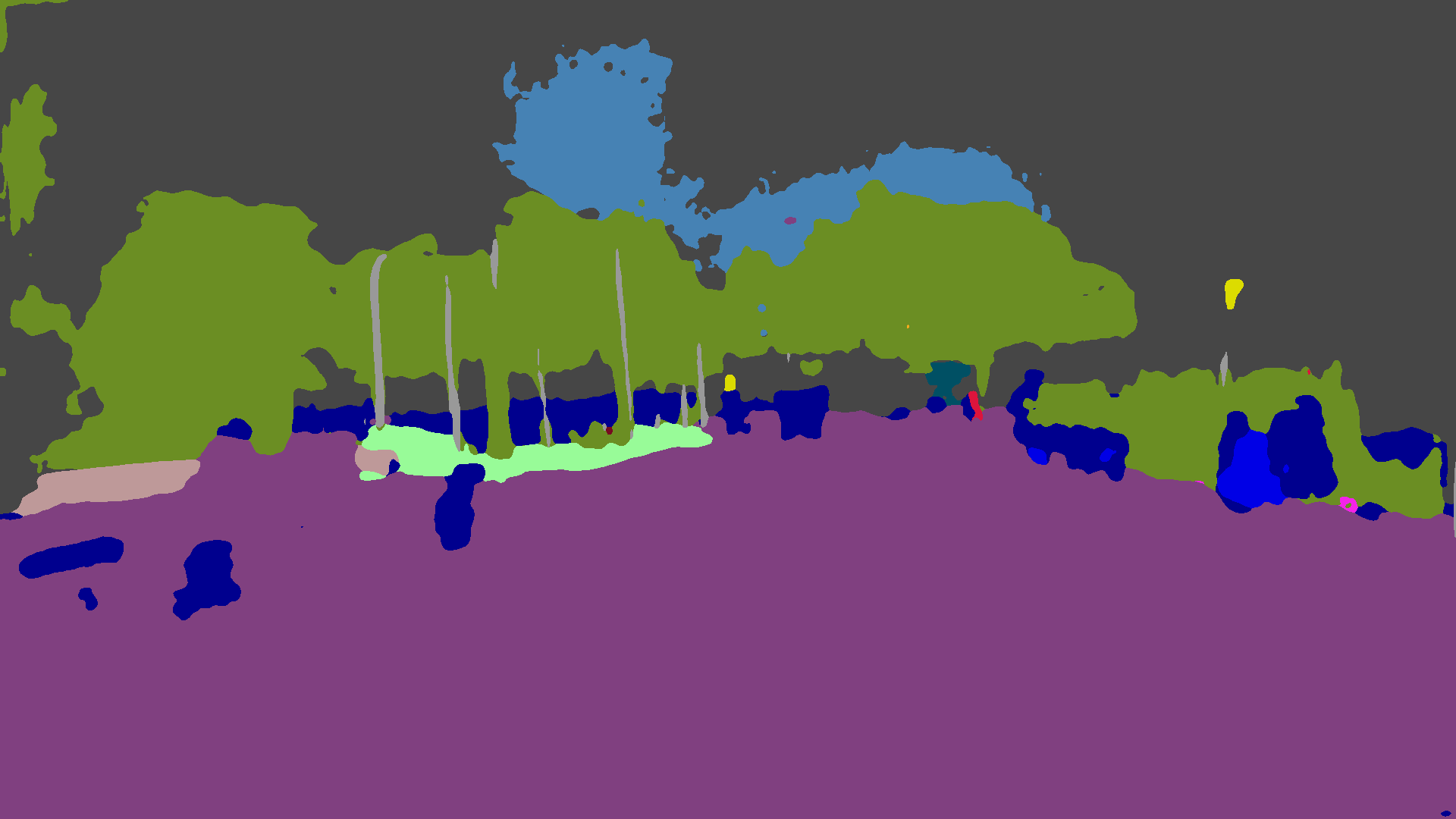}}
    \caption{Examples of our annotations and qualitative semantic segmentation results on \emph{Dark Zurich-test}. From top to bottom row: nighttime image, invalid mask annotation overlaid on the image (valid pixels are colored green), semantic annotation, AdaptSegNet~\cite{adapt:structured:output:cvpr18}, DMAda~\cite{daytime:2:nighttime}, and GCMA (ours).}
    \label{fig:sem:seg:extra}
\end{figure*}

In the first case, the first term in \eqref{eq:proof:nonempty:improved} is strictly positive, so \eqref{eq:proof:true:positive:iou:theta1} implies
\begin{equation} \label{eq:proof:correct:predict:ineq:strict}
|\text{TP}(1/C)| < |\text{TP}(\theta_1)| + |\text{TI}(\theta_1)|.
\end{equation}
We establish the inequality we are after by writing
\begin{align}
&{\text{IoU}} = \nonumber \\
&{=}\;\frac{|\text{TP}(1/C)|}{|\text{TP}(1/C)| + |\text{FN}(1/C)| + |\text{FP}(1/C)|} \nonumber \\
&{=}\;\frac{|\text{TP}(1/C)|}{|\text{TP}(\theta_1)| + |\text{FN}(\theta_1)| + |\text{TI}(\theta_1)| + |\text{FI}(\theta_1)| + |\text{FP}(1/C)|} \nonumber \\
&{\leq}\;\frac{|\text{TP}(1/C)|}{|\text{TP}(\theta_1)| + |\text{TI}(\theta_1)| + |\text{FP}(\theta_1)| + |\text{FN}(\theta_1)| + |\text{FI}(\theta_1)|} \nonumber \\
&{<}\;\frac{|\text{TP}(\theta_1)| + |\text{TI}(\theta_1)|}{|\text{TP}(\theta_1)| + |\text{TI}(\theta_1)| + |\text{FP}(\theta_1)| + |\text{FN}(\theta_1)| + |\text{FI}(\theta_1)|} \nonumber \\
&{=}\;\text{UIoU}(\theta_1),
\end{align}
where we have used the definition of IoU in the second line, \eqref{eq:proof:true:positive:false:negative:iou} in the third line, $\text{FP}(\theta_1) \subseteq \text{FP}(1/C)$ in the fourth line, \eqref{eq:proof:correct:predict:ineq:strict} in the fifth line, and the definition of UIoU that has been introduced in \eqref{eq:uiou} in the last line.

In the second case, the second term in \eqref{eq:proof:nonempty:improved} is strictly positive, which implies that
\begin{equation} \label{eq:proof:fp:ineq}
|\text{FP}(1/C)| > |\text{FP}(\theta_1)|.
\end{equation}
Besides, applying the nonnegativity of the first term in \eqref{eq:proof:nonempty:improved} to \eqref{eq:proof:true:positive:iou:theta1} leads to
\begin{equation} \label{eq:proof:correct:predict:ineq}
|\text{TP}(1/C)| \leq |\text{TP}(\theta_1)| + |\text{TI}(\theta_1)|.
\end{equation}
Similarly to the first case, we establish the inequality we are after by writing
\begin{align}
&{\text{IoU}} = \nonumber \\
&{=}\;\frac{|\text{TP}(1/C)|}{|\text{TP}(\theta_1)| + |\text{TI}(\theta_1)| + |\text{FP}(1/C)| + |\text{FN}(\theta_1)| + |\text{FI}(\theta_1)|} \nonumber \\
&{<}\;\frac{|\text{TP}(1/C)|}{|\text{TP}(\theta_1)| + |\text{TI}(\theta_1)| + |\text{FP}(\theta_1)| + |\text{FN}(\theta_1)| + |\text{FI}(\theta_1)|} \nonumber \\
&{\leq}\;\frac{|\text{TP}(\theta_1)| + |\text{TI}(\theta_1)|}{|\text{TP}(\theta_1)| + |\text{TI}(\theta_1)| + |\text{FP}(\theta_1)| + |\text{FN}(\theta_1)| + |\text{FI}(\theta_1)|} \nonumber \\
&{=}\;\text{UIoU}(\theta_1),
\end{align}
where we have used the definition of IoU as well as \eqref{eq:proof:true:positive:false:negative:iou} in the second line, \eqref{eq:proof:fp:ineq} in the third line, \eqref{eq:proof:correct:predict:ineq} in the fourth line, and the definition of UIoU in the last line.
\end{proof}

\section{Additional Qualitative Results}
\label{supp:sec:results}

In Fig.~\ref{fig:sem:seg:extra}, we compare our GCMA approach against AdaptSegNet~\cite{adapt:structured:output:cvpr18} and DMAda~\cite{daytime:2:nighttime} on additional images from \emph{Dark Zurich-test}, further demonstrating the superiority of GCMA. For these images, we also present our annotations for invalid masks and semantic labels, which show that a significant portion of ground-truth invalid regions is indeed assigned a reliable semantic label through our annotation protocol and can thus be included in the evaluation.

\begin{figure}
    \centering
    \subfloat[$\mathcal{D}^1_{lr}$: Cityscapes]{\includegraphics[height=0.115\textwidth]{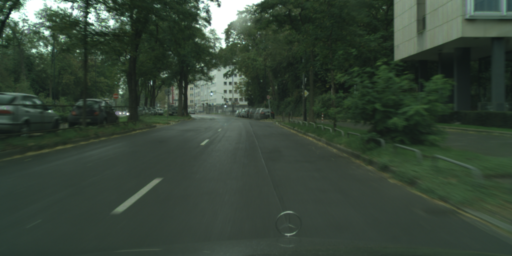}}
    \hfil
    \subfloat[$\mathcal{D}^1_{ur}$: \emph{Dark Zurich-day}]{\includegraphics[height=0.115\textwidth]{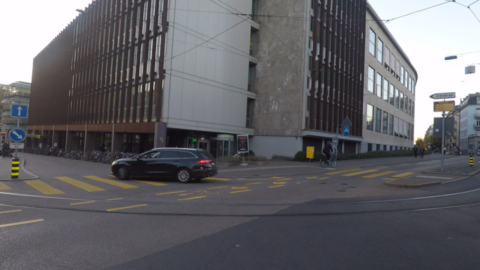}}
    \\
    \vspace{-0.3cm}
    \subfloat[$\mathcal{D}^2_{ls}$: Cityscapes-twilight style]{\includegraphics[height=0.115\textwidth]{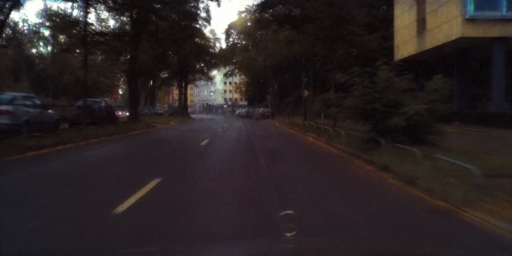}}
    \hfil
    \subfloat[$\mathcal{D}^2_{ur}$: \emph{Dark Zurich-twilight}]{\includegraphics[height=0.115\textwidth]{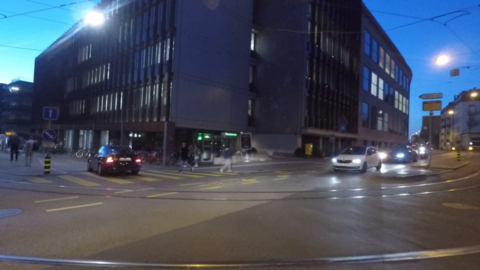}}
    \\
    \vspace{-0.3cm}
    \subfloat[$\mathcal{D}^3_{ls}$: Cityscapes-nighttime style]{\includegraphics[height=0.115\textwidth]{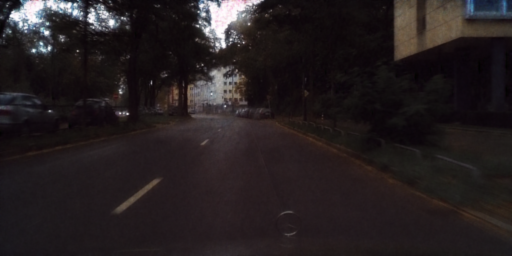}}
    \hfil
    \subfloat[$\mathcal{D}^3_{ur}$: \emph{Dark Zurich-night}]{\includegraphics[height=0.115\textwidth]{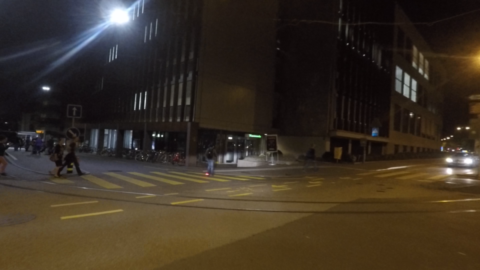}}
    \caption{Sample images from the training sets used in GCMA.}
    \label{fig:training:datasets}
\end{figure}

\section{Configuration of Training Sets for GCMA}
\label{supp:sec:training:datasets}

In Fig.~\ref{fig:training:datasets}, we show examples from the six training sets we introduced in Sec.~\ref{sec:gcma:general}, which are used for implementing GCMA. Cityscapes is used to instantiate the labeled sets, while \emph{Dark Zurich} is used for the unlabeled sets.

\begin{figure*}[!tb]
    \centering
    \subfloat{\includegraphics[width=0.24\textwidth]{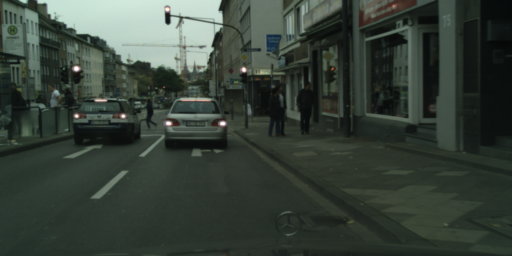}}
    \hfil
    \subfloat{\includegraphics[width=0.24\textwidth]{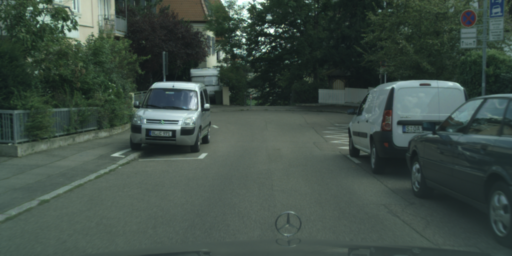}}
    \hfil
    \subfloat{\includegraphics[width=0.24\textwidth]{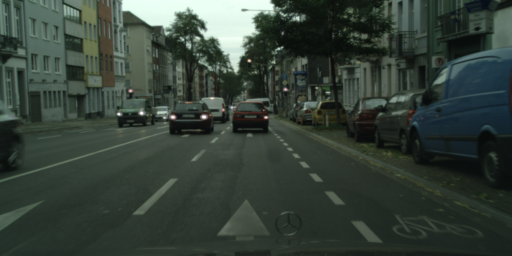}}
    \hfil
    \subfloat{\includegraphics[width=0.24\textwidth]{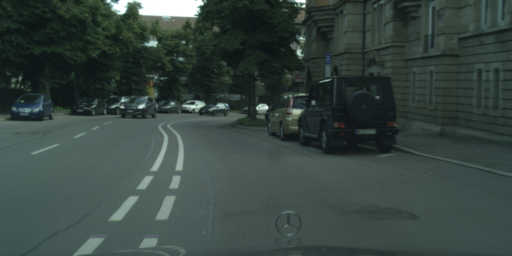}}
    \\
    \subfloat{\includegraphics[width=0.24\textwidth]{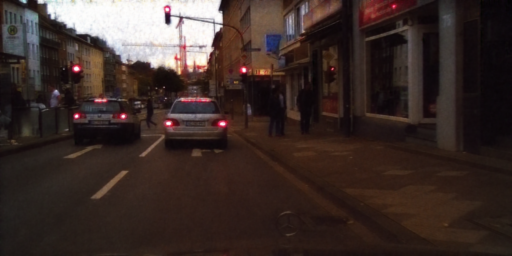}}
    \hfil
    \subfloat{\includegraphics[width=0.24\textwidth]{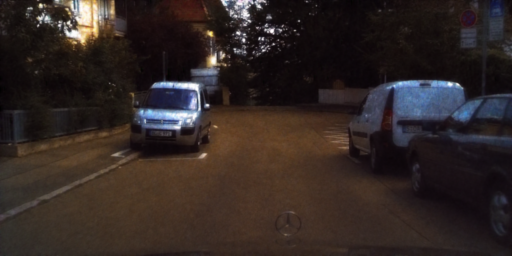}}
    \hfil
    \subfloat{\includegraphics[width=0.24\textwidth]{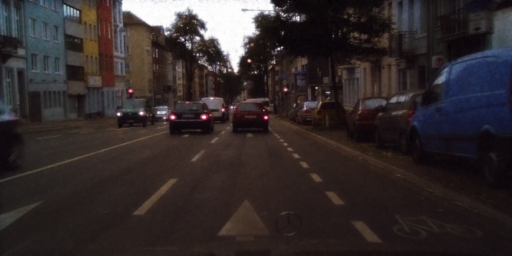}}
    \hfil
    \subfloat{\includegraphics[width=0.24\textwidth]{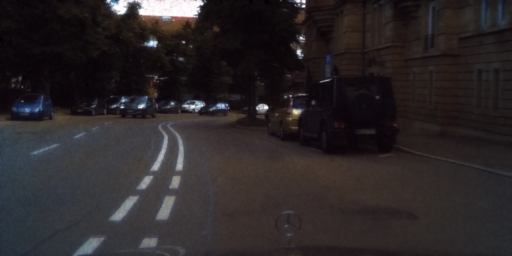}}
    \caption{Top row: Examples of images from Cityscapes ($\mathcal{D}^1_{lr}$ in GCMA), bottom row: corresponding images from Cityscapes-nighttime style ($\mathcal{D}^3_{ls}$ in GCMA).}
    \label{fig:stylized:night:examples}
\end{figure*}

More examples of Cityscapes images stylized to nighttime using a CycleGAN model~\cite{cycleGAN} that is trained to translate Cityscapes to \emph{Dark Zurich-night} are presented in Fig.~\ref{fig:stylized:night:examples}.

\section{Parameter Selection for Prediction Fusion}
\label{supp:sec:params:fusion}

\begin{figure*}[!tb]
    \centering
    \subfloat[Dark image $I^z$]{\includegraphics[width=0.24\textwidth]{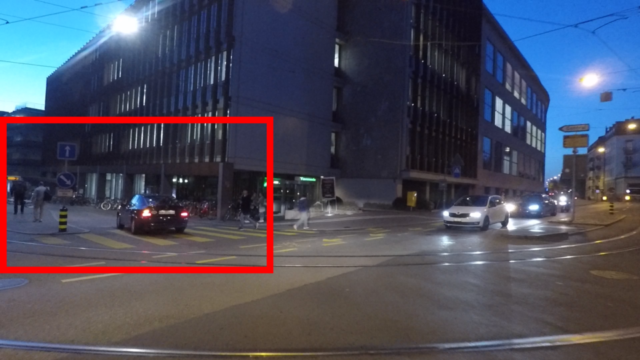}}
    \hfil
    \subfloat[$\alpha_l = \alpha_h = 1$]{\includegraphics[width=0.24\textwidth]{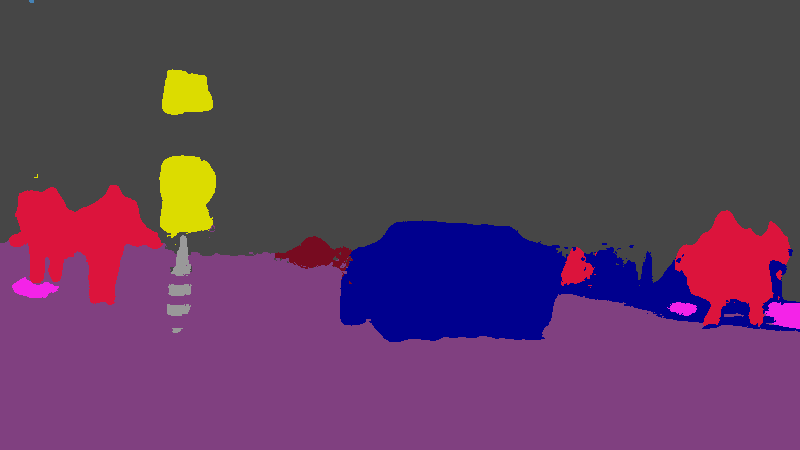}\label{fig:params:fusion:comparison:a1}}
    \hfil
    \subfloat[$\alpha_l = \alpha_h = 0.6$]{\includegraphics[width=0.24\textwidth]{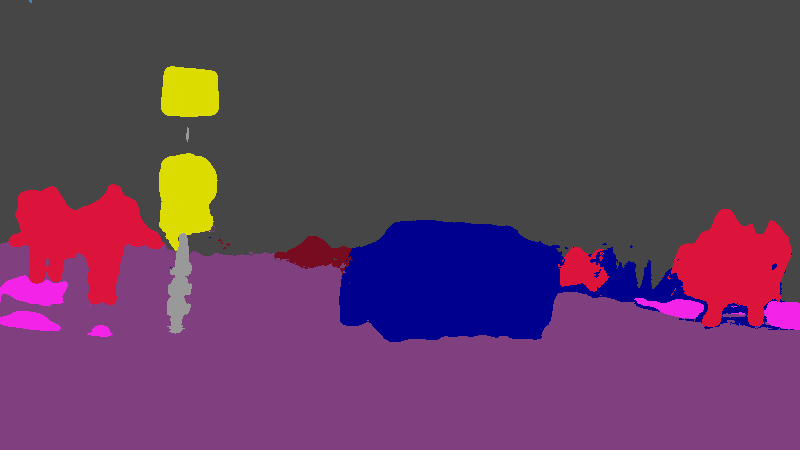}\label{fig:params:fusion:comparison:a06}}
    \hfil
    \subfloat[$\alpha_l = 0.3$, $\alpha_h = 0.6$, $\eta = 0.2$]{\includegraphics[width=0.24\textwidth]{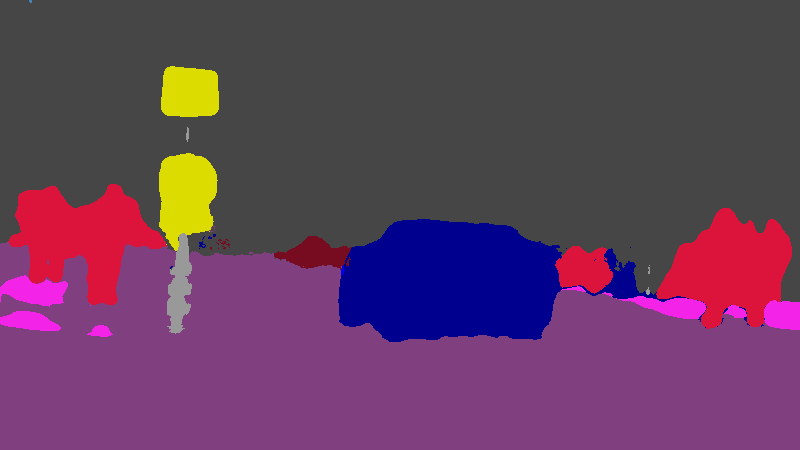}\label{fig:params:fusion:comparison:aL03aH06}}
    \caption{Dark image $I^z$ from \emph{Dark Zurich} and our refined predictions $\hat{\mathbf{S}}^{z}$ for the region indicated by the red box for different values of the parameters involved in the proposed confidence-adaptive prediction fusion. When $\alpha_l = \alpha_h$, reducing $\alpha_h$ to a value lower than $1$, e.g.\ \protect\subref{fig:params:fusion:comparison:a1}$\rightarrow$\protect\subref{fig:params:fusion:comparison:a06}, reduces false positives and/or false negatives both for static and dynamic classes, e.g.\ \emph{pole}, \emph{sidewalk}, \emph{road} and \emph{car}. When $\alpha_h < 1$, reducing $\alpha_l$ to a value lower than $\alpha_h$, e.g.\ \protect\subref{fig:params:fusion:comparison:a06}$\rightarrow$\protect\subref{fig:params:fusion:comparison:aL03aH06}, improves accuracy on pixels that are assigned to a dynamic class in either prediction, e.g.\ \emph{car}, because of the formulation of \eqref{eq:alpha}. Best viewed with zoom.}
    \label{fig:params:fusion:comparison}
\end{figure*}

For our confidence-adaptive prediction fusion, we demonstrate the benefit of selecting $\alpha_l < \alpha_h < 1$---the rationale of which is exposed in Sec.~\ref{sec:gcma:guidance:fusion}---through a visual example in Fig.~\ref{fig:params:fusion:comparison}.

\end{document}